\providecommand{\usingElsarticleFinal}{1}
\if \usingElsarticleFinal 1
  \newcommand{\doitext}{10.1016/j.eswa.2025.130564}
  \newcommand{\doilink}{https://doi.org/10.1016/j.eswa.2025.130564}
  \documentclass[preprint,10pt,3p]{elsarticle-final}
  \else
  \documentclass[preprint,10pt,3p]{elsarticle}
  \fi

\newcommand{\flatstructure}{1}
\if \flatstructure 1
  \usepackage[flat]{dhnamlib-prefix}
  \else
  \usepackage[recursive]{dhnamlib-prefix}
  \fi

\usepackage{config}
\usepackage{elsarticle-preprint-config}

\input{package.tex\inputfilesuffix}
\input{command.tex\inputfilesuffix}

\IfFileExists{./arxiv-config.tex}{
  
\newcommand{\includingHighlights}{0}
\newcommand{\forArxiv}{1}

}{
}

\providecommand{\showJournalSubmission}{1}
\if \showJournalSubmission 1
  \journal{Expert Systems with Applications}
  \else
  \fi

\begin{document}

\begin{frontmatter}
  \input{frontmatter.tex\inputfilesuffix}
\end{frontmatter}



\section{Introduction}
Semantic parsing is the task of mapping natural language to logical forms \citep{DBLP:journals/coling/WarrenP82}, which can be evaluated on given knowledge bases (KBs) to produce corresponding denotations \citep{DBLP:conf/acl/LiangJK11}.
For example, a question answering system can use a semantic parser to convert a user's question to a query (logical form), which then derives an answer (denotation) from a database (KB) \citep{DBLP:conf/aaai/ZelleM96,DBLP:conf/acl/CaiY13}.
Traditional semantic parsers depend on grammars with lexicons that map spans of utterances to atomic units, which are subsequently composed into logical forms by following the grammars \citep{DBLP:conf/uai/ZettlemoyerC05,DBLP:conf/acl/WongM07,DBLP:conf/acl/LiangJK11}.
In contrast, the emergence of sequence-to-sequence (seq2seq) frameworks \citep{DBLP:conf/nips/SutskeverVL14,DBLP:journals/corr/BahdanauCB14} has led to the development of neural semantic parsers in which neural networks convert natural language token sequences to action sequences that construct logical forms \citep{DBLP:conf/acl/JiaL16,DBLP:conf/acl/DongL16}.

Neural semantic parsers have used grammars that utilize types for constrained action decoding, in which the actions are designed to generate only well-typed logical forms.
The actions can be defined as production rules that expand typed placeholders to sub-expressions of logical forms \citep{DBLP:conf/acl/YinN17,DBLP:conf/acl/RabinovichSK17,DBLP:conf/emnlp/KrishnamurthyDG17},
or as typed atomic units that are inserted into partially constructed logical forms \citep{DBLP:conf/acl/GuuPLL17,DBLP:conf/acl/ChengRSL17,DBLP:conf/acl/LiangBLFL17,DBLP:conf/acl/LapataD18,DBLP:conf/acl/BerantGGLN18}.
In particular, semantic parsers that take production rules as actions are easily adapted to diverse applications with different logical form languages, once the corresponding production rules are defined.


Recent work has incorporated grammars into semantic parsers based on seq2seq pre-trained language models (PLMs) \citep{DBLP:conf/acl/LewisLGGMLSZ20,DBLP:journals/jmlr/RaffelSRLNMZLL20} or large language models (LLMs) \citep{DBLP:conf/nips/BrownMRSKDNSSAA20,DBLP:journals/corr/abs-2107-03374}, where the models have decoders and tokenizers.
The semantic parsers sequentially generate tokens that extend prefixes of logical forms under the guidance of the grammars that keep the prefixes always valid.
The semantic parsers use context-free grammars (CFGs) to ensure syntactic validity \citep{DBLP:conf/acl/Wu0XH0ZCYC20,DBLP:conf/emnlp/ShinLTCRPPKED21,DBLP:conf/nips/WangW0CSK23},
and additionally use context-sensitive constraints to ensure semantic validity \citep{DBLP:conf/emnlp/ScholakSB21,DBLP:conf/iclr/PoesiaP00SMG22}.

However, the grammars for semantic parsing with seq2seq PLMs or LLMs lack the ability to utilize \emph{large} information of KBs.
The information includes KB elements, such as entities or relations, and categories that the elements belong to.
As logical forms contain representations of KB elements, the information of KBs is necessary to decode the logical forms.
Previous work has generated valid KB elements with the guidance of trie data structures \citep{DBLP:books/daglib/0023376} that store entities \citep{DBLP:conf/iclr/CaoI0P21} or predicates \citep{DBLP:conf/emnlp/ShuYLKMQL22};
however, incorporating the method into semantic parsing is difficult when logical forms have complex syntax with various types.


\begin{revision}
  In contrast, recent work on knowledge base question answering (KBQA) has focused on retrieving information from large KBs by conditioning on the input questions.
  %
  %
  A semantic parser can use the retrieved information to better predict a logical form that produces a denotation as an answer.
  The retrieved information, such as sub-graphs of a KB, can be fed into a seq2seq PLM or an LLM to generate a logical form \citep{DBLP:conf/emnlp/ShuYLKMQL22,DBLP:conf/acl/YeYHZX22,DBLP:conf/emnlp/LiuYMRXZ22,DBLP:conf/aaai/00010BZ0025,DBLP:conf/aaai/Xu0ZLZ0000C25},
  or be used to replace placeholders of a logical form template with KB elements \citep{DBLP:conf/acl/LuoETPG0MDSLZL24,DBLP:journals/eswa/LiLL24,DBLP:conf/coling/FengH25}.
  Nevertheless, the methods that retrieve the information from KBs do not guarantee \emph{grammatical validity} of generated logical forms, and may sacrifice \emph{efficiency} for the sake of accuracy.
  A grammatically invalid logical form cannot produce a denotation, even if it captures most of the meaning of the input utterance.
  In addition, inference efficiency is crucial for both production-level deployment and learning strategies, such as weakly-supervised learning \citep{DBLP:conf/emnlp/KrishnamurthyDG17,DBLP:conf/naacl/Dasigi0MZH19} or zero-shot learning \citep{DBLP:conf/emnlp/XuSCL20,DBLP:conf/acl/YinWSN22,DBLP:conf/acl/WuXLHLCYW023}, that involve inference during training.
  Therefore, an efficient constrained decoding method that ensures grammatical validity is desirable.
\end{revision}

In this work, we propose a grammar augmented with \emph{candidate expressions} for semantic parsing on a large KB with a seq2seq PLM.
Our grammar combines previous constrained decoding methods that construct compositional structures \citep{DBLP:conf/acl/YinN17,DBLP:conf/emnlp/KrishnamurthyDG17} and generate KB elements \citep{DBLP:conf/iclr/CaoI0P21,DBLP:conf/emnlp/ShuYLKMQL22}, which correspond to candidate expressions.
The two different methods are seamlessly unified into our grammar, which formulates constrained decoding as the problem of restricting actions for a given \emph{intermediate representation}.
We also introduce two special rules, sub-type inference and union types, and a mask caching algorithm.
In particular, sub-type inference and the mask caching algorithm increases the speed of the constrained decoding method for our grammar, so the method has small overhead during decoding.


We experiment on \kqapro \citep{DBLP:conf/acl/CaoSPNX0LHZ22} and \overnight \citep{DBLP:conf/acl/WangBL15},
where \kqapro is a benchmark for large-scale KBQA and \overnight is a benchmark for KBQA on multiple domains.
Our semantic parser is based on \bartmodel \citep{DBLP:conf/acl/LewisLGGMLSZ20}, and is trained with strong supervision from gold action sequences or with weak supervision from gold denotations.
In experiments, the constraints by candidate expressions increased accuracy of our semantic parser,
and the semantic parser had a fast decoding speed with sub-type inference and mask caching.

We list the contributions of our work as follows:
\begin{enumerate}[leftmargin=*, itemsep=0pt]
\item
  We propose a grammar integrated with types and candidate expressions for semantic parsing with a seq2seq PLM.
  Type constraints guide a semantic parser to construct compositional structures, where we introduce two special rules: sub-type inference and union types \Crefp{sec:grammars-with-types}.
  Candidate expressions guide a semantic parser to generate various categories of KB elements efficiently with multiple trie data structures \Crefp{sec:candidate-expressions}.
\item
  We devise a mask caching algorithm that increases the speed of constrained decoding with type constraints for a seq2seq PLM \Crefp{sec:mask-caching}.
\item
  Our semantic parser achieved state-of-the-art accuracies on two benchmarks: \kqapro which has a large KB and \overnight which has multiple KBs for eight domains \Crefp{sec:results-on-strongsup,sec:results-on-weaksup}.
  In addition, our constrained decoding method had small time cost \Crefp{sec:results-on-decoding-speed}.
\end{enumerate}

\section{Background and Related Work}

\subsection{Background} \label{sec:background}

We follow \citetseq{DBLP:conf/acl/YinN17,DBLP:conf/emnlp/KrishnamurthyDG17} whose grammars define actions as production rules that build well-typed formal representations such as abstract syntax trees or logical forms.
The grammars can be designed for complex syntax, and be applied for various logical form languages \citep{DBLP:conf/emnlp/YinN18,DBLP:conf/acl/NeubigZYH18,DBLP:conf/acl/GuoZGXLLZ19,DBLP:conf/naacl/Dasigi0MZH19,DBLP:conf/acl/WangSLPR20,DBLP:conf/acl/Gupta0020,DBLP:conf/acl/CaoC0ZZ020,DBLP:conf/acl/ChenLYLLJ21}.
However, the previous semantic parsers with the grammars depend on custom decoders based on long short-term memories (LSTMs) \citep{DBLP:journals/neco/HochreiterS97}, and their constraints for KBs are specialized for a decoder that generates a whole KB element by taking a single action \citep{DBLP:conf/emnlp/KrishnamurthyDG17,DBLP:conf/acl/ChenLYLLJ21}.
In contrast, our semantic parser is based on \bartmodel \citep{DBLP:conf/acl/LewisLGGMLSZ20}, which has a pre-trained decoder with a specific tokenizer, where a KB element is represented as a sequence of tokens that are generated by taking actions under the guidance of candidate expressions \Crefp{sec:candidate-expressions}.
In addition, we enhance the grammars with sub-type inference and union types \Crefp{sec:grammars-with-types}, and increase the speed of our constrained decoding method with mask caching for a seq2seq PLM \Crefp{sec:mask-caching}.
We also follow previous work \citep{DBLP:conf/iclr/CaoI0P21,DBLP:conf/emnlp/ShuYLKMQL22,DBLP:journals/mlc/DouGPWCLZ23} that uses trie data structures \citep{DBLP:books/daglib/0023376} to decode KB elements with seq2seq PLMs.
The previous constrained decoding methods have used one trie for entities \citep{DBLP:conf/iclr/CaoI0P21}, two distinct tries for unary and binary predicates \citep{DBLP:conf/emnlp/ShuYLKMQL22}, or a distinct trie for table names and column names for each database \citep{DBLP:journals/mlc/DouGPWCLZ23}; therefore, the methods use one or two tries to decode a logical form.
However, the previous methods lack constraints for compositional structures \citep{DBLP:conf/iclr/CaoI0P21,DBLP:journals/mlc/DouGPWCLZ23} or define the constraints as conditional statements that are hard-coded for specific logical form languages \citep{DBLP:conf/emnlp/ShuYLKMQL22}.
Therefore, applying the previous methods to semantic parsing is difficult when a logical form language involves complex compositional structures and various categories of KB elements that are attached to the compositional structures.
In contrast, our grammar is defined as a set of node classes that specify compositional structures by types \Crefp{sec:grammars-with-types} and KB elements by candidate expressions \Crefp{sec:candidate-expressions}.
As a result, our constrained decoding method uses a distinct trie for each KB element category in \kqapro, and uses a distinct set of tries for each KB of a domain in \overnight; therefore, our method uses various tries to decode a logical form.

\subsection{Related Work: Constrained Decoding} \label{sec:related-work:constrained-decoding}
Constrained decoding methods that are combined with parsing algorithms have been developed for semantic parsers based on seq2seq PLMs or LLMs. 
\citet{DBLP:conf/acl/Wu0XH0ZCYC20} used the LR(1) algorithm \citep{DBLP:journals/iandc/Knuth65}, \citet{DBLP:conf/emnlp/ShinLTCRPPKED21} used the Earley's algorithm \citep{DBLP:journals/cacm/Earley70}, \citet{DBLP:conf/emnlp/ScholakSB21} used an incremental parsing algorithm \citep{attoparsec}, and \citet{DBLP:conf/iclr/PoesiaP00SMG22} used the LL(*) algorithm \citep{DBLP:conf/pldi/ParrF11}.
The previous work shows that the parsing algorithms are effective for semantic parsers to increase accuracies.

However, the parsing algorithms, which are based on CFGs, are inefficient for constraints on large KBs, as many production rules are needed to define the constraints in CFGs.
The LR(1) algorithm \citep{DBLP:journals/iandc/Knuth65} requires a parsing table that consists of states, where a state has pairs of a dotted production rules and lookaheads.
The states with different lookaheads are not identical, although the states have the same list of production rules, so the size of the LR(1) parsing table is large when many production rules are incorporated into the table.
The Earley's algorithm \citep{DBLP:journals/cacm/Earley70} tracks several states during decoding, where a state has a dotted production rule and a position in decoded tokens.
The algorithm updates multiple states for each time step, so the decoding time is long when the states are derived from many production rules.
The LL(*) algorithm \citep{DBLP:conf/pldi/ParrF11} represents an arbitrary number of lookaheads as a regular expression.
The regular expression is defined from production rules that have the same left-hand side, so the size of the regular expression is large when the production rules are many.
In contrast, our grammar incorporates candidate expressions that define constraints on a large KB, and the constraints are efficiently implemented with trie data structures \Crefp{sec:candidate-expressions}.

In addition, the publicly available implementations of the previous constrained decoding methods iterate on tokens in vocabularies for each time step to check whether a token is valid as a next token, and the iteration process with CPUs is the bottleneck of parallel computing with GPUs.
To reduce the number of iterations, \citetseq{DBLP:conf/emnlp/ShinLTCRPPKED21,DBLP:conf/emnlp/ScholakSB21} iterate on only \topk tokens with respect to decoders' probabilities.
However, considering only \topk tokens raises a trade-off: (1) if the number \numk is large, the decoding time is increased; (2) otherwise, valid tokens with low probabilities are ignored, although these tokens are important for the initial search steps of weakly-supervised learning in which a model that performs search is not sufficiently trained.
Other constrained decoding methods \citep{DBLP:conf/acl/YinN17,DBLP:conf/emnlp/KrishnamurthyDG17,DBLP:conf/emnlp/YinN18} that predict production rules as actions with custom LSTM decoders \citep{DBLP:journals/neco/HochreiterS97} also iterate on valid actions for each time step.
In contrast, our constrained decoding method does not repeat on actions that are valid with respect to type constraints with a CPU, as the method caches a mask vector that specifies valid actions for each type, and as the mask vector is used to update a mask tensor with a GPU \Crefp{sec:mask-caching}.

Finally, the previous constrained decoding methods treat logical forms as sequences of tokens, then have long decoding time due to long output sequences, as all symbols in logical forms are tokenized \citep{DBLP:conf/acl/Wu0XH0ZCYC20,DBLP:conf/emnlp/ShinLTCRPPKED21,DBLP:conf/emnlp/ScholakSB21,DBLP:conf/iclr/PoesiaP00SMG22}.
\revisedtwo{
  However, symbols---such as functions, operators or named constants---are frequently used in logical forms, as the symbols are essential to synthesize logical forms.
  Therefore, our method sequentially predicts production rules themselves that generate expressions that consist of symbols, to reduce the lengths of output sequences and decoding time \citep{DBLP:conf/acl/YinN17,DBLP:conf/emnlp/KrishnamurthyDG17}.
}
In addition, the sub-type inference rule further reduces the lengths of action sequences \Crefp{sec:grammars-with-types}.

Bottom-up parsing has been applied to neural semantic parsers, and it uses constrained decoding.
\citetseq{DBLP:conf/naacl/RubinB21} define production rules for relational algebra \citep{codd1970relational},
and apply the production rules to compose logical forms in a bottom-up manner.
In contrast, our constrained decoding method applies production rules in a top-down manner.
Other bottom-up semantic parsers execute sub-logical forms during inference, and define constraints on the execution results \citep{DBLP:conf/acl/LiangBLFL17,DBLP:conf/nips/LiangNBLL18,DBLP:conf/acl/YinNYR20,DBLP:conf/coling/Gu022,DBLP:conf/acl/Gu0023}.
This approach can filter out meaningless logical forms that cannot be captured by type constraints.
Similarly, top-down semantic parsers also have benefited from constraints on the results of executing sub-logical forms \citep{wang2018robust,DBLP:conf/acl/ChenLYLLJ21}.
However, executing sub-logical forms during inference, in either bottom-up or top-down manner, entails substantial computational cost.

Weakly-supervised learning has been applied to traditional semantic parsers \citep{DBLP:conf/acl/LiangJK11,DBLP:conf/emnlp/BerantCFL13,DBLP:conf/acl/BerantL14,DBLP:conf/emnlp/ZhangPL17} and neural semantic parsers \citep{DBLP:conf/emnlp/KrishnamurthyDG17,DBLP:conf/acl/GuuPLL17,DBLP:conf/acl/LiangBLFL17,DBLP:conf/acl/BerantGGLN18,DBLP:conf/emnlp/MisraC0Y18,DBLP:conf/naacl/Dasigi0MZH19,DBLP:conf/nips/LiangNBLL18,DBLP:conf/emnlp/GuoLLZ21,DBLP:conf/naacl/WolfsonDB22}, in which constrained decoding is important for search steps.
However, weakly-supervised semantic parsing is less addressed for seq2seq PLMs than for custom LSTMs,
and existing work \citep{DBLP:conf/naacl/WolfsonDB22} does not use the parameters of a semantic parser during search steps.
In contrast, we follow \citetseq{DBLP:conf/naacl/Dasigi0MZH19} whose search steps use parameters that are the output of maximization steps, and we apply the learning process to our semantic parser based on \bartmodel, which is a seq2seq PLM \Crefp{sec:semantic-parsing}.
As we increase the speed of constrained decoding with sub-type inference \Crefp{sec:grammars-with-types} and mask caching \Crefp{sec:mask-caching}, the time cost of search steps is greatly reduced.
Our constrained decoding method would also be efficient for other learning strategies, such as zero-shot learning, that involve search steps during training \citep{DBLP:conf/emnlp/XuSCL20,DBLP:conf/acl/YinWSN22,DBLP:conf/acl/WuXLHLCYW023}.

\begin{revision}
  \subsection{Related Work: Question Answering over Knowledge Graphs} \label{sec:related-work:qa-over-kgs}
  A knowledge graph \citep{DBLP:conf/semweb/AuerBKLCI07,DBLP:conf/sigmod/BollackerEPST08,DBLP:journals/cacm/VrandecicK14}, which consists of entities and relations between them, is a widely used type of KB, and thus KBQA systems that use knowledge graphs have been actively developed.
  To derive an answer from a knowledge graph, KBQA methods typically depend on semantic parsing or information retrieval.
  The initial semantic parsing approach uses features computed from questions and sub-logical forms to gradually construct logical forms \citep{DBLP:conf/acl/CaiY13,DBLP:conf/emnlp/BerantCFL13,DBLP:conf/acl/BerantL14,DBLP:conf/acl/YihCHG15}.
  In contrast, the initial information retrieval approach extracts sub-graphs that include entities as candidate answers for a question, and computes similarities between the question and the sub-graphs to identify the top-ranked entities \citep{DBLP:conf/emnlp/BordesCW14,DBLP:conf/acl/DongWZX15}; therefore, this approach directly retrieves entities as answers.
  Since PLMs \citep{DBLP:conf/acl/LewisLGGMLSZ20,DBLP:journals/jmlr/RaffelSRLNMZLL20} and LLMs \citep{DBLP:conf/nips/BrownMRSKDNSSAA20,DBLP:journals/corr/abs-2107-03374} have been introduced, the two approaches have greatly evolved.

  To infer a logical form for a knowledge graph, the semantic parsing approach retrieves proper information from the knowledge graph.
  In particular, semantic parsers based on PLMs or LLMs take the information, then generate logical forms.
  The information can be KB elements, such as entities or relations \citep{DBLP:conf/emnlp/ShuYLKMQL22,DBLP:conf/emnlp/LiuYMRXZ22,DBLP:conf/aaai/Xu0ZLZ0000C25}, sub-graphs \citep{DBLP:conf/aaai/Xu0ZLZ0000C25}, paths from topic entities \citep{DBLP:conf/acl/YihCHG15} which exist in questions \citep{DBLP:conf/aaai/00010BZ0025}, or exemplary logical forms synthesized from the paths \citep{DBLP:conf/emnlp/ShuYLKMQL22,DBLP:conf/acl/YeYHZX22}.
  %
  %
  Alternatively, PLMs or LLMs generate logical form templates, whose placeholders are later replaced with KB elements that are retrieved from KBs \citep{DBLP:conf/acl/LuoETPG0MDSLZL24,DBLP:journals/eswa/LiLL24,DBLP:conf/coling/FengH25}.
  In addition, the retrieved information and logical forms can be interactively updated during inference \citep{DBLP:conf/coling/Gu022,DBLP:conf/acl/Gu0023,DBLP:conf/acl/FangZG24}.
  However, the information retrieved from knowledge graphs does not guarantee the grammatical validity of logical forms, so constrained decoding has been employed for semantic parsing.
  Previous work proposed constrained decoding methods that use tries \citep{DBLP:conf/emnlp/ShuYLKMQL22} or execution results \citep{DBLP:conf/coling/Gu022,DBLP:conf/acl/Gu0023}, but the former is unsuitable for complex syntax \Crefp{sec:background}, and the latter is time-consuming \Crefp{sec:related-work:constrained-decoding}.
  In contrast, our constrained decoding method supports complex syntax and is efficient.

  The information retrieval approach shares similar ideas, such as retrieving sub-graphs, with the semantic parsing approach, although the two approaches use different mechanisms for answer reasoning.
  To retrieve entities as answers, the information retrieval approach first retrieves sub-graphs that include entities or reasoning paths that lead to entities, where the entities are candidate answers, then the approach ranks the entities by using the retrieved information.
  The retrieval modules can be trained either independently \citep{DBLP:conf/acl/ZhangZY000C22,DBLP:journals/corr/abs-2410-13080} or jointly with the ranking modules \citep{DBLP:conf/emnlp/SunBC19,DBLP:conf/nips/ChoiLCK23,DBLP:conf/iclr/JiangZ0W23}.
  Recent work uses LLMs to retrieve reasoning paths and to derive answers from the paths \citep{DBLP:conf/iclr/LuoLHP24}, and these processes can be repeated until the LLMs find final answers \citep{DBLP:conf/iclr/SunXTW0GNSG24,DBLP:conf/acl/SuiHLHWH25,DBLP:conf/aaai/Ma0CSWPTSLZC25,DBLP:conf/acl/JiangZZS0ZW25,DBLP:conf/iclr/MaXJLQYMG25,DBLP:conf/www/LiuZLYPY25}.
  The information retrieval approach is effective for finding one or a few entities as an answer.
  However, the approach is inappropriate when an answer consists of many entities or when complex operations, such as aggregation, are needed.
  %
  %
\end{revision}

\section{Semantic Parsing} \label{sec:semantic-parsing}

\begin{figure}
  \begin{center}
    \input{\figurepathoverview.tex\inputfilesuffix}
    \caption{
      Semantic parsing on an example of \kqapro.
      \(\kb\) is a KB, \(\uttersc\) is an utterance, \(\actseq\) is an action sequence,
      \(\lfir{\actseq}\) is an intermediate representation built by \(\actseq\),
      \(\lflf{\actseq}\) is a logical form which corresponds to \(\lfir{\actseq}\),
      and \(\denot\) is the denotation when \(\lflf{\actseq}\) is evaluated on \(\kb\).
    }
    \label{fig:sp-example}
  \end{center}
\end{figure}

We first formally define a semantic parser as a function \(f_\theta: \UtterSet \rightarrow \ActSeqSet\) that maps a natural language utterance \(\uttersc \in \UtterSet\) to an action sequence \(\actseq \in \ActSeqSet = \ActSeqSetExpr\) that builds an intermediate representation \(\lfir{\actseq}\) that corresponds to a logical form \(\lflf{\actseq}\), which is evaluated on a KB \(\kb\) to produce a denotation \(\denot\) \Crefp{fig:sp-example}.
As a seq2seq model with a probability distribution over actions at each time step, we formulate a semantic parser as:
\begin{align}
  f_\theta(\uttersc) = \arg\max_{\actseq \in \ActSeqSet} p_\theta(\actseq \mid \uttersc)
  = \arg\max_{\actseq \in \ActSeqSet} \prod_{i=1}^{\setsize{\actseq}} p_{\theta}(\actsci \mid \actseqimo, \uttersc)
\end{align}
where \(\theta\) is the set of parameters of the model.
In practice, a semantic parser finds a sub-optimal solution by greedy search or beam search within a limited number of operations.

The semantic parser can be trained with strong supervision from gold action sequences, which are converted from gold logical forms.
When a training set \(\SplitBase\) has a gold action sequence \(\actseq\) for each example, the parameters \(\theta\) are optimized by the maximum likelihood objective \(\jml(\theta \scbar \SplitBase)\):
\begin{align}
  \jml(\theta \scbar \SplitBase) = \sum_{(\uttersc,\actseq) \in \SplitBase} \log p_\theta(\actseq \mid \uttersc) \label{eq:loss}
  .
\end{align}
However, annotating gold logical forms requires expertise, so the cost to construct the training set is expensive.

\begin{algorithm}[t]
  \begin{center}
    \input{\algorithmpathweaksup-learning.tex\inputfilesuffix}
    \newcommand{\weaksupdataset}{\{(x_j, y_j, \ActSeqSetWS_j)\}_{j}^{\setsize{\SplitBaseWS}}}
    \caption{
      Process of weakly-supervised learning.
      \(\theta_0\) is the initial set of parameters,
      \(\SplitBaseWS\) is a training set,
      \(\SplitVal\) is a validation set,
      where an example of the datasets has an utterance \(\uttersc\) and a gold denotation \(\lfgd\).
      \textsc{Search} finds action sequences \(\ActSeqSetWS\) for a given utterance \(\uttersc\), then constructs \(\SplitSearch\).
      \textsc{Maximize} performs gradient descent with \(\weaksupdataset\), which are merged from \(\SplitBaseWS\) and \(\SplitSearch\), for several epochs.
    }
    \label{alg:weaksup-learning}
  \end{center}
\end{algorithm}

Instead, the semantic parser can be trained with weak supervision from gold denotations, which are less expensive to annotate than gold logical forms.
When a training set \(\SplitBaseWS\) has a gold denotation \(\lfgd\) for each example, the parameters \(\theta\) are optimized by the maximum marginal likelihood objective \(\jmml(\theta \scbar \SplitBaseWS)\) \citep{DBLP:conf/acl/LiangJK11} \refappendixp{sec:derivation-of-grad}:
\begin{align}
  \jmml(\theta \scbar \SplitBaseWS)
  & = \sum_{(\uttersc, \lfgd) \in \SplitBaseWS} \jmml(\theta \scbar \uttersc, \lfgd)
    \label{eq:weaksup-loss} \\
  \jmml(\theta \scbar \uttersc, \lfgd)
  & = \log p_\theta(\lfgd \mid \uttersc)
    = \log \sum_{\actseq \in \ActSeqSet} p(\lfgd \mid \actseq) p_\theta(\actseq \mid \uttersc)
    \label{eq:weaksup-loss-example} \\
  \nabla_\theta \jmml(\theta \scbar \uttersc, \lfgd)
  & = \sum_{\actseq \in \ActSeqSet} p_\theta(\actseq \mid \uttersc, \lfgd) \nabla_\theta \log p_\theta(\actseq \mid \uttersc)
    = \mathop{\mathbb{E}}_{p_\theta(\actseq \mid \uttersc, \lfgd)} \nabla_\theta \log p_\theta(\actseq \mid \uttersc)
    \label{eq:grad-weaksup-loss-example}
  .
\end{align}
where
\begin{align}
  p_\theta(\actseq \mid \uttersc, \lfgd) &= \frac{p(\lfgd \mid \actseq) p_\theta(\actseq \mid \uttersc)}{\sum_{\actseq' \in \ActSeqSet} p(\lfgd \mid \actseq') p_\theta(\actseq' \mid \uttersc)} \\
  p(\lfgd \mid \actseq) &= \begin{cases}
    1, &\textrm{if \(\denot = \lfgd\)} \\
    0, &\textrm{otherwise} \end{cases} \label{eq:lf-score} .
\end{align}
%
The exact computation of \refeq{eq:grad-weaksup-loss-example} is intractable due to the large size of \(\ActSeqSet\),
which is the set of all action sequences,
so the learning process repeats a search step and a maximization step \citep{DBLP:conf/naacl/Dasigi0MZH19}.
In the search step, a search algorithm such as beam search finds action sequences \(\ActSeqSetWS \subset \ActSeqSet\) for a given utterance \(\uttersc\), where an action sequence \(\actseq \in \ActSeqSetWS\) has a high value of \(p_\theta(\actseq \mid \uttersc)\). 
In the maximization step, the parameters \(\theta\) are updated by gradient descent with approximated computation of \refeq{eq:grad-weaksup-loss-example}, where \(\ActSeqSetWS\) replaces \(\ActSeqSet\), to maximize \(\jmml(\theta \scbar \SplitBaseWS)\)
\Crefp{alg:weaksup-learning}.

However, to adapt a seq2seq PLM to our semantic parsing framework, we must reduce the discrepancy in formats between the actions and the natural language tokens that are predicted by the seq2seq PLM.
Therefore, we divide \(\ActSet\), which is the set of actions, into two subsets:
\(\ActSetCom\) which contains actions that build \textbf{com}positional structures or atomic units, and \(\ActSetNlt\) which contains actions that generate \textbf{n}atural \textbf{l}anguage \textbf{t}okens \citep{DBLP:conf/acl/YinN17}.
An action in \(\ActSetNlt\) constructs a node \(\lfexpr{\nlt{ *}}\) in which \(\lfexpr{*}\) is a natural language token.
Then, (1) the embedding of an action in \(\ActSetCom\) is learned from scratch and (2) the embedding of an action in \(\ActSetNlt\) is fine-tuned from the pre-trained embedding of the corresponding token.

\section{Grammars with Types} \label{sec:grammars-with-types}

\begin{table}[t]
  \begin{center}
    \lffontsizecaption
    \caption{
      Subset of a grammar definition for \kqapro.
      Each row specifies the properties of a node class.
      About logical form templates,
      \(\lfexpr{@i}\) means the logical form of a child node for an index \(\lfexpr{i}\),
      \(\lfexpr{@*}\) means the logical forms of all child nodes,
      and \(\lfexpr{\#(expr)}\) means that the logical form is the result of the evaluation of \(\lfexpr{(expr)}\).
    }

    \tablefontsizetwo
    \lffontsizetwo
    \input{\tablepathgrammar.tex\inputfilesuffix}

    \label{tab:grammar}
  \end{center}
\end{table}

\begin{table}[t]
  \begin{center}
    \caption{Example of building an intermediate representation by taking actions for \kqapro. 
      For each step, an action expands the leftmost non-terminal, written in bold, into a logical form expression, underlined in the next step.}

    \tablefontsizefour
    \lffontsizethree
    \input{\tablepathactions-to-ir.tex\inputfilesuffix}

    \label{tab:parsing-example}
  \end{center}
\end{table}

An action \(\actsc\) is a production rule that is applied to the leftmost non-terminal \(\lmnt(\lfir{\actseqic})\) in an intermediate representation \(\lfir{\actseqic}\) built from a past action sequence \(\actseqic\).
The action \(\actsc = \actfromnc(\ncsc)\) corresponds to a node class \(\ncsc\) that is defined by a grammar that specifies a return type and parameter types for \(\ncsc\) \Crefp{tab:grammar}.
The action \(\actsc\) expands the return type of \(\ncsc\) to an expression that is composed of the name of \(\ncsc\) and the parameter types of \(\ncsc\):
\begin{align*}
  \lfexpr{\lfnt{\textit{return-type}}} &\rightarrow \lfexpr{\lrb \textit{class-name} } \lfexpr{\lfnt{\textit{param-type-0}} \lfnt{\textit{param-type-1}} ...\rrb} \\
  \intertext{or \(\actsc\) expands to the name of \(\ncsc\) when \(\ncsc\) does not have any parameter type:}
  \lfexpr{\lfnt{\textit{return-type}}} &\rightarrow \lfexpr{\textit{class-name}}
\end{align*}
where the types become non-terminals.
We express an intermediate representation as an \nohyphens{S-expression} which consists of symbols and parentheses \citep{DBLP:conf/hopl/McCarthy78}.
The S-expression is a tree structure, in which the first symbol in a pair of parentheses is the parent node and the remaining symbols or sub-expressions are child nodes.

Under type constraints, an action \(\actsc\) can be applied to the leftmost non-terminal \(\lmnt(\lfir{\actseqic})\) when \(\actsc\)'s left-hand side \(\lhs(\actsc)\) and \(\lmnt(\lfir{\actseqic})\) have the same type or compatible types \citep{DBLP:conf/acl/YinN17,DBLP:conf/emnlp/KrishnamurthyDG17} \Crefp{fig:candidate-actions}.
For type compatibility, we introduce two new special rules, \emph{sub-type inference} and \emph{union types}, and we also adapt two existing special rules, \emph{optional types} and \emph{repeatable types} \citep{DBLP:conf/acl/YinN17}:

\paragraphhead{Sub-type inference} allows an action \(\actsc\) to be applied to the leftmost non-terminal \(\lmnt(\lfir{\actseqic})\) when the left-hand side \(\lhs(\actsc)\) has a sub-type of \(\lmnt(\lfir{\actseqic})\).
For example, \mbox{\(\actsc = \actfromnc(\lfexpr{query-rel-qualifier})\)} has \(\lfexpr{\lfnt{result-rel-q-value}}\) as \(\lhs(\actsc)\), then \(\actsc\) can be applied to \mbox{\(\lmnt(\lfir{\actseqic}) = \lfexpr{\lfnt{result}}\)}, as \(\lfexpr{\lftyp{result-rel-q-value}}\) is a sub-type of \(\lfexpr{\lftyp{result}}\).
Sub-type inference does not change the size of the search space, which is the set of all complete action sequences.
However, sub-type inference shortens the length of an action sequence by skipping production rules that convert non-terminals of super-types to non-terminals of sub-types; e.g. \(\lfexpr{\lfnt{result}} \rightarrow \lfexpr{\lfnt{result-rel-q-value}}\).
%
Therefore, sub-type inference increases decoding speed, as decoding time is proportional to the length of an action sequence \Crefp{sec:results-on-decoding-speed}.

\paragraphhead{Union types} allow the left-hand side \(\lhs(\actsc)\) of an action \(\actsc\) to have multiple types, then \(\actsc\) can be applied to the leftmost non-terminal \(\lmnt(\lfir{\actseqic})\) when the type of \(\lmnt(\lfir{\actseqic})\) is same as or compatible with a type that belongs to \(\lhs(\actsc)\).
We assign a union type to \(\lhs(\actsc)\) for an action \(\actsc = \actfromnc(\lfexpr{\nlt{ *}}) \in \ActSetNlt\).
For example, \mbox{\(\actsc = \actfromnc(\lfexpr{\nltwb{country}})\)} has \(\lfexpr{\lfnt{kp-entry kp-relation ...}}\) as \(\lhs(\actsc)\), whose type is \(\lfexpr{(\uniontyp \lftyp{kp-entry} \lftyp{kp-relation} ...)}\), then \(\actsc\) can be applied to \(\lmnt(\lfir{\actseqic})\) such as \(\lfexpr{\lfnt{kp-entry}}\) or \(\lfexpr{\lfnt{kp-relation}}\),
but it cannot be applied to \(\lfexpr{\lfnt{vp-quantity}}\) which requires another action \(a'\),
such as \mbox{\(\actfromnc(\lfexpr{\nltwb{7}})\)}, whose left-hand side \(\lhs(a')\) is \(\lfexpr{\lfnt{vp-quantity ...}}\).
Therefore, union types are effective for distinguishing valid actions from \(\ActSetNlt\) with respect to \(\lmnt(\lfir{\actseqic})\).

\paragraphhead{Optional types} allow the leftmost non-terminal \(\lmnt(\lfir{\actseqic})\) that has \(\lfexpr{\lfntoptional}\) as a suffix to be skipped when the special action \(\reduceact\) is taken.
The special non-terminal that has \(\lfexpr{\lfntoptional}\) is derived from a parameter type declared with the \(\lfexpr{\&optional}\) keyword.
For example, a node class \(\lfexpr{constant-date}\) that is defined in \overnight has \(\lfexpr{\lftyp{year}}\), \(\lfexpr{\lftyp{month}}\) and \(\lfexpr{\lftyp{day}}\) as parameter types where \(\lfexpr{\lftyp{month}}\) and \(\lfexpr{\lftyp{day}}\) are optional, then the optional types become non-terminals \(\lfexpr{\lfnt{month}\lfntoptional}\) and \(\lfexpr{\lfnt{day}\lfntoptional}\); once \(\lfexpr{\lfnt{month}\lfntoptional}\) is skipped by taking \(\reduceact\), the remaining non-terminal \(\lfexpr{\lfnt{day}\lfntoptional}\) is also skipped, and the expression of \(\lfexpr{constant-date}\) becomes complete.

\paragraphhead{Repeatable types} allow the leftmost non-terminal \(\lmnt(\lfir{\actseqic})\) that has \(\lfexpr{\lfntrepeat}\) as a suffix to be repeated until the special action \(\reduceact\) is taken.
The special non-terminal that has \(\lfexpr{\lfntrepeat}\) is derived from a parameter type declared with the \(\lfexpr{\&rest}\) keyword.
For example, a node class \(\lfexpr{keyword-relation}\) has \(\lfexpr{\lftyp{kp-relation}}\) as the second parameter type, which is declared with the \(\lfexpr{\&rest}\) keyword, then the type becomes a non-terminal \(\lfexpr{\lfnt{kp-relation}\lfntrepeat}\), which is repeated as \(\lmnt(\lfir{\actseqic})\) until \(\reduceact\) is taken.
\paragraphheadvspace

The parsing procedure is to sequentially take actions, which expand the leftmost non-terminals to sub-expressions, until no non-terminal remains \Crefp{tab:parsing-example}.
When the past action sequence is \(\actseqtmo\), an action \(\actsct\) replaces the leftmost non-terminal \(\lmnt(\lfir{\actseqtmo})\) with the right-hand side of \(\actsct\), then the intermediate representation is updated as \(\lfir{\actseqte}\).
For each step during parsing, a semantic parser should distinguish which actions are valid for the current intermediate representation.
Therefore, a semantic parser needs a function \mbox{\(\ActBase: \TypeForActFunc\)} that maps an intermediate representation \mbox{\(\lfir{\actseqic} \in \IRSet\)} to a set of valid actions \mbox{\(\ActBaseFn{\lfir{\actseqic}} \subset \ActSet\)}.

We define \(\ActTypeFn{\lfir{\actseqic}}\) as the set of all valid actions with respect to types.
For an action \mbox{\(\actsc \in \ActTypeFn{\lfir{\actseqic}}\)}, the leftmost non-terminal \(\lmnt(\lfir{\actseqic})\) and the left-hand side \(\lhs(\actsc)\) have compatible types.
Therefore, \(\ActType\) guides a semantic parser to gradually compose well-typed intermediate representations.
When parsing is finished, the final expression is a complete intermediate representation \(\lfir{\actseq}\), which can be converted to a logical form \(\lflf{\actseq}\), as each node has a corresponding logical form template (\revised{\Cref{tab:grammar,tab:ir-to-lf}; \Cref{sec:implementation} ``Logical form templates''}).
\revised{To facilitate understanding of our constrained decoding method \Crefp{sec:grammars-with-types,sec:candidate-expressions,sec:mask-caching}, we provide a working example \inrefappendix{sec:working-example}.}

\section{Candidate Expressions} \label{sec:candidate-expressions}

\begin{table}[t]
  \begin{center}
    \caption{
      Node classes for \kqapro, subsets of their candidate expressions, and the numbers of candidate expressions.
    }
    \tablefontsizetwo
    \lffontsizetwo
    \input{\tablepathkqapro-candidate-expressions.tex\inputfilesuffix}
    \label{tab:kqapro-candidate-expressions}
  \end{center}

  \begin{center}
    \anotherTableCaption{
      Node classes for \overnight and the numbers of their candidate expressions for each domain.
      We only consider KB elements that have natural language counterparts as candidate expressions,
      because many KB elements in \overnight are artificially created.
    }
    \tablefontsizetwo
    \lffontsizetwo
    \input{\tablepathovernight-candidate-expressions.tex\inputfilesuffix}
    \label{tab:overnight-candidate-expressions}
  \end{center}
\end{table}

\begin{figure}[t]
  \begin{center}

    \begingroup
    \lffontsizefour
    \tablefontsizesix
    \input{\figurepathcandidate-actions.tex\inputfilesuffix}
    \endgroup

    \newcommand{\redtxt}[1]{\textcolor{red}{#1}}
    \newcommand{\blutxt}[1]{\textcolor{blue}{#1}}
    \lffontsizecaption
    \caption{
      Example of two action sets \(\ActTypeFn{\lfir{\actseqic}}\) and \(\ActCandFn{\lfir{\actseqic}}\) for an intermediate representation \(\lfir{\actseqic}\) for \kqapro.
      \(\lfexpr{\boldim{\lfnt{kp-relation}\lfntrepeat}}\) is \(\lmnt(\lfir{\actseqic})\) which is the leftmost non-terminal in \(\lfir{\actseqic}\),
      \(\lfexpr{\emphsy{keyword-relation}}\) is \(\parent(\lmnt(\lfir{\actseqic}))\) which is the parent node of \(\lmnt(\lfir{\actseqic})\),
      and \(\lfexpr{\uwave{\nltwb{country}}}\) is \(\childseq(\parent(\lmnt(\lfir{\actseqic})))\) which is the sequence of child \(\lfexpr{\nlt{ *}}\) nodes of \(\parent(\lmnt(\lfir{\actseqic}))\) and whose length is currently one.
      \(\lfexpr{\blutxt{\lfnt{kp-relation ...} \(\rightarrow\) \nltwb{of}}}\) and \(\lfexpr{\blutxt{\lfnt{kp-relation ...} \(\rightarrow\) \nltwb{for}}}\) are actions that result in valid prefixes of candidate expressions with respect to a KB \(\kb\).
      \(\lfexpr{\redtxt{\lfnt{kp-relation ...} \(\rightarrow\) \nltwb{with}}}\) is an action that results in an invalid prefix of a candidate expression.
      \(\lfexpr{\lfnt{kp-relation ...}}\) is \(\lhs(\actsc)\) which is the left-hand side of an action \mbox{\(\actsc \in \lfexpr{\lfnt{kp-relation ...} \(\rightarrow\) \nltwb{of} | \nltwb{with} | \nltwb{for}}\)}.
    }
    \label{fig:candidate-actions}
  \end{center}
\end{figure}

Our grammar with types builds compositional structures of intermediate representations,
but the grammar is insufficient to synthesize valid KB elements.
A KB element is constructed by a node, such as \(\lfexpr{keyword-relation}\), and the node has a sequence of \(\lfexpr{\nlt{ *}}\) nodes as children. 
Unless the sequence of \(\lfexpr{\nlt{ *}}\) nodes becomes an existing KB element, a logical form that involves the sequence cannot produce a meaningful denotation.

We augment the grammar with candidate expressions to generate existing KB elements.
A candidate expression \(\cesc \in \CESetFn{\ncsc}\) for a node class \(\ncsc\) is a predefined instance of a KB element category that corresponds to \(\ncsc\), where the elements of \(\CESetFn{\ncsc}\) is determined by a KB \(\kb\).
For example, in \kqapro, the KB element category ``relation'', which corresponds to a node class \(\ncsc = \lfexpr{keyword-relation}\), has predefined instances, such as \(\lfexpr{\lfstr{country of citizenship}}\) and \(\lfexpr{\lfstr{country for sport}}\), as candidate expressions in \(\CESetFn{\ncsc}\) \Crefp{tab:kqapro-candidate-expressions}.
In \overnight, a set \(\CESetFn{\ncsc}\) of candidate expressions varies depending on a domain, as each domain has a different KB \(\kb\) \Crefp{tab:overnight-candidate-expressions}.
In addition, a set \(\CESetFn{\ncsc}\) of candidate expressions are shared with a node \(\nisc\) that is instantiated from the node class \(\ncsc\); therefore \(\CESetFn{\ncsc} = \CESetFn{\nisc}\).

We define \(\ActCandFn{\lfir{\actseqic}}\) as the set of valid actions with respect to candidate expressions.
\(\ActCandFn{\lfir{\actseqic}}\) depends on \(\parent(\lmnt(\lfir{\actseqic}))\) which is the parent node of the leftmost non-terminal \(\lmnt(\lfir{\actseqic})\) \Crefp{fig:candidate-actions}.
\revised{The parent node \(\parent(\lmnt(\lfir{\actseqic}))\) has \(\childseq(\parent(\lmnt(\lfir{\actseqic})))\) which is the sequence of child \(\lfexpr{\nlt{ *}}\) nodes and whose concatenation should be always a prefix of a candidate expression \(\cesc \in \CESet(\parent(\lmnt(\lfir{\actseqic})))\).}
Therefore, an action \mbox{\(\actsct \in \ActCandFn{\lfir{\actseqtmo}}\)} replaces the leftmost non-terminal \(\lmnt(\lfir{\actseqtmo})\) with an \(\lfexpr{\nlt{ *}}\) node as a child of \(\parent(\lmnt(\lfir{\actseqtmo}))\), then the concatenation of \(\childseq(\parent(\lmnt(\lfir{\actseqte})))\) becomes an extended prefix of a candidate expression \mbox{\(\cesc \in \CESet(\parent(\lmnt(\lfir{\actseqtmo})))\)}.

We implement \(\ActCand\) with trie data structures \citep{DBLP:books/daglib/0023376} that store candidate expressions which are split into natural language tokens.
For each node class \(\ncsc\), we convert its candidate expressions in \(\CESetFn{\ncsc}\) into token sequences, which are then added to the trie \(\trieFn{\ncsc}\).
The trie \(\trieFn{\ncsc}\) is shared with a node \(\nisc\) instantiated from the node class \(\ncsc\); therefore \(\trieFn{\ncsc} = \trieFn{\nisc}\).
A constructed trie \(\trieFn{\nisc}\) takes a token sequence as a prefix of a candidate expression \(\cesc \in \CESetFn{\nisc}\), then retrieves valid tokens that can extend the prefix.
Therefore, an action \(\actsc \in \ActCandFn{\lfir{\actseqic}}\) is represented as \(\lfexpr{\lfnt{...}} \rightarrow \lfexpr{\nlt{ *}}\), in which the token \(\lfexpr{*}\) is retrieved from the trie \(\trieFn{\parent(\lmnt(\lfir{\actseqic}))}\)
when given a token sequence that is extracted from \(\childseq(\parent(\lmnt(\lfir{\actseqic})))\).
Previous work has used one trie for entities \citep{DBLP:conf/iclr/CaoI0P21}, two distinct tries for predicates \citep{DBLP:conf/emnlp/ShuYLKMQL22}, or a distinct trie for each database \citep{DBLP:journals/mlc/DouGPWCLZ23}, whereas we use a distinct trie for each pair of a node class \(\ncsc\) and a KB \(\kb\), where \(\ncsc\) corresponds to a KB element category.

Finally, we introduce \(\ActHybr\) which is a hybrid function of \(\ActType\) and \(\ActCand\).
For an intermediate representation \(\lfir{\actseqic}\), \(\ActHybr\) returns a set of valid actions from \(\ActCandFn{\lfir{\actseqic}}\) when candidate expressions are defined for \(\parent(\lmnt(\lfir{\actseqic}))\), or from \(\ActTypeFn{\lfir{\actseqic}}\) otherwise:
\begin{align}
  \ActHybrFn{\lfir{\actseqic}} =
  \begin{cases}
    \ActCandFn{\lfir{\actseqic}} & \quad \textrm{if \,} \textsc{HasCandExpr}(\parent(\lmnt(\lfir{\actseqic}))) \\
    \ActTypeFn{\lfir{\actseqic}} & \quad \textrm{otherwise} .
  \end{cases} \label{eq:act-hybr}
\end{align}
Therefore, \(\ActHybr\) uses \(\ActType\) to construct compositional structures, and uses \(\ActCand\) to generate KB elements that are attached to the compositional structures.

\section{Mask Caching} \label{sec:mask-caching}

\begin{algorithm}[t!]
  \begin{center}
    \input{\algorithmpathcompute-mask-vector.tex\inputfilesuffix}
    \caption{
      Method to compute a mask vector for an action sequence.
      The input values are a past action sequence \(\actseqic\) and a cache memory \(\Cache\).
      The return value is a cached mask vector \(\Cache[\Type(\lmnt(\lfir{\actseqic}))]\).
    }
    \label{alg:compute-mask-vector}
  \end{center}
\end{algorithm}

\begin{algorithm}[t!]
  \begin{center}
    \input{\algorithmpathcompute-mask-tensor.tex\inputfilesuffix}
    \caption{
      Method to compute a mask tensor for a batch of action sequences.
      The input values are a batch \(\Batch\) of past action sequences and a cache memory \(\Cache\).
      The return value is a mask tensor \(\Tensor\).
    }
    \label{alg:compute-mask-tensor}
  \end{center}
\end{algorithm}

Our semantic parser searches for an action sequence \(\actseq\) when given an utterance \(\uttersc\),
where a search algorithm, such as greedy search or beam search, depends on a scoring function.
Without constrained decoding, a scoring function \(s\) is defined as:
\begin{align}
  s(\actsc \scbar \actseqtmo, \uttersc, \theta)
  = \log p_{\theta}(\actsc \mid \actseqtmo, \uttersc) + \log p_{\theta}(\actseqtmo \mid \uttersc)
  = \log p_{\theta}((\actseqtmo, \actsc) \mid \uttersc)
  \label{eq:score-func-base}
\end{align}
which assigns a priority to an action \(\actsc \in \ActNoneFn{\actseqtmo} = \ActSet\) as a candidate for the next action \(\actsct\).
An implementation of a search algorithm for a seq2seq PLM computes, in parallel with a GPU, a score vector that contains \(s(\actsc \scbar \actseqtmo, \uttersc, \theta)\) for every action \(\actsc \in \ActSet\).

For constrained decoding with a function \(\ActBase \in \SetOfActsWC\), we define a scoring function \(s_{\ActBase}\) that combines the scoring function \(s\) and a masking function \(m_{\ActBase}\):
\begin{align}
  s_{\ActBase}(\actsc \scbar \actseqtmo, \uttersc, \theta) &= s(\actsc \scbar \actseqtmo, \uttersc, \theta) + m_{\ActBase}(\actsc \scbar \actseqtmo) \\
  m_{\ActBase}(\actsc \scbar \actseqtmo) &=
  \begin{cases}
    0 & \quad \textrm{if \,} \actsc \in \ActBaseFn{\lfir{\actseqtmo}} \\
    - \infty & \quad \textrm{otherwise} .
  \end{cases} \label{eq:act-hybr}
\end{align}
In contrast to the scoring function \(s\), the masking function \(m_{\ActBase}\) involves an operation that checks whether an action \(\actsc\) exists in \(\ActBaseFn{\lfir{\actseqtmo}}\), and the operation cannot be efficiently executed in parallel with a GPU.
Therefore, computing a mask vector that contains \(m_{\ActBase}(\actsc \scbar \actseqtmo)\) for every action \(\actsc \in \ActSet\) costs \(O(\setsize{\ActSet})\) time as the size \(\setsize{\ActBaseFn{\lfir{\actseqtmo}}}\) ranges from 1 to \(\setsize{\ActSet}\).

We devise a mask caching algorithm that reduces the time cost to create a mask tensor during constrained decoding.
A grammar has a limited number of types, so we cache and retrieve a mask vector for \(\Type(\lmnt(\lfir{\actseqic}))\) when given a past action sequence \(\actseqic\) \Crefp{alg:compute-mask-vector}. To compute a mask tensor \(\Tensor\) for a batch \(\Batch\) of past action sequences, we initialize all element of \(\Tensor\) as \(-\infty\), then update a mask vector \(\Tensor_{i}\) with respect to a past action sequence \(\actseqic = \Batch_i\), where \(\Tensor_{i}\) is updated as a cached mask vector from \(\textsc{ComputeMaskVector}(\actseqic, \Cache)\) unless the condition \(\textsc{HasCandExpr}(\parent(\lmnt(\lfir{\actseqic})))\) is satisfied \Crefp{alg:compute-mask-tensor}.
In particular, updating \(\Tensor_{i}\) as \(\textsc{ComputeMaskVector}(\actseqic, \Cache)\) is efficiently processed by a GPU in constant time.


\newcommand{\fE}{E}             
\newcommand{\fF}{F}             

\begin{revision}
  \begingroup
  We analyze the time complexity of \(\textsc{ComputeMaskTensor}(\Batch, \Cache)\) \Crefp{alg:compute-mask-tensor}.
  To ease the analysis, we assume that the current time step is \(\tmostep\) and the batch size \(\setsize{\Batch}\) is 1, so \(\Batch = \left[\actseqtmo\right]\).
  We also use the following symbols:
  \(\fE\) is the prefix length \(\setsize{\childseq(\parent(\lmnt(\lfir{\actseqtmo})))}\) which is smaller than \(t\),
  and \(\fF\) is the action set size \(\setsize{\ActCandFn{\lfir{\actseqtmo}}}\) which is smaller than \(\setsize{\ActSet}\).
  The time complexity depends on the condition \(\textsc{HasCandExpr}(\parent(\lmnt(\lfir{\actseqtmo})))\).
  When the condition is true, the time complexity is \(O(\fE + \fF)\), where \(\fE\) is the time to search a trie and \(\fF\) is the time to update a mask tensor.
  Otherwise, the time complexity is \(O(1)\), as the mask caching algorithm retrieves a cached mask vector to update a mask tensor.

  We now present the time complexity of constrained decoding with \(\ActHybr\) between the time step \(\tmostep\) and the time step \(\tstep\).
  The complexity is overhead that is added to unconstrained decoding.
  For the time step \(\tmostep\), our method computes a mask tensor \(\textsc{ComputeMaskTensor}([\actseqtmo], \Cache)\), and updates an intermediate representation from \(\lfir{\actseqtmo}\) to \(\lfir{\actseqte}\) after an action \(\actsct\) is predicted by a seq2seq model.
  Therefore, we can derive the time complexity by adding the time to compute a mask tensor and the time to update an intermediate representation \CrefpContent{sec:implementation}{Intermediate representations}.
  The worst-case time complexity is \(O(\fF + \tstep)\) in which \(E\) is omitted because \(\fE < \tstep\).
  The average-case time complexity is \(O(\fE + \fF)\) or \(O(1)\) when \(\textsc{HasCandExpr}(\parent(\lmnt(\lfir{\actseqtmo})))\) is true or false respectively.
  In addition, we briefly address an issue of time complexity when implementing our method with the transformers library \citep{DBLP:conf/emnlp/WolfDSCDMCRLFDS20} \inrefappendix{sec:search-implementation}.

  \endgroup
\end{revision}

\section{Implementation Details and Experimental Setup} \label{sec:implementation}

We implement our semantic parser with the proposed grammar on two complementary KBQA benchmarks: \kqapro, which is a large-scale benchmark \citep{DBLP:conf/acl/CaoSPNX0LHZ22}, and \overnight, which is a multi-domain benchmark \citep{DBLP:conf/acl/WangBL15}.

\begin{revision}
  \paragraphheaddot{Benchmarks}
  The \kqapro benchmark evaluates complex reasoning abilities, such as multi-hop inference, over a dense subset of Wikidata \citep{DBLP:journals/cacm/VrandecicK14}.
  In particular, \kqapro addresses reasoning with qualifiers, which enable Wikidata to express n-ary facts.
  While other large-scale KBQA benchmarks support either complex reasoning \citep{DBLP:conf/naacl/TalmorB18,DBLP:conf/www/GuKVSLY021} or reasoning with n-ary facts \citep{DBLP:conf/cikm/JiaPRW21}, none supports both.
  As a result, \kopl \citep{DBLP:conf/acl/CaoSPNX0LHZ22}, the logical form language of \kqapro, has 27 functions, including those related to qualifiers, and the latent grammar of the language is complex (\Cref{fig:kqapro-type-hierarchy} and \Cref{tab:kqapro-grammar-part-1,tab:kqapro-grammar-part-2}).
  Therefore, applying constrained decoding to \kqapro is challenging.

  The \overnight benchmark focuses on domain-specific linguistic variability by addressing the linguistic gap between natural language and logical forms.
  As each domain in \overnight uses a different KB, the examples across the domains exhibit different linguistic phenomena.
  For example, in the Basketball domain, the utterance \textit{``Where did kobe bryant play in 2004?''} uses the word \textit{``Where''} to refer to the relation \(\lfexpr{relation:team}\) \Crefp{tab:overnight-example-sp-basketball-relation-entity}.
  To infer a correct logical form, a semantic parser should utilize a domain-specific KB.
  Therefore, the ability to customize constrained decoding to a given domain-specific KB is necessary.
\end{revision}

\paragraphheaddot{Data splits}
We use the standard \kqapro data splits: the training set \(\SplitTrain\), the validation set \(\SplitVal\) and the test set \(\SplitTest\); they contain 94,376, 11,797 and 11,797 examples respectively \citep{DBLP:conf/acl/CaoSPNX0LHZ22}.
Each example includes a question, a logical form written in \kopl \citep{DBLP:conf/acl/CaoSPNX0LHZ22} and an answer.
We map a question to an utterance \(\uttersc\), and an answer to a gold denotation \(\lfgd\).
We also augment an example in \(\SplitTrain\) with an action sequence \(\actseq\), which is converted from the \kopl logical form of the example. However, gold denotations in \(\SplitTest\) for \kqapro are not publicly available, so we evaluate our semantic parsers on \(\SplitTest\) only for main experiments \Crefp{tab:kqapro-strongsup-main-result,tab:kqapro-weaksup-main-result}.

\overnight has training and test data splits, and we extract 20\% of the training data split as a validation set \citep{DBLP:conf/acl/WangBL15}; therefore, the training set \(\SplitTrain\), the validation set \(\SplitVal\) and the test set \(\SplitTest\) contain 8,751, 2,191 and 2,740 examples respectively.
Each example includes an utterance and a logical form written in \lambdadcs \citep{DBLP:journals/corr/Liang13}.
We augment an example in \(\SplitTrain\) with an action sequence \(\actseq\) which is converted from the \lambdadcs logical form of the example, and with the denotation \(\denot\) as a gold denotation \(\lfgd\).
By default, we train a strongly-supervised model with \(\SplitTrain\) which consists of examples of all eight domains, but additionally train each model for a domain with examples that belong to the domain in \(\SplitTrain\) \Crefp{tab:overnight-strongsup-main-result}.

For weakly-supervised learning, we use a small subset of \(\SplitTrain\) for pre-training with strong supervision and use \(\SplitTrainWS\) for fine-tuning with weak supervision.
We use 0.1\% of \(\SplitTrain\) for \kqapro and 1\% of \(\SplitTrain\) for \overnight, as the subsets for pre-training \Crefp{tab:statistics-of-train-set}.
\(\SplitTrainWS\) is a set of examples that have only utterances and gold denotations in \(\SplitTrain\).

\paragraphheaddot{Models}
We develop our semantic parser with \bartmodel \citep{DBLP:conf/acl/LewisLGGMLSZ20}, which is a seq2seq PLM.
For fair comparison with previous work, we especially use the \bartbase model, with which previous semantic parsers were developed \citep{DBLP:conf/acl/CaoSPNX0LHZ22, DBLP:conf/emnlp/NieCSST0LZ22, DBLP:conf/aaai/NieS0DH00LZ23}.
We additionally use the \bartlarge model to address the effect of our constrained decoding method on a large PLM \Crefp{sec:results-on-large-plms}.

\paragraphheaddot{Grammars}
Our grammar defines the actions in \(\ActSet = \ActSetCom \cup \ActSetNlt\) \Crefp{sec:semantic-parsing}.
The size \(\setsize{\ActSetCom}\) is 53 for \kqapro and 36 for \overnight if sub-type inference is applied; otherwise, \(\setsize{\ActSetCom}\) is 76 for \kqapro and 48 for \overnight \refappendixp{sec:node-classes}.
We apply sub-type inference to our semantic parser by default, and additionally disable sub-type inference to address the effect of decoding speed \Crefp{sec:results-on-decoding-speed}.
The size \(\setsize{\ActSetNlt}\) is 50,260, which is the same as the number of non-special tokens of \bartmodel.

From the grammar, different \(\ActBase\) functions are derived: (1) \(\ActHybr\), (2) \(\ActType\), (3) \(\ActTypeWU\) which replaces different union types with the same type and (4) \(\ActNone\) which always returns \(\ActSet\), the set of all actions, without applying any constraint.
The action sets that are returned from the four functions have the following subset relations:
\begin{align}
  \ActHybrFn{\lfir{\actseqic}} \subset \ActTypeFn{\lfir{\actseqic}} \subset \ActTypeWUFn{\lfir{\actseqic}} \subset \ActNoneFn{\lfir{\actseqic}} = \ActSet . \label{eq:subset-relations}
\end{align}
We address the effect of the functions \(\ActFuncSet = \SetOfAllActs\) \inCref{sec:results-on-strongsup,sec:results-on-weaksup}.

\paragraphheaddot{Intermediate representations}
An intermediate representation \(\lfir{\actseqic}\) is stored in a linked list that consists of nodes and complete sub-expressions.
An action \(\actsc\) that is not \(\reduceact\) attaches a node to the linked list.
When the parent node \(\parent(\lmnt(\lfir{\actseqic}))\) has no more non-terminals as child nodes, or when the last action is \(\reduceact\),
\(\parent(\lmnt(\lfir{\actseqic}))\) and its children are reduced \citep{DBLP:conf/acl/ChengRSL17}.
The reduction operation pops \(\parent(\lmnt(\lfir{\actseqic}))\) and its children from the linked list, then again attaches them to the linked list as a complete sub-expression.
\revised{If a newly attached complete sub-expression replaces the last non-terminal of a parent node, reduction is again performed and can be repeated.
Therefore, the worst-case time complexity of updating an intermediate representation from \(\lfir{\actseqtmo}\) to \(\lfir{\actseqte}\) is \(O(\tstep)\) and the average-case time complexity is \(O(1)\).}
In addition, a linked list can be shared as a sub-linked list for other linked lists, so search algorithms do not need to the copy the previous intermediate representation \(\lfir{\actseqtmo}\) when multiple branches with different actions from \(\ActBaseFn{\lfir{\actseqtmo}}\) occur for the time step \(t\).

\begin{revision}
  \paragraphheaddot{Logical form templates}
  An intermediate representation \(\lfir{\actseq}\) can be converted to a logical form \(\lflf{\actseq}\) by using logical form templates \Crefp{tab:grammar,tab:kqapro-grammar-part-1,tab:kqapro-grammar-part-2,tab:grammar,tab:overnight-grammar-part-1,tab:overnight-grammar-part-2}.
  The conversion process is to apply the logical form template of each node to the intermediate representation of the node in a bottom-up manner \Crefp{tab:ir-to-lf},
  and the time complexity of the process is \mbox{\(O(\setsize{\actseq} + \setsize{\lflf{\actseq}})\)} \refappendixp{sec:lf-template}.
  The logical form templates can be designed for various formats, such as S-expressions, database queries, or executable program code, once the formats are specified with CFGs.
  In our application to \kqapro and \overnight, both \kopl and \lambdadcs logical forms are represented as S-expressions.
\end{revision}

\begin{table}[t]
  \begin{center}
    \caption{
      Statistics of the training set \(\SplitTrain\) and its subset that is used for pre-training a model in advance of fine-tuning the model with weak supervision.
    }
    \begin{subtable}[h]{\textwidth}
      \begin{center}
        \caption{
          Statistics for \kqapro.
          An example can belong to more than one category.
          The subset for pre-training is 0.1\% of \(\SplitTrain\).
        }
        \tablefontsizeone
        \input{\tablepathkqapro-statistics-of-train-set.tex\inputfilesuffix}
      \end{center}
    \end{subtable}

    \vspace{\subtablevspacelen}

    \begin{subtable}[h]{\textwidth}
      \begin{center}
        \caption{
          Statistics for \overnight.
          Each example belongs to exactly one domain.
          The subset for pre-training is 1\% of \(\SplitTrain\).
        }
        \tablefontsizeone
        \input{\tablepathovernight-statistics-of-train-set.tex\inputfilesuffix}
      \end{center}
    \end{subtable}
    \label{tab:statistics-of-train-set}
  \end{center}
\end{table}

\paragraphheaddot{Search}
We use greedy search and beam search algorithms that are implemented in the transformers library \citep{DBLP:conf/emnlp/WolfDSCDMCRLFDS20}.
The search implementations can take a scoring function \(s_{\ActBase}\) as an argument to predict an action sequence \(\actseq\) when given an utterance \(\uttersc\) \Crefp{sec:mask-caching}.
At test time, our semantic parser uses greedy search by default, and additionally uses beam search with a beam size of 4.


\paragraphheaddot{Execution}
A search process finds an action sequence \(\actseq\), from which an executable logical form \(\lflf{\actseq}\) is derived.
In \kqapro, a logical form \(\lflf{\actseq}\) is written as an \mbox{S-expression}, so a transpiler \citep{Hissp_version:0.3.0} converts \(\lflf{\actseq}\) into Python code on the fly,
then the code is executed over a KB \(\kb\) to produce the denotation \(\denot\).
In \overnight, a logical form \(\lflf{\actseq}\) is a \lambdadcs expression, which is executed over a KB \(\kb\) for a domain to produce the denotation \(\denot\).
To execute a \lambdadcs expression, we use the SEMPRE library \citep{DBLP:conf/emnlp/BerantCFL13}.

\paragraphheaddot{Evaluation}
As an evaluation measure, we use accuracy, which is the fraction of examples where the predicted denotation \(\denot\) and the gold denotation \(\lfgd\) are identical.
In addition, to observe the progress of search during weakly-supervised learning, we use oracle accuracy, which is the fraction of examples where a search algorithm finds at least one action sequence \(\actseq\) that constructs a consistent logical form \(\lflf{\actseq}\), whose denotation \(\denot\) is identical to a gold denotation \(\lfgd\) for a given utterance \(\uttersc\) \Crefp{fig:kqapro-weaksup-search,fig:overnight-weaksup-search}.

\paragraphheaddot{Training procedure}
In strongly-supervised learning, the parameters \(\theta\) are optimized by maximizing the objective \(\jml(\theta \scbar \SplitTrain)\) \refeqp{eq:loss}, and evaluate the semantic parser \(f_\theta\) with each function \(\ActBase \in \ActFuncSet\) on \(\SplitVal\).
Once the training is complete, each function \(\ActBase \in \ActFuncSet\) has a checkpoint of parameters, with which \(\ActBase\) achieves the highest accuracy on \(\SplitVal\) during training.
\InCref{sec:results-on-strongsup}, we report accuracies by the checkpoints.

In weakly-supervised learning, the parameters of a weakly-supervised model with a function \(\ActBase \in \ActFuncSet\) are initialized from those from a strongly-supervised model \Crefp{tab:kqapro-varying-result,tab:overnight-varying-result} that uses \(\ActBase\) and that is pre-trained with a small subset of \(\SplitTrain\) \Crefp{tab:statistics-of-train-set}.
The parameters are fine-tuned by maximizing the objective \(\jmml(\theta \scbar \SplitTrainWS)\) \refeqp{eq:weaksup-loss}.
Unlike strongly-supervised learning, we independently train each model with a different function \(\ActBase \in \ActFuncSet\), because parameters cannot be shared during training as each \(\ActBase \in \ActFuncSet\) finds a different set \(\ActSeqSetWS\).
We also use data parallelism to reduce training time \citep{accelerate}; therefore, examples that are used in a search step or an optimization step are distributed to multiple processes.
\InCref{sec:results-on-weaksup}, we report accuracies of the models with functions in \(\ActFuncSet\).

\paragraphheaddot{Hyperparameters}
In strongly-supervised learning, we adapt the hyperparameters from \bartkopl \citep{DBLP:conf/acl/CaoSPNX0LHZ22}, which is a previous semantic parser on \kqapro.
The number of epochs is 25, and the batch size is 16.
For each update on parameters when given a batch, the learning rate linearly increases from 0 to 3e-5 for the first 2.5 epochs, then linearly decreases to 0.
The parameters of a model are optimized by AdamW \citep{DBLP:journals/corr/abs-1711-05101}, which takes the learning rate and other arguments with the following values; \(\beta_1\) is 0.9, \(\beta_2\) is 0.999, \(\epsilon\) is 1e-8 and the weight decay rate \(\lambda\) is 1e-5.

In weakly-supervised learning, we use 4 processes for data parallelism, and repeat the cycle of the search step and the optimization step 16 times. 
For each search step, the beam size is 8, and beam search returns 8 action sequences for each example.
For each optimization step, the number of epochs is 8.
The batch size of each process is 16, where the gradients of distributed batches are averaged for each update on parameters.
The learning rate is constantly 2e-5.
The hyperparameters for AdamW are same with those of strongly-supervised learning.

\begin{table}[t]
  \begin{center}
    \caption{
      Accuracies \accunit of strongly-supervised semantic parsers for \kqapro on the overall \(\SplitVal\), the overall \(\SplitTest\) and each category of examples in \(\SplitTest\).
    }

    \tablefontsizeone

    \input{\tablepathkqapro-strongsup-main.tex\inputfilesuffix}
    \label{tab:kqapro-strongsup-main-result}
  \end{center}

  \begin{center}
    \anotherTableCaption{
      Accuracies \accunit of strongly-supervised semantic parsers for \overnight on the overall \(\SplitVal\), the overall \(\SplitTest\) and each domain of examples in \(\SplitTest\).
      The postfix \MD for some model means that the model is trained with examples of all domains, or that some parameters of the model are shared across all domains.
    }

    \tablefontsizeone

    \input{\tablepathovernight-strongsup-main.tex\inputfilesuffix}
    \label{tab:overnight-strongsup-main-result}
  \end{center}
\end{table}

\section{Results on Strongly-Supervised Learning} \label{sec:results-on-strongsup}
We compare semantic parsers of ours and previous work \inCref{sec:main-results-on-strongsup},
compare our semantic parsers based on \bartlarge models \inCref{sec:results-on-large-plms},
analyze functions in \(\ActFuncSet\) with different subsets of \(\SplitTrain\) \inCref{sec:effect-of-constraints},
and perform ablation study of candidate expressions with different subsets of \(\SplitTrain\) \inCref{sec:effect-of-candidate-expressions}.

\subsection{Main Results} \label{sec:main-results-on-strongsup}

\begin{table}[t]
  \begin{center}
    \caption{
      Accuracies \accunit of strongly-supervised semantic parsers with \bartlarge for \kqapro on the overall \(\SplitVal\) and each category of examples in \(\SplitVal\).
    }

    \tablefontsizeone

    \input{\tablepathkqapro-strongsup-bart-large.tex\inputfilesuffix}
    \label{tab:kqapro-strongsup-result-bart-large}
  \end{center}

  \begin{center}
    \anotherTableCaption{
      Accuracies \accunit of strongly-supervised semantic parsers with \bartlarge for \overnight on the overall \(\SplitVal\), the overall \(\SplitTest\) and each \revisedtwo{domain} of examples in \(\SplitTest\).
    }

    \tablefontsizeone

    \input{\tablepathovernight-strongsup-bart-large.tex\inputfilesuffix}
    \label{tab:overnight-strongsup-result-bart-large}
  \end{center}
\end{table}

We report the accuracies of our semantic parsers, and compare the accuracies with those of previous semantic parsers \citep{DBLP:conf/acl/CaoSPNX0LHZ22,DBLP:conf/emnlp/NieCSST0LZ22,DBLP:conf/aaai/NieS0DH00LZ23} \Crefp{tab:kqapro-strongsup-main-result,tab:overnight-strongsup-main-result}.
The accuracies are computed on the overall \(\SplitVal\), the overall \(\SplitTest\), and each category \citep{DBLP:conf/acl/CaoSPNX0LHZ22} or domain \citep{DBLP:conf/acl/WangBL15} of examples in \(\SplitTest\).
If semantic parsers with some functions in \(\SetOfActsNoHybr\) achieved the same result, we report their accuracies without duplication.

The previous semantic parsers are \bartkopl\citep{DBLP:conf/acl/CaoSPNX0LHZ22}, \graphqir\citep{DBLP:conf/emnlp/NieCSST0LZ22} and \semanchor\citep{DBLP:conf/aaai/NieS0DH00LZ23}.
The three previous semantic parsers, as well as ours, are developed with \bartbase.
The \bartkopl model predicts logical forms written in \kopl, which is linearized in postfix representations with respect to dependencies between function calls. 
The \graphqir model predicts intermediate representations written in the \graphqirlang language, which resembles English. 
The \semanchor model predicts logical forms written in SPARQL for \kqapro and in \lambdadcs for \overnight, and the model learns from sub-tasks about semantic anchors. 

All of our semantic parsers that are trained with \(\SplitTrain\) achieved higher accuracies on the overall \(\SplitTest\) than the previous semantic parsers \Crefp{tab:kqapro-strongsup-main-result,tab:overnight-strongsup-main-result}.
The models with \(\ActNone\) achieved decent accuracies without using any constraint during parsing;
this result shows that a seq2seq PLM can be effectively fine-tuned to predict a sequence of actions that are production rules.
Some models with \(\ActTypeWU\) or \(\ActType\) slightly increased our accuracies.
The models with \(\ActHybr\) noticeably increased our accuracies.
Finally, when the beam size was 4, the models with \(\ActHybr\) achieved the highest accuracies on the overall \(\SplitVal\) and the overall \(\SplitTest\).

However, our semantic parsers achieved lower accuracies in \(\SplitTest\) than \graphqir on some categories of \kqapro and some domains of \overnight.
\graphqir achieved the highest accuracies in the categories of \emph{Logical}, \emph{Count} and \emph{Zero-shot} of \kqapro, and in the domains of \emph{Publications}, \emph{Recipes} and \emph{Restaurants} of \overnight.
\ours and \graphqir have different designs of actions for intermediate representations: production rules for S-expressions, and tokens for English-like expressions.
These comparisons indicate that the two designs generalize differently on categories and on domains.


\subsection{Results on Large PLMs} \label{sec:results-on-large-plms}

We report the accuracies of our semantic parsers with \bartlarge \Crefp{tab:kqapro-strongsup-result-bart-large,tab:overnight-strongsup-result-bart-large}, which generalize better on seq2seq tasks than \bartbase \Crefp{tab:kqapro-strongsup-main-result,tab:overnight-strongsup-main-result}.
The models with \(\ActTypeWU\) and \(\ActType\) achieved higher accuracies than the models with \(\ActNone\), but the models with \(\ActTypeWU\) and the models with \(\ActType\) achieved the same accuracies.
These results suggest that \revisedtwo{union types} are not significantly effective for a large model.
In contrast, the models with \(\ActHybr\) noticeably increased accuracies.
These results suggest that  candidate expressions are effective for a large model, as training examples cannot include all possible mappings of natural language phrases and the corresponding KB elements, and as a seq2seq PLM cannot remember all the mappings exactly.

\subsection{Effect of Constraints on Actions} \label{sec:effect-of-constraints}

\begin{table}[t]
  \begin{center}
    \caption{Accuracies \accunit of strongly-supervised semantic parsers for \kqapro on \(\SplitVal\) with functions in \(\ActFuncSet\).}

    \tablefontsizeone

    \input{\tablepathkqapro-varying.tex\inputfilesuffix}

    \label{tab:kqapro-varying-result}
  \end{center}

  \begin{center}
    \anotherTableCaption{Accuracies \accunit of strongly-supervised semantic parsers for \overnight on \(\SplitTest\) with functions in \(\ActFuncSet\).}

    \tablefontsizeone

    \input{\tablepathovernight-varying.tex\inputfilesuffix}

    \label{tab:overnight-varying-result}
  \end{center}
\end{table}

We report the accuracies of our semantic parsers with different subsets of \(\SplitTrain\) to show the differences between functions in \(\ActFuncSet\) \Crefp{tab:kqapro-varying-result,tab:overnight-varying-result}.
The functions in \(\ActFuncSet\) are listed in decreasing order of the number of applied constraints: \(\ActHybr\), \(\ActType\), \(\ActTypeWU\) and \(\ActNone\).
In the same order, the size of the action set \(\ActBaseFn{\lfir{\actseqic}}\) for a function \(\ActBase \in \ActFuncSet\) increases \refeqp{eq:subset-relations}.
Therefore, a function \(\ActBase \in \ActFuncSet\) with more constraints results in smaller search space.
As the constraints reject actions that lead to an incorrect logical form, a search algorithm benefits from the small search space.

In particular, candidate expressions, which \(\ActHybr\) uses, made the biggest contribution to the improvement in accuracy.
Candidate expressions are effective when a KB element is differently represented in an utterance \(\uttersc\).
For example, \inCref{fig:sp-example}, the relation \(\lfexpr{"country of citizenship"}\), which is a KB element, is represented as \textit{``a citizen of''} in the utterance \(\uttersc\), and the candidate expressions of the node class \(\lfexpr{keyword-relation}\) can guide a semantic parser to generate the relation.
\begin{revision}
  As another example, \inCref{tab:overnight-example-sp-basketball-relation-entity},
  the relation \(\lfexpr{relation:team}\) is implicitly represented as \textit{``Where''},
  and the candidate expressions of the node class \(\lfexpr{keyword-relation-entity}\) can guide a semantic parser to generate the relation with the KB that is specific to the Basketball domain.
\end{revision}

Although the effects of union types and of other types were small, the type constraints consistently increased accuracies.
Union types, which are especially effective for \kqapro, distinguish among actions in \(\ActSetNlt\), so the union types are useful for node classes (e.g., \(\lfexpr{constant-number}\)) that do not have candidate expressions due to their unlimited number of possible instances.
Other types, which construct compositional structures or atomic units, are effective when the number of training examples is small.

\begin{table}[t]
  \begin{center}
    \caption{Reduced accuracies \accunit of strongly-supervised semantic parsers for \kqapro on \(\SplitVal\) when candidate expressions for specific node classes were not used.}

    \lffontsizefive
    \tablefontsizeone
    \input{\tablepathkqapro-candidate-ablation.tex\inputfilesuffix}

    \label{tab:kqapro-candidate-ablation}
  \end{center}

  \begin{center}
    \anotherTableCaption{Reduced accuracies \accunit of strongly-supervised semantic parsers for \overnight on \(\SplitTest\) when candidate expressions for specific node classes were not used.}

    \lffontsizefive
    \tablefontsizeone
    \input{\tablepathovernight-candidate-ablation.tex\inputfilesuffix}

    \label{tab:overnight-candidate-ablation}
  \end{center}
\end{table}

\subsection{Effect of Candidate Expressions for Each Node Class} \label{sec:effect-of-candidate-expressions}

We report decreases in accuracies of our semantic parsers when candidate expressions for specific node classes were not used \Crefp{tab:kqapro-candidate-ablation,tab:overnight-candidate-ablation}.
The semantic parsers were trained with \(\ActHybr\) and with different subsets of \(\SplitTrain\), then candidate expressions of a node class were disabled for each evaluation.
The candidate expressions for most node classes contributed to accuracies unless the number of training examples was large.

In \kqapro, the node class \(\lfexpr{keyword-entity}\), which has the most candidate expressions \Crefp{tab:kqapro-candidate-expressions}, contributed the most to accuracies.
Other node classes, such as \(\lfexpr{keyword-concept}\), \(\lfexpr{keyword-relation}\) and \(\lfexpr{keyword-attribute-string}\), have many fewer candidate expressions than \(\lfexpr{keyword-entity}\), but they also contributed to accuracies.
The contributions to accuracies would increase when a KB becomes larger and node classes have more candidate expressions.
%

In \overnight, node classes, such as \(\lfexpr{keyword-entity}\), \(\lfexpr{keyword-ent-type}\) and \(\lfexpr{keyword-relation-entity}\), contributed to accuracies, although the numbers of candidate expressions are small \Crefp{tab:overnight-candidate-expressions}.
Most KB elements, \revisedtwo{which have natural language counterparts}, appear as sub-action sequences of gold action sequences in \(\SplitTrain\), but our semantic parser without candidate expressions wrongly predicted some KB elements during inference, when the KB elements were differently represented in utterances.
For example, in the Restaurants domain of \overnight, when an utterance was \textit{``How many meals are served?''} and candidate expressions for the node class \(\lfexpr{keyword-ent-type}\) were disabled, our semantic parser inferred \(\lfexpr{meal}\) as a KB element rather than \(\lfexpr{food}\) which is the correct KB element for \(\lfexpr{keyword-ent-type}\) \Crefp{tab:overnight-example-sp-no-kw-ent-type-in-restaurants}, although \(\lfexpr{food}\) appears as sub-action sequences of gold action sequences in 8.9\% of training examples for the domain.

\section{Results on Weakly-Supervised Learning} \label{sec:results-on-weaksup}

We report the accuracies of our weakly-supervised semantic parsers, where \(\ActHybr\) was most effective among functions in \(\ActFuncSet\) \Crefp{tab:kqapro-weaksup-main-result,tab:overnight-weaksup-main-result}.
The \(\ActHybr\) function significantly contributed to accuracies, and the models with \(\ActHybr\) achieved the highest overall accuracies on \(\SplitVal\) and \(\SplitTest\) when the beam size was 4.
The search space with \(\ActHybr\) consists of logical forms that are valid with respect to types and candidate expressions.
These results suggest that a model with \(\ActHybr\) effectively finds consistent logical forms from the compact search space during search steps,
and the parameters of the model are well updated with the found logical forms during optimization steps,
where the search steps also benefit from the well-optimized parameters.
\Crefp{fig:kqapro-weaksup-learning,fig:overnight-weaksup-learning}.


\begin{table}[t]
  \begin{center}
    \caption{
      Accuracies \accunit of weakly-supervised semantic parsers for \kqapro on the overall \(\SplitVal\), the overall \(\SplitTest\) and each category of examples in \(\SplitTest\).
    }

    \tablefontsizeone

    \input{\tablepathkqapro-weaksup-main.tex\inputfilesuffix}
    \label{tab:kqapro-weaksup-main-result}
  \end{center}

  \begin{center}
    \anotherTableCaption{
      Accuracies \accunit of weakly-supervised semantic parsers for \overnight on the overall \(\SplitVal\), the overall \(\SplitTest\) and each \revisedtwo{domain} of examples in \(\SplitTest\).
    }

    \tablefontsizeone

    \input{\tablepathovernight-weaksup-main.tex\inputfilesuffix}
    \label{tab:overnight-weaksup-main-result}
  \end{center}
\end{table}

However, the models with \(\ActTypeWU\) and \(\ActType\) achieved lower overall accuracies on \(\SplitVal\) and \(\SplitTest\) in \kqapro than the model with \(\ActNone\), although \(\ActNone\) does not use any constraint during parsing \Crefp{tab:kqapro-weaksup-main-result}.
Compared to the models with \(\ActTypeWU\) and \(\ActType\), the model with \(\ActNone\) had lower oracle accuracies during search steps, but higher accuracies during optimization steps \Crefp{fig:kqapro-weaksup-learning}.
This phenomenon results from the problem of spurious logical forms \citep{DBLP:conf/acl/PasupatL16}.
A logical form \(\lflf{\actseq}\) is spurious when it produces a denotation \(\denot\) that is identical to the gold denotation \(\lfgd\) but \(\lflf{\actseq}\) does not reflect the semantics of the input utterance \(\uttersc\).
For example, the category \emph{Verify} is a binary prediction task, where the gold denotation \(\lfgd\) of an example is either ``yes'' or ``no'';
in this case, any syntactically valid logical form that is evaluated to produce the binary value can easily become a spurious logical form.
In contrast, type constraints by \(\ActTypeWU\) and \(\ActType\) are effective in \overnight (\Cref{tab:overnight-weaksup-main-result} and \Cref{fig:overnight-weaksup-learning}),
because a denotation is usually a set of values and few spurious logical forms occur from the denotation.
Therefore, using constraints by types without candidate expressions is ineffective for weakly-supervised learning, when many action sequences that construct spurious logical forms are not filtered during search steps.


\begin{figure}[t]
  \begin{center}
    \begin{subfigure}[h]{\subfigurewidth}
      \begin{center}
        \begingroup
        \lffontsizefour
        \tablefontsizesix
        \newcommand{\domaintype}{kqapro}
        \newcommand{\itertype}{search}
        \newcommand{\iterunit}{Search step}
        \newcommand{\measure}{Oracle accuracy}
        \input{\figurepathweaksup-learning.tex\inputfilesuffix}
        \endgroup

        \lffontsizecaption
        \caption{
          Oracle accuracies \accunit on \(\SplitTrainWS\) for each search step.
        }
        \label{fig:kqapro-weaksup-search}
      \end{center}
    \end{subfigure}
    \hspace{0.06\textwidth}
    \begin{subfigure}[h]{\subfigurewidth}
      \begin{center}
        \begingroup
        \lffontsizefour
        \tablefontsizesix
        \newcommand{\domaintype}{kqapro}
        \newcommand{\itertype}{optim}
        \newcommand{\iterunit}{Optimization step}
        \newcommand{\measure}{Accuracy}
        \input{\figurepathweaksup-learning.tex\inputfilesuffix}
        \endgroup

        \lffontsizecaption
        \caption{
          Accuracies \accunit on \(\SplitVal\) for each optimization step.
        }
        \label{fig:kqapro-weaksup-optim}
      \end{center}
    \end{subfigure}
    \caption{
      Progress of weakly-supervised learning for \kqapro.
    }
    \label{fig:kqapro-weaksup-learning}
  \end{center}
  %
  \evgap
  \begin{center}
    \begin{subfigure}[h]{\subfigurewidth}
      \begin{center}
        \begingroup
        \lffontsizefour
        \tablefontsizesix
        \newcommand{\domaintype}{overnight}
        \newcommand{\itertype}{search}
        \newcommand{\iterunit}{Search step}
        \newcommand{\measure}{Oracle accuracy}
        \input{\figurepathweaksup-learning.tex\inputfilesuffix}
        \endgroup

        \lffontsizecaption
        \caption{
          Oracle accuracies \accunit on \(\SplitTrainWS\) for each search step.
        }
        \label{fig:overnight-weaksup-search}
      \end{center}
    \end{subfigure}
    \hspace{0.06\textwidth}
    \begin{subfigure}[h]{\subfigurewidth}
      \begin{center}
        \begingroup
        \lffontsizefour
        \tablefontsizesix
        \newcommand{\domaintype}{overnight}
        \newcommand{\itertype}{optim}
        \newcommand{\iterunit}{Optimization step}
        \newcommand{\measure}{Accuracy}
        \input{\figurepathweaksup-learning.tex\inputfilesuffix}
        \endgroup

        \lffontsizecaption
        \caption{
          Accuracies \accunit on \(\SplitVal\) for each optimization step.
        }
        \label{fig:overnight-weaksup-optim}
      \end{center}
    \end{subfigure}
    \caption{
      Progress of weakly-supervised learning for \overnight.
    }
    \label{fig:overnight-weaksup-learning}
  \end{center}
\end{figure}

\section{Results on Decoding Speed} \label{sec:results-on-decoding-speed}

We report average times taken by strongly-supervised semantic parsers of ours and previous work to decode output sequences \Crefp{tab:kqapro-decoding-time,tab:overnight-decoding-time}.
As an output sequence, our semantic parser predicts an action sequence, and a previous semantic parser predicts a token sequence.
When mask caching is disabled, our semantic parser with a function \(\ActBase \in \SetOfActsWC\) caches the set \(\ActBaseFn{\lfir{\actseqic}}\) of actions for \(\Type(\lmnt(\lfir{\actseqic}))\) instead of a mask vector.
The previous semantic parsers are \bartkopl for \kqapro and \graphqir for \overnight, where the semantic parsers are publicly available.
We also report the average, maximum and minimum lengths of output sequences for each semantic parsers \Crefp{tab:action-seq-length-statistics}.
For fair comparison, all experiments were performed on the same CPU and GPU: {AMD EPYC 7502 32-Core Processor} and {NVIDIA RTX A5000}.


Sub-type inference was applied to our semantic parsers with all functions in \(\ActFuncSet\), and consistently reduced decoding time.
Our semantic parsers benefit from sub-type inference, as decoding time is proportional to the length of an action sequence.
In particular, sub-type inference was effective for \overnight, as the average length of action sequences was greatly reduced \Crefp{tab:overnight-action-seq-length-statistics}.

Mask caching was applied to our semantic parsers with functions in \(\SetOfActsWC\), and decently reduced decoding time.
In particular, mask caching was effective when a beam size or a batch size was large, as computing a mask tensor from a set of valid actions is the bottleneck of parallel computing.
We also devised an alternative algorithm that reduces iterations with a CPU to create a mask tensor \refappendixp{sec:alternative-decoding-speed-optimization}, but mask caching was more effective than the alternative algorithm.

\begin{table}[t]  
  \begin{center}
    \caption{
      Average time to decode an output sequence in \(\SplitTest\) of \kqapro.
    }
    \tablefontsizeone
    \input{\tablepathkqapro-decoding-time.tex\inputfilesuffix}
    \label{tab:kqapro-decoding-time}
  \end{center}
  \begin{center}
    \anotherTableCaption{
      Average time to decode an output sequence in \(\SplitTest\) of \overnight.
      Because the size \(\setsize{\SplitTest}\) of \overnight is small, we duplicate examples of \(\SplitTest\) 5 times.
    }
    \tablefontsizeone
    \input{\tablepathovernight-decoding-time.tex\inputfilesuffix}
    \label{tab:overnight-decoding-time}
  \end{center}
\end{table}

\begin{revision}
  \section{Discussion}
  We discuss our method in terms of effectiveness, efficiency, flexibility, and limitations.


  \paragraphheaddot{Effectiveness}
  Constrained decoding with \(\ActHybr\), which uses candidate expressions, increased accuracies of semantic parsers for both \kqapro and \overnight \Crefp{tab:kqapro-strongsup-main-result,tab:overnight-strongsup-main-result,tab:kqapro-strongsup-result-bart-large,tab:overnight-strongsup-result-bart-large,tab:kqapro-varying-result,tab:overnight-varying-result,tab:kqapro-candidate-ablation,tab:overnight-candidate-ablation,tab:kqapro-weaksup-main-result,tab:overnight-weaksup-main-result}.
  The function \(\ActHybr\) was effective whether a PLM was small \Crefp{sec:main-results-on-strongsup} or large \Crefp{sec:results-on-large-plms}.
  In particular, we found that \(\ActHybr\) was effective when a KB element was differently represented in an utterance \Crefp{sec:effect-of-constraints,sec:effect-of-candidate-expressions}, and when a node class had many candidate expressions \Crefp{sec:effect-of-candidate-expressions}.
  In addition, constrained decoding with \(\ActHybr\) effectively found consistent logical forms during search steps of weakly-supervised learning \Crefp{sec:results-on-weaksup}.
  Therefore, we demonstrate the effectiveness of constrained decoding with candidate expressions.

  \paragraphheaddot{Efficiency}
  We achieved fast decoding speed with sub-type inference and mask caching \Crefp{tab:kqapro-decoding-time,tab:overnight-decoding-time}.
  Sub-type inference reduces the lengths of action sequences \Crefp{tab:action-seq-length-statistics}.
  Mask caching reduces the time to update a mask tensor, especially when a beam size or a batch size is large.
  The worst-case time complexity of constrained decoding with \(\ActHybr\) for time step \(\tstep\) is \(O(\fF + \tstep)\), in which \(\fF\) is the action set size \(\setsize{\ActCandFn{\lfir{\actseqtmo}}}\) which is smaller than \(\setsize{\ActSet}\) \Crefp{sec:mask-caching}.
  Because the size \(\fF\) can be large, the factor is the bottleneck of decoding speed.
  If we use mask caching to store a few top-sized trie nodes, which have lots of child nodes, we can further increase the decoding speed.

  \paragraphheaddot{Flexibility}
  We applied our grammar to the complex syntax of \kqapro and the multiple domains of \overnight.
  The grammar uses types for compositional structures \Crefp{sec:grammars-with-types} and candidate expressions for KB elements \Crefp{sec:candidate-expressions} that are attached to the compositional structures.
  This combination enables constrained decoding to use multiple tries, which can be constructed from a large KB or a domain-specific KB.
  As a result, the grammar was applied to \kopl and \lambdadcs, and can be further extended to other formal languages.

  \paragraphheaddot{Limitations}
  The effectiveness of our method remains uncertain when the method is applied to other semantic parsers, especially those that feed the information retrieved from KBs into PLMs or LLMs \Crefp{sec:related-work:qa-over-kgs}.
  The semantic parsers have been shown to be effective for other large-scale datasets, such as ComplexWebQuestions \citep{DBLP:conf/naacl/TalmorB18} and GrailQA \citep{DBLP:conf/www/GuKVSLY021}, which we did not experiment with.
  Although \citet{DBLP:conf/emnlp/ShuYLKMQL22} demonstrated that constrained decoding was effective when given the information retrieved from a KB, other advanced retrieval methods have been proposed since then.
  Therefore, combining our constrained decoding method with sophisticated retrieval methods would be a promising direction for future research.

  %
\end{revision}

\section{Conclusion}

We present a grammar augmented with candidate expressions for semantic parsing on a large KB with a seq2seq PLM.
Our grammar has a scalable and efficient design that incorporates both various types and many candidate expressions for a large KB.
The grammar guides a semantic parser to construct compositional structures by using types, and to generate KB elements by using candidate expressions.
We extend type rules with sub-type inference and union types, and implement constraints of candidate expressions with multiple tries.
In addition, the proposed mask caching algorithm increased the speed of our constrained decoding method.
The fast decoding speed improves inference efficiency during test time and in weakly-supervised learning.
We experimented on \kqapro and \overnight, where our semantic parsers are trained with strong supervision and weak supervision.
Semantic parsing with candidate expressions achieved state-of-the-art accuracies on \kqapro and \overnight.

\begin{table}[t]
  \begin{center}
    \caption{Statistics of lengths of output sequences in \(\SplitTrain\).}
    \begin{subtable}[h]{\subtablewidth}
      \begin{center}
        \caption{
          \kqapro
        }
        \tablefontsizetwo
        \input{\tablepathkqapro-action-seq-length-statistics.tex\inputfilesuffix}
        \label{tab:kqapro-action-seq-length-statistics}
      \end{center}
    \end{subtable}
    \begin{subtable}[h]{\subtablewidth}
      \begin{center}
        \caption{
          \overnight
        }
        \tablefontsizetwo
        \input{\tablepathovernight-action-seq-length-statistics.tex\inputfilesuffix}
        \label{tab:overnight-action-seq-length-statistics}
      \end{center}
    \end{subtable}
    \label{tab:action-seq-length-statistics}
  \end{center}
\end{table}

\section*{CRediT authorship contribution statement}

\textbf{Daehwan Nam: }
Conceptualization, Methodology, Software, Validation, Formal analysis, Investigation, Data Curation,
Writing - Original Draft, Writing - Review \& Editing, Visualization.
\textbf{Gary Geunbae Lee: }
Resources, Writing - Review \& Editing, Supervision, Project administration, Funding acquisition.




  \section*{Acknowledgments}

  This work was supported by the IITP~(Institute of Information \& Communications Technology Planning \& Evaluation)-ITRC~(Information Technology Research Center) grant funded by the Korea government (Ministry of Science and ICT) (IITP-2025-RS-2024-00437866) [Contribution Rate: 45\%].
  This research was supported by Culture, Sports and Tourism R\&D Program through the Korea Creative Content Agency grant funded by the Ministry of Culture, Sports and Tourism in 2025 (Project Name: Development of an AI-Based Korean Diagnostic System for Efficient Korean Speaking Learning by Foreigners, Project Number: RS-2025-02413038) [Contribution Rate: 45\%].
  This work was supported by Institute of Information \& communications Technology Planning \& Evaluation~(IITP) grant funded by the Korea government~(MSIT) (No.RS-2019-II191906, Artificial Intelligence Graduate School Program (POSTECH)) [Contribution Rate: 10\%].



\newcommand{\proofreading}{0}
\if \proofreading 1
  \clearpage
  \else
  \fi

\bibliographystyle{elsarticle-num-names}

\providecommand{\forArxiv}{0}
\if \forArxiv 1
  \bibliography{paper.bib}
  \else
  \bibliography{\papercollectionpath bibliography.bib}

\begin{thebibliography}{106}
\expandafter\ifx\csname natexlab\endcsname\relax\def\natexlab#1{#1}\fi
\providecommand{\url}[1]{\texttt{#1}}
\providecommand{\href}[2]{#2}
\providecommand{\path}[1]{#1}
\providecommand{\DOIprefix}{doi:}
\providecommand{\ArXivprefix}{arXiv:}
\providecommand{\URLprefix}{URL: }
\providecommand{\Pubmedprefix}{pmid:}
\providecommand{\doi}[1]{\href{http://dx.doi.org/#1}{\path{#1}}}
\providecommand{\Pubmed}[1]{\href{pmid:#1}{\path{#1}}}
\providecommand{\bibinfo}[2]{#2}
\ifx\xfnm\relax \def\xfnm[#1]{\unskip,\space#1}\fi
\bibitem[{Warren and Pereira(1982)}]{DBLP:journals/coling/WarrenP82}
\bibinfo{author}{D.~H.~D. Warren}, \bibinfo{author}{F.~C.~N. Pereira},
\newblock \bibinfo{title}{An efficient easily adaptable system for interpreting
  natural language queries},
\newblock \bibinfo{journal}{Am. J. Comput. Linguistics} \bibinfo{volume}{8}
  (\bibinfo{year}{1982}) \bibinfo{pages}{110--122}.
\bibitem[{Liang et~al.(2011)Liang, Jordan, and Klein}]{DBLP:conf/acl/LiangJK11}
\bibinfo{author}{P.~Liang}, \bibinfo{author}{M.~I. Jordan},
  \bibinfo{author}{D.~Klein},
\newblock \bibinfo{title}{Learning dependency-based compositional semantics},
\newblock in: \bibinfo{editor}{D.~Lin}, \bibinfo{editor}{Y.~Matsumoto},
  \bibinfo{editor}{R.~Mihalcea} (Eds.), \bibinfo{booktitle}{The 49th Annual
  Meeting of the Association for Computational Linguistics: Human Language
  Technologies, Proceedings of the Conference, 19-24 June, 2011, Portland,
  Oregon, {USA}}, \bibinfo{publisher}{The Association for Computer
  Linguistics}, \bibinfo{year}{2011}, pp. \bibinfo{pages}{590--599}. \URLprefix
  \url{https://aclanthology.org/P11-1060/}.
\bibitem[{Zelle and Mooney(1996)}]{DBLP:conf/aaai/ZelleM96}
\bibinfo{author}{J.~M. Zelle}, \bibinfo{author}{R.~J. Mooney},
\newblock \bibinfo{title}{Learning to parse database queries using inductive
  logic programming},
\newblock in: \bibinfo{editor}{W.~J. Clancey}, \bibinfo{editor}{D.~S. Weld}
  (Eds.), \bibinfo{booktitle}{Proceedings of the Thirteenth National Conference
  on Artificial Intelligence and Eighth Innovative Applications of Artificial
  Intelligence Conference, {AAAI} 96, {IAAI} 96, Portland, Oregon, USA, August
  4-8, 1996, Volume 2}, \bibinfo{publisher}{{AAAI} Press / The {MIT} Press},
  \bibinfo{year}{1996}, pp. \bibinfo{pages}{1050--1055}. \URLprefix
  \url{http://www.aaai.org/Library/AAAI/1996/aaai96-156.php}.
\bibitem[{Cai and Yates(2013)}]{DBLP:conf/acl/CaiY13}
\bibinfo{author}{Q.~Cai}, \bibinfo{author}{A.~Yates},
\newblock \bibinfo{title}{Large-scale semantic parsing via schema matching and
  lexicon extension},
\newblock in: \bibinfo{booktitle}{Proceedings of the 51st Annual Meeting of the
  Association for Computational Linguistics, {ACL} 2013, 4-9 August 2013,
  Sofia, Bulgaria, Volume 1: Long Papers}, \bibinfo{publisher}{The Association
  for Computer Linguistics}, \bibinfo{year}{2013}, pp.
  \bibinfo{pages}{423--433}. \URLprefix
  \url{https://aclanthology.org/P13-1042/}.
\bibitem[{Zettlemoyer and Collins(2005)}]{DBLP:conf/uai/ZettlemoyerC05}
\bibinfo{author}{L.~S. Zettlemoyer}, \bibinfo{author}{M.~Collins},
\newblock \bibinfo{title}{Learning to map sentences to logical form: Structured
  classification with probabilistic categorial grammars},
\newblock in: \bibinfo{booktitle}{{UAI} '05, Proceedings of the 21st Conference
  in Uncertainty in Artificial Intelligence, Edinburgh, Scotland, July 26-29,
  2005}, \bibinfo{publisher}{{AUAI} Press}, \bibinfo{year}{2005}, pp.
  \bibinfo{pages}{658--666}. \URLprefix
  \url{https://dslpitt.org/uai/displayArticleDetails.jsp?mmnu=1\&smnu=2\&article\_id=1209\&proceeding\_id=21}.
\bibitem[{Wong and Mooney(2007)}]{DBLP:conf/acl/WongM07}
\bibinfo{author}{Y.~W. Wong}, \bibinfo{author}{R.~J. Mooney},
\newblock \bibinfo{title}{Learning synchronous grammars for semantic parsing
  with lambda calculus},
\newblock in: \bibinfo{editor}{J.~A. Carroll}, \bibinfo{editor}{A.~van~den
  Bosch}, \bibinfo{editor}{A.~Zaenen} (Eds.), \bibinfo{booktitle}{{ACL} 2007,
  Proceedings of the 45th Annual Meeting of the Association for Computational
  Linguistics, June 23-30, 2007, Prague, Czech Republic},
  \bibinfo{publisher}{The Association for Computational Linguistics},
  \bibinfo{year}{2007}, pp. \bibinfo{pages}{960--967}. \URLprefix
  \url{https://aclanthology.org/P07-1121/}.
\bibitem[{Sutskever et~al.(2014)Sutskever, Vinyals, and
  Le}]{DBLP:conf/nips/SutskeverVL14}
\bibinfo{author}{I.~Sutskever}, \bibinfo{author}{O.~Vinyals},
  \bibinfo{author}{Q.~V. Le},
\newblock \bibinfo{title}{Sequence to sequence learning with neural networks},
\newblock in: \bibinfo{editor}{Z.~Ghahramani}, \bibinfo{editor}{M.~Welling},
  \bibinfo{editor}{C.~Cortes}, \bibinfo{editor}{N.~D. Lawrence},
  \bibinfo{editor}{K.~Q. Weinberger} (Eds.), \bibinfo{booktitle}{Advances in
  Neural Information Processing Systems 27: Annual Conference on Neural
  Information Processing Systems 2014, December 8-13 2014, Montreal, Quebec,
  Canada}, \bibinfo{year}{2014}, pp. \bibinfo{pages}{3104--3112}. \URLprefix
  \url{https://proceedings.neurips.cc/paper/2014/hash/a14ac55a4f27472c5d894ec1c3c743d2-Abstract.html}.
\bibitem[{Bahdanau et~al.(2015)Bahdanau, Cho, and
  Bengio}]{DBLP:journals/corr/BahdanauCB14}
\bibinfo{author}{D.~Bahdanau}, \bibinfo{author}{K.~Cho},
  \bibinfo{author}{Y.~Bengio},
\newblock \bibinfo{title}{Neural machine translation by jointly learning to
  align and translate},
\newblock in: \bibinfo{editor}{Y.~Bengio}, \bibinfo{editor}{Y.~LeCun} (Eds.),
  \bibinfo{booktitle}{3rd International Conference on Learning Representations,
  {ICLR} 2015, San Diego, CA, USA, May 7-9, 2015, Conference Track
  Proceedings}, \bibinfo{year}{2015}, pp. \bibinfo{pages}{118--126}. \URLprefix
  \url{http://arxiv.org/abs/1409.0473}.
\bibitem[{Jia and Liang(2016)}]{DBLP:conf/acl/JiaL16}
\bibinfo{author}{R.~Jia}, \bibinfo{author}{P.~Liang},
\newblock \bibinfo{title}{Data recombination for neural semantic parsing},
\newblock in: \bibinfo{booktitle}{Proceedings of the 54th Annual Meeting of the
  Association for Computational Linguistics, {ACL} 2016, August 7-12, 2016,
  Berlin, Germany, Volume 1: Long Papers}, \bibinfo{publisher}{The Association
  for Computer Linguistics}, \bibinfo{year}{2016}. \URLprefix
  \url{https://doi.org/10.18653/v1/p16-1002}.
  \DOIprefix\doi{10.18653/v1/p16-1002}.
\bibitem[{Dong and Lapata(2016)}]{DBLP:conf/acl/DongL16}
\bibinfo{author}{L.~Dong}, \bibinfo{author}{M.~Lapata},
\newblock \bibinfo{title}{Language to logical form with neural attention},
\newblock in: \bibinfo{booktitle}{Proceedings of the 54th Annual Meeting of the
  Association for Computational Linguistics, {ACL} 2016, August 7-12, 2016,
  Berlin, Germany, Volume 1: Long Papers}, \bibinfo{publisher}{The Association
  for Computer Linguistics}, \bibinfo{year}{2016}, pp. \bibinfo{pages}{33--43}.
  \URLprefix \url{https://doi.org/10.18653/v1/p16-1004}.
  \DOIprefix\doi{10.18653/v1/p16-1004}.
\bibitem[{Yin and Neubig(2017)}]{DBLP:conf/acl/YinN17}
\bibinfo{author}{P.~Yin}, \bibinfo{author}{G.~Neubig},
\newblock \bibinfo{title}{A syntactic neural model for general-purpose code
  generation},
\newblock in: \bibinfo{editor}{R.~Barzilay}, \bibinfo{editor}{M.~Kan} (Eds.),
  \bibinfo{booktitle}{Proceedings of the 55th Annual Meeting of the Association
  for Computational Linguistics, {ACL} 2017, Vancouver, Canada, July 30 -
  August 4, Volume 1: Long Papers}, \bibinfo{publisher}{Association for
  Computational Linguistics}, \bibinfo{year}{2017}, pp.
  \bibinfo{pages}{440--450}. \URLprefix
  \url{https://doi.org/10.18653/v1/P17-1041}.
  \DOIprefix\doi{10.18653/v1/P17-1041}.
\bibitem[{Rabinovich et~al.(2017)Rabinovich, Stern, and
  Klein}]{DBLP:conf/acl/RabinovichSK17}
\bibinfo{author}{M.~Rabinovich}, \bibinfo{author}{M.~Stern},
  \bibinfo{author}{D.~Klein},
\newblock \bibinfo{title}{Abstract syntax networks for code generation and
  semantic parsing},
\newblock in: \bibinfo{editor}{R.~Barzilay}, \bibinfo{editor}{M.~Kan} (Eds.),
  \bibinfo{booktitle}{Proceedings of the 55th Annual Meeting of the Association
  for Computational Linguistics, {ACL} 2017, Vancouver, Canada, July 30 -
  August 4, Volume 1: Long Papers}, \bibinfo{publisher}{Association for
  Computational Linguistics}, \bibinfo{year}{2017}, pp.
  \bibinfo{pages}{1139--1149}. \URLprefix
  \url{https://doi.org/10.18653/v1/P17-1105}.
  \DOIprefix\doi{10.18653/v1/P17-1105}.
\bibitem[{Krishnamurthy et~al.(2017)Krishnamurthy, Dasigi, and
  Gardner}]{DBLP:conf/emnlp/KrishnamurthyDG17}
\bibinfo{author}{J.~Krishnamurthy}, \bibinfo{author}{P.~Dasigi},
  \bibinfo{author}{M.~Gardner},
\newblock \bibinfo{title}{Neural semantic parsing with type constraints for
  semi-structured tables},
\newblock in: \bibinfo{editor}{M.~Palmer}, \bibinfo{editor}{R.~Hwa},
  \bibinfo{editor}{S.~Riedel} (Eds.), \bibinfo{booktitle}{Proceedings of the
  2017 Conference on Empirical Methods in Natural Language Processing, {EMNLP}
  2017, Copenhagen, Denmark, September 9-11, 2017},
  \bibinfo{publisher}{Association for Computational Linguistics},
  \bibinfo{year}{2017}, pp. \bibinfo{pages}{1516--1526}. \URLprefix
  \url{https://doi.org/10.18653/v1/d17-1160}.
  \DOIprefix\doi{10.18653/v1/d17-1160}.
\bibitem[{Guu et~al.(2017)Guu, Pasupat, Liu, and
  Liang}]{DBLP:conf/acl/GuuPLL17}
\bibinfo{author}{K.~Guu}, \bibinfo{author}{P.~Pasupat}, \bibinfo{author}{E.~Z.
  Liu}, \bibinfo{author}{P.~Liang},
\newblock \bibinfo{title}{From language to programs: Bridging reinforcement
  learning and maximum marginal likelihood},
\newblock in: \bibinfo{editor}{R.~Barzilay}, \bibinfo{editor}{M.~Kan} (Eds.),
  \bibinfo{booktitle}{Proceedings of the 55th Annual Meeting of the Association
  for Computational Linguistics, {ACL} 2017, Vancouver, Canada, July 30 -
  August 4, Volume 1: Long Papers}, \bibinfo{publisher}{Association for
  Computational Linguistics}, \bibinfo{year}{2017}, pp.
  \bibinfo{pages}{1051--1062}. \URLprefix
  \url{https://doi.org/10.18653/v1/P17-1097}.
  \DOIprefix\doi{10.18653/v1/P17-1097}.
\bibitem[{Cheng et~al.(2017)Cheng, Reddy, Saraswat, and
  Lapata}]{DBLP:conf/acl/ChengRSL17}
\bibinfo{author}{J.~Cheng}, \bibinfo{author}{S.~Reddy}, \bibinfo{author}{V.~A.
  Saraswat}, \bibinfo{author}{M.~Lapata},
\newblock \bibinfo{title}{Learning structured natural language representations
  for semantic parsing},
\newblock in: \bibinfo{editor}{R.~Barzilay}, \bibinfo{editor}{M.~Kan} (Eds.),
  \bibinfo{booktitle}{Proceedings of the 55th Annual Meeting of the Association
  for Computational Linguistics, {ACL} 2017, Vancouver, Canada, July 30 -
  August 4, Volume 1: Long Papers}, \bibinfo{publisher}{Association for
  Computational Linguistics}, \bibinfo{year}{2017}, pp.
  \bibinfo{pages}{44--55}. \URLprefix
  \url{https://doi.org/10.18653/v1/P17-1005}.
  \DOIprefix\doi{10.18653/v1/P17-1005}.
\bibitem[{Liang et~al.(2017)Liang, Berant, Le, Forbus, and
  Lao}]{DBLP:conf/acl/LiangBLFL17}
\bibinfo{author}{C.~Liang}, \bibinfo{author}{J.~Berant}, \bibinfo{author}{Q.~V.
  Le}, \bibinfo{author}{K.~D. Forbus}, \bibinfo{author}{N.~Lao},
\newblock \bibinfo{title}{Neural symbolic machines: Learning semantic parsers
  on freebase with weak supervision},
\newblock in: \bibinfo{editor}{R.~Barzilay}, \bibinfo{editor}{M.~Kan} (Eds.),
  \bibinfo{booktitle}{Proceedings of the 55th Annual Meeting of the Association
  for Computational Linguistics, {ACL} 2017, Vancouver, Canada, July 30 -
  August 4, Volume 1: Long Papers}, \bibinfo{publisher}{Association for
  Computational Linguistics}, \bibinfo{year}{2017}, pp.
  \bibinfo{pages}{23--33}. \URLprefix
  \url{https://doi.org/10.18653/v1/P17-1003}.
  \DOIprefix\doi{10.18653/v1/P17-1003}.
\bibitem[{Dong and Lapata(2018)}]{DBLP:conf/acl/LapataD18}
\bibinfo{author}{L.~Dong}, \bibinfo{author}{M.~Lapata},
\newblock \bibinfo{title}{Coarse-to-fine decoding for neural semantic parsing},
\newblock in: \bibinfo{editor}{I.~Gurevych}, \bibinfo{editor}{Y.~Miyao} (Eds.),
  \bibinfo{booktitle}{Proceedings of the 56th Annual Meeting of the Association
  for Computational Linguistics, {ACL} 2018, Melbourne, Australia, July 15-20,
  2018, Volume 1: Long Papers}, \bibinfo{publisher}{Association for
  Computational Linguistics}, \bibinfo{year}{2018}, pp.
  \bibinfo{pages}{731--742}. \URLprefix
  \url{https://aclanthology.org/P18-1068/}.
  \DOIprefix\doi{10.18653/v1/P18-1068}.
\bibitem[{Goldman et~al.(2018)Goldman, Latcinnik, Nave, Globerson, and
  Berant}]{DBLP:conf/acl/BerantGGLN18}
\bibinfo{author}{O.~Goldman}, \bibinfo{author}{V.~Latcinnik},
  \bibinfo{author}{E.~Nave}, \bibinfo{author}{A.~Globerson},
  \bibinfo{author}{J.~Berant},
\newblock \bibinfo{title}{Weakly supervised semantic parsing with abstract
  examples},
\newblock in: \bibinfo{editor}{I.~Gurevych}, \bibinfo{editor}{Y.~Miyao} (Eds.),
  \bibinfo{booktitle}{Proceedings of the 56th Annual Meeting of the Association
  for Computational Linguistics, {ACL} 2018, Melbourne, Australia, July 15-20,
  2018, Volume 1: Long Papers}, \bibinfo{publisher}{Association for
  Computational Linguistics}, \bibinfo{year}{2018}, pp.
  \bibinfo{pages}{1809--1819}. \URLprefix
  \url{https://www.aclweb.org/anthology/P18-1168/}.
  \DOIprefix\doi{10.18653/v1/P18-1168}.
\bibitem[{Lewis et~al.(2020)Lewis, Liu, Goyal, Ghazvininejad, Mohamed, Levy,
  Stoyanov, and Zettlemoyer}]{DBLP:conf/acl/LewisLGGMLSZ20}
\bibinfo{author}{M.~Lewis}, \bibinfo{author}{Y.~Liu},
  \bibinfo{author}{N.~Goyal}, \bibinfo{author}{M.~Ghazvininejad},
  \bibinfo{author}{A.~Mohamed}, \bibinfo{author}{O.~Levy},
  \bibinfo{author}{V.~Stoyanov}, \bibinfo{author}{L.~Zettlemoyer},
\newblock \bibinfo{title}{{BART:} denoising sequence-to-sequence pre-training
  for natural language generation, translation, and comprehension},
\newblock in: \bibinfo{editor}{D.~Jurafsky}, \bibinfo{editor}{J.~Chai},
  \bibinfo{editor}{N.~Schluter}, \bibinfo{editor}{J.~R. Tetreault} (Eds.),
  \bibinfo{booktitle}{Proceedings of the 58th Annual Meeting of the Association
  for Computational Linguistics, {ACL} 2020, Online, July 5-10, 2020},
  \bibinfo{publisher}{Association for Computational Linguistics},
  \bibinfo{year}{2020}, pp. \bibinfo{pages}{7871--7880}. \URLprefix
  \url{https://doi.org/10.18653/v1/2020.acl-main.703}.
  \DOIprefix\doi{10.18653/v1/2020.acl-main.703}.
\bibitem[{Raffel et~al.(2020)Raffel, Shazeer, Roberts, Lee, Narang, Matena,
  Zhou, Li, and Liu}]{DBLP:journals/jmlr/RaffelSRLNMZLL20}
\bibinfo{author}{C.~Raffel}, \bibinfo{author}{N.~Shazeer},
  \bibinfo{author}{A.~Roberts}, \bibinfo{author}{K.~Lee},
  \bibinfo{author}{S.~Narang}, \bibinfo{author}{M.~Matena},
  \bibinfo{author}{Y.~Zhou}, \bibinfo{author}{W.~Li}, \bibinfo{author}{P.~J.
  Liu},
\newblock \bibinfo{title}{Exploring the limits of transfer learning with a
  unified text-to-text transformer},
\newblock \bibinfo{journal}{J. Mach. Learn. Res.} \bibinfo{volume}{21}
  (\bibinfo{year}{2020}) \bibinfo{pages}{140:1--140:67}. \URLprefix
  \url{http://jmlr.org/papers/v21/20-074.html}.
\bibitem[{Brown et~al.(2020)Brown, Mann, Ryder, Subbiah, Kaplan, Dhariwal,
  Neelakantan, Shyam, Sastry, Askell, Agarwal, Herbert{-}Voss, Krueger,
  Henighan, Child, Ramesh, Ziegler, Wu, Winter, Hesse, Chen, Sigler, Litwin,
  Gray, Chess, Clark, Berner, McCandlish, Radford, Sutskever, and
  Amodei}]{DBLP:conf/nips/BrownMRSKDNSSAA20}
\bibinfo{author}{T.~B. Brown}, \bibinfo{author}{B.~Mann},
  \bibinfo{author}{N.~Ryder}, \bibinfo{author}{M.~Subbiah},
  \bibinfo{author}{J.~Kaplan}, \bibinfo{author}{P.~Dhariwal},
  \bibinfo{author}{A.~Neelakantan}, \bibinfo{author}{P.~Shyam},
  \bibinfo{author}{G.~Sastry}, \bibinfo{author}{A.~Askell},
  \bibinfo{author}{S.~Agarwal}, \bibinfo{author}{A.~Herbert{-}Voss},
  \bibinfo{author}{G.~Krueger}, \bibinfo{author}{T.~Henighan},
  \bibinfo{author}{R.~Child}, \bibinfo{author}{A.~Ramesh},
  \bibinfo{author}{D.~M. Ziegler}, \bibinfo{author}{J.~Wu},
  \bibinfo{author}{C.~Winter}, \bibinfo{author}{C.~Hesse},
  \bibinfo{author}{M.~Chen}, \bibinfo{author}{E.~Sigler},
  \bibinfo{author}{M.~Litwin}, \bibinfo{author}{S.~Gray},
  \bibinfo{author}{B.~Chess}, \bibinfo{author}{J.~Clark},
  \bibinfo{author}{C.~Berner}, \bibinfo{author}{S.~McCandlish},
  \bibinfo{author}{A.~Radford}, \bibinfo{author}{I.~Sutskever},
  \bibinfo{author}{D.~Amodei},
\newblock \bibinfo{title}{Language models are few-shot learners},
\newblock in: \bibinfo{editor}{H.~Larochelle}, \bibinfo{editor}{M.~Ranzato},
  \bibinfo{editor}{R.~Hadsell}, \bibinfo{editor}{M.~Balcan},
  \bibinfo{editor}{H.~Lin} (Eds.), \bibinfo{booktitle}{Advances in Neural
  Information Processing Systems 33: Annual Conference on Neural Information
  Processing Systems 2020, NeurIPS 2020, December 6-12, 2020, virtual},
  \bibinfo{year}{2020}. \URLprefix
  \url{https://proceedings.neurips.cc/paper/2020/hash/1457c0d6bfcb4967418bfb8ac142f64a-Abstract.html}.
\bibitem[{Chen et~al.(2021)Chen, Tworek, Jun, Yuan, de~Oliveira~Pinto, Kaplan,
  Edwards, Burda, Joseph, Brockman, Ray, Puri, Krueger, Petrov, Khlaaf, Sastry,
  Mishkin, Chan, Gray, Ryder, Pavlov, Power, Kaiser, Bavarian, Winter, Tillet,
  Such, Cummings, Plappert, Chantzis, Barnes, Herbert{-}Voss, Guss, Nichol,
  Paino, Tezak, Tang, Babuschkin, Balaji, Jain, Saunders, Hesse, Carr, Leike,
  Achiam, Misra, Morikawa, Radford, Knight, Brundage, Murati, Mayer, Welinder,
  McGrew, Amodei, McCandlish, Sutskever, and
  Zaremba}]{DBLP:journals/corr/abs-2107-03374}
\bibinfo{author}{M.~Chen}, \bibinfo{author}{J.~Tworek},
  \bibinfo{author}{H.~Jun}, \bibinfo{author}{Q.~Yuan}, \bibinfo{author}{H.~P.
  de~Oliveira~Pinto}, \bibinfo{author}{J.~Kaplan},
  \bibinfo{author}{H.~Edwards}, \bibinfo{author}{Y.~Burda},
  \bibinfo{author}{N.~Joseph}, \bibinfo{author}{G.~Brockman},
  \bibinfo{author}{A.~Ray}, \bibinfo{author}{R.~Puri},
  \bibinfo{author}{G.~Krueger}, \bibinfo{author}{M.~Petrov},
  \bibinfo{author}{H.~Khlaaf}, \bibinfo{author}{G.~Sastry},
  \bibinfo{author}{P.~Mishkin}, \bibinfo{author}{B.~Chan},
  \bibinfo{author}{S.~Gray}, \bibinfo{author}{N.~Ryder},
  \bibinfo{author}{M.~Pavlov}, \bibinfo{author}{A.~Power},
  \bibinfo{author}{L.~Kaiser}, \bibinfo{author}{M.~Bavarian},
  \bibinfo{author}{C.~Winter}, \bibinfo{author}{P.~Tillet},
  \bibinfo{author}{F.~P. Such}, \bibinfo{author}{D.~Cummings},
  \bibinfo{author}{M.~Plappert}, \bibinfo{author}{F.~Chantzis},
  \bibinfo{author}{E.~Barnes}, \bibinfo{author}{A.~Herbert{-}Voss},
  \bibinfo{author}{W.~H. Guss}, \bibinfo{author}{A.~Nichol},
  \bibinfo{author}{A.~Paino}, \bibinfo{author}{N.~Tezak},
  \bibinfo{author}{J.~Tang}, \bibinfo{author}{I.~Babuschkin},
  \bibinfo{author}{S.~Balaji}, \bibinfo{author}{S.~Jain},
  \bibinfo{author}{W.~Saunders}, \bibinfo{author}{C.~Hesse},
  \bibinfo{author}{A.~N. Carr}, \bibinfo{author}{J.~Leike},
  \bibinfo{author}{J.~Achiam}, \bibinfo{author}{V.~Misra},
  \bibinfo{author}{E.~Morikawa}, \bibinfo{author}{A.~Radford},
  \bibinfo{author}{M.~Knight}, \bibinfo{author}{M.~Brundage},
  \bibinfo{author}{M.~Murati}, \bibinfo{author}{K.~Mayer},
  \bibinfo{author}{P.~Welinder}, \bibinfo{author}{B.~McGrew},
  \bibinfo{author}{D.~Amodei}, \bibinfo{author}{S.~McCandlish},
  \bibinfo{author}{I.~Sutskever}, \bibinfo{author}{W.~Zaremba},
\newblock \bibinfo{title}{Evaluating large language models trained on code},
\newblock \bibinfo{journal}{CoRR} \bibinfo{volume}{abs/2107.03374}
  (\bibinfo{year}{2021}). \URLprefix \url{https://arxiv.org/abs/2107.03374}.
  \href{http://arxiv.org/abs/2107.03374}{{\tt arXiv:2107.03374}}.
\bibitem[{Wu et~al.(2021)Wu, Chen, Xin, Han, Sun, Zhang, Chen, Yang, and
  Cai}]{DBLP:conf/acl/Wu0XH0ZCYC20}
\bibinfo{author}{S.~Wu}, \bibinfo{author}{B.~Chen}, \bibinfo{author}{C.~Xin},
  \bibinfo{author}{X.~Han}, \bibinfo{author}{L.~Sun},
  \bibinfo{author}{W.~Zhang}, \bibinfo{author}{J.~Chen},
  \bibinfo{author}{F.~Yang}, \bibinfo{author}{X.~Cai},
\newblock \bibinfo{title}{From paraphrasing to semantic parsing: Unsupervised
  semantic parsing via synchronous semantic decoding},
\newblock in: \bibinfo{editor}{C.~Zong}, \bibinfo{editor}{F.~Xia},
  \bibinfo{editor}{W.~Li}, \bibinfo{editor}{R.~Navigli} (Eds.),
  \bibinfo{booktitle}{Proceedings of the 59th Annual Meeting of the Association
  for Computational Linguistics and the 11th International Joint Conference on
  Natural Language Processing, {ACL/IJCNLP} 2021, (Volume 1: Long Papers),
  Virtual Event, August 1-6, 2021}, \bibinfo{publisher}{Association for
  Computational Linguistics}, \bibinfo{year}{2021}, pp.
  \bibinfo{pages}{5110--5121}. \URLprefix
  \url{https://doi.org/10.18653/v1/2021.acl-long.397}.
  \DOIprefix\doi{10.18653/v1/2021.acl-long.397}.
\bibitem[{Shin et~al.(2021)Shin, Lin, Thomson, Chen, Roy, Platanios, Pauls,
  Klein, Eisner, and Durme}]{DBLP:conf/emnlp/ShinLTCRPPKED21}
\bibinfo{author}{R.~Shin}, \bibinfo{author}{C.~H. Lin},
  \bibinfo{author}{S.~Thomson}, \bibinfo{author}{C.~Chen},
  \bibinfo{author}{S.~Roy}, \bibinfo{author}{E.~A. Platanios},
  \bibinfo{author}{A.~Pauls}, \bibinfo{author}{D.~Klein},
  \bibinfo{author}{J.~Eisner}, \bibinfo{author}{B.~V. Durme},
\newblock \bibinfo{title}{Constrained language models yield few-shot semantic
  parsers},
\newblock in: \bibinfo{editor}{M.~Moens}, \bibinfo{editor}{X.~Huang},
  \bibinfo{editor}{L.~Specia}, \bibinfo{editor}{S.~W. Yih} (Eds.),
  \bibinfo{booktitle}{Proceedings of the 2021 Conference on Empirical Methods
  in Natural Language Processing, {EMNLP} 2021, Virtual Event / Punta Cana,
  Dominican Republic, 7-11 November, 2021}, \bibinfo{publisher}{Association for
  Computational Linguistics}, \bibinfo{year}{2021}, pp.
  \bibinfo{pages}{7699--7715}. \URLprefix
  \url{https://doi.org/10.18653/v1/2021.emnlp-main.608}.
  \DOIprefix\doi{10.18653/v1/2021.emnlp-main.608}.
\bibitem[{Wang et~al.(2023)Wang, Wang, Wang, Cao, Saurous, and
  Kim}]{DBLP:conf/nips/WangW0CSK23}
\bibinfo{author}{B.~Wang}, \bibinfo{author}{Z.~Wang},
  \bibinfo{author}{X.~Wang}, \bibinfo{author}{Y.~Cao}, \bibinfo{author}{R.~A.
  Saurous}, \bibinfo{author}{Y.~Kim},
\newblock \bibinfo{title}{Grammar prompting for domain-specific language
  generation with large language models},
\newblock in: \bibinfo{editor}{A.~Oh}, \bibinfo{editor}{T.~Naumann},
  \bibinfo{editor}{A.~Globerson}, \bibinfo{editor}{K.~Saenko},
  \bibinfo{editor}{M.~Hardt}, \bibinfo{editor}{S.~Levine} (Eds.),
  \bibinfo{booktitle}{Advances in Neural Information Processing Systems 36:
  Annual Conference on Neural Information Processing Systems 2023, NeurIPS
  2023, New Orleans, LA, USA, December 10 - 16, 2023}, \bibinfo{year}{2023}.
  \URLprefix
  \url{http://papers.nips.cc/paper\_files/paper/2023/hash/cd40d0d65bfebb894ccc9ea822b47fa8-Abstract-Conference.html}.
\bibitem[{Scholak et~al.(2021)Scholak, Schucher, and
  Bahdanau}]{DBLP:conf/emnlp/ScholakSB21}
\bibinfo{author}{T.~Scholak}, \bibinfo{author}{N.~Schucher},
  \bibinfo{author}{D.~Bahdanau},
\newblock \bibinfo{title}{{PICARD:} parsing incrementally for constrained
  auto-regressive decoding from language models},
\newblock in: \bibinfo{editor}{M.~Moens}, \bibinfo{editor}{X.~Huang},
  \bibinfo{editor}{L.~Specia}, \bibinfo{editor}{S.~W. Yih} (Eds.),
  \bibinfo{booktitle}{Proceedings of the 2021 Conference on Empirical Methods
  in Natural Language Processing, {EMNLP} 2021, Virtual Event / Punta Cana,
  Dominican Republic, 7-11 November, 2021}, \bibinfo{publisher}{Association for
  Computational Linguistics}, \bibinfo{year}{2021}, pp.
  \bibinfo{pages}{9895--9901}. \URLprefix
  \url{https://doi.org/10.18653/v1/2021.emnlp-main.779}.
  \DOIprefix\doi{10.18653/v1/2021.emnlp-main.779}.
\bibitem[{Poesia et~al.(2022)Poesia, Polozov, Le, Tiwari, Soares, Meek, and
  Gulwani}]{DBLP:conf/iclr/PoesiaP00SMG22}
\bibinfo{author}{G.~Poesia}, \bibinfo{author}{A.~Polozov},
  \bibinfo{author}{V.~Le}, \bibinfo{author}{A.~Tiwari},
  \bibinfo{author}{G.~Soares}, \bibinfo{author}{C.~Meek},
  \bibinfo{author}{S.~Gulwani},
\newblock \bibinfo{title}{Synchromesh: Reliable code generation from
  pre-trained language models},
\newblock in: \bibinfo{booktitle}{The Tenth International Conference on
  Learning Representations, {ICLR} 2022, Virtual Event, April 25-29, 2022},
  \bibinfo{publisher}{OpenReview.net}, \bibinfo{year}{2022}. \URLprefix
  \url{https://openreview.net/forum?id=KmtVD97J43e}.
\bibitem[{Cormen et~al.(2009)Cormen, Leiserson, Rivest, and
  Stein}]{DBLP:books/daglib/0023376}
\bibinfo{author}{T.~H. Cormen}, \bibinfo{author}{C.~E. Leiserson},
  \bibinfo{author}{R.~L. Rivest}, \bibinfo{author}{C.~Stein},
  \bibinfo{title}{Introduction to Algorithms, 3rd Edition},
  \bibinfo{publisher}{{MIT} Press}, \bibinfo{year}{2009}. \URLprefix
  \url{http://mitpress.mit.edu/books/introduction-algorithms}.
\bibitem[{Cao et~al.(2021)Cao, Izacard, Riedel, and
  Petroni}]{DBLP:conf/iclr/CaoI0P21}
\bibinfo{author}{N.~D. Cao}, \bibinfo{author}{G.~Izacard},
  \bibinfo{author}{S.~Riedel}, \bibinfo{author}{F.~Petroni},
\newblock \bibinfo{title}{Autoregressive entity retrieval},
\newblock in: \bibinfo{booktitle}{9th International Conference on Learning
  Representations, {ICLR} 2021, Virtual Event, Austria, May 3-7, 2021},
  \bibinfo{publisher}{OpenReview.net}, \bibinfo{year}{2021}. \URLprefix
  \url{https://openreview.net/forum?id=5k8F6UU39V}.
\bibitem[{Shu et~al.(2022)Shu, Yu, Li, Karlsson, Ma, Qu, and
  Lin}]{DBLP:conf/emnlp/ShuYLKMQL22}
\bibinfo{author}{Y.~Shu}, \bibinfo{author}{Z.~Yu}, \bibinfo{author}{Y.~Li},
  \bibinfo{author}{B.~F. Karlsson}, \bibinfo{author}{T.~Ma},
  \bibinfo{author}{Y.~Qu}, \bibinfo{author}{C.~Lin},
\newblock \bibinfo{title}{{TIARA:} multi-grained retrieval for robust question
  answering over large knowledge base},
\newblock in: \bibinfo{editor}{Y.~Goldberg}, \bibinfo{editor}{Z.~Kozareva},
  \bibinfo{editor}{Y.~Zhang} (Eds.), \bibinfo{booktitle}{Proceedings of the
  2022 Conference on Empirical Methods in Natural Language Processing, {EMNLP}
  2022, Abu Dhabi, United Arab Emirates, December 7-11, 2022},
  \bibinfo{publisher}{Association for Computational Linguistics},
  \bibinfo{year}{2022}, pp. \bibinfo{pages}{8108--8121}. \URLprefix
  \url{https://doi.org/10.18653/v1/2022.emnlp-main.555}.
  \DOIprefix\doi{10.18653/v1/2022.emnlp-main.555}.
\bibitem[{Ye et~al.(2022)Ye, Yavuz, Hashimoto, Zhou, and
  Xiong}]{DBLP:conf/acl/YeYHZX22}
\bibinfo{author}{X.~Ye}, \bibinfo{author}{S.~Yavuz},
  \bibinfo{author}{K.~Hashimoto}, \bibinfo{author}{Y.~Zhou},
  \bibinfo{author}{C.~Xiong},
\newblock \bibinfo{title}{{RNG-KBQA:} generation augmented iterative ranking
  for knowledge base question answering},
\newblock in: \bibinfo{editor}{S.~Muresan}, \bibinfo{editor}{P.~Nakov},
  \bibinfo{editor}{A.~Villavicencio} (Eds.), \bibinfo{booktitle}{Proceedings of
  the 60th Annual Meeting of the Association for Computational Linguistics
  (Volume 1: Long Papers), {ACL} 2022, Dublin, Ireland, May 22-27, 2022},
  \bibinfo{publisher}{Association for Computational Linguistics},
  \bibinfo{year}{2022}, pp. \bibinfo{pages}{6032--6043}. \URLprefix
  \url{https://doi.org/10.18653/v1/2022.acl-long.417}.
  \DOIprefix\doi{10.18653/v1/2022.acl-long.417}.
\bibitem[{Liu et~al.(2022)Liu, Yavuz, Meng, Radev, Xiong, and
  Zhou}]{DBLP:conf/emnlp/LiuYMRXZ22}
\bibinfo{author}{Y.~Liu}, \bibinfo{author}{S.~Yavuz},
  \bibinfo{author}{R.~Meng}, \bibinfo{author}{D.~Radev},
  \bibinfo{author}{C.~Xiong}, \bibinfo{author}{Y.~Zhou},
\newblock \bibinfo{title}{Uni-parser: Unified semantic parser for question
  answering on knowledge base and database},
\newblock in: \bibinfo{editor}{Y.~Goldberg}, \bibinfo{editor}{Z.~Kozareva},
  \bibinfo{editor}{Y.~Zhang} (Eds.), \bibinfo{booktitle}{Proceedings of the
  2022 Conference on Empirical Methods in Natural Language Processing, {EMNLP}
  2022, Abu Dhabi, United Arab Emirates, December 7-11, 2022},
  \bibinfo{publisher}{Association for Computational Linguistics},
  \bibinfo{year}{2022}, pp. \bibinfo{pages}{8858--8869}. \URLprefix
  \url{https://doi.org/10.18653/v1/2022.emnlp-main.605}.
  \DOIprefix\doi{10.18653/v1/2022.emnlp-main.605}.
\bibitem[{Gao et~al.(2025)Gao, Cao, Bu, Zhu, Guan, and
  Yu}]{DBLP:conf/aaai/00010BZ0025}
\bibinfo{author}{J.~Gao}, \bibinfo{author}{J.~Cao}, \bibinfo{author}{R.~Bu},
  \bibinfo{author}{N.~Zhu}, \bibinfo{author}{W.~Guan}, \bibinfo{author}{H.~Yu},
\newblock \bibinfo{title}{Promoting knowledge base question answering by
  directing llms to generate task-relevant logical forms},
\newblock in: \bibinfo{editor}{T.~Walsh}, \bibinfo{editor}{J.~Shah},
  \bibinfo{editor}{Z.~Kolter} (Eds.), \bibinfo{booktitle}{AAAI-25, Sponsored by
  the Association for the Advancement of Artificial Intelligence, February 25 -
  March 4, 2025, Philadelphia, PA, {USA}}, \bibinfo{publisher}{{AAAI} Press},
  \bibinfo{year}{2025}, pp. \bibinfo{pages}{23914--23922}. \URLprefix
  \url{https://doi.org/10.1609/aaai.v39i22.34564}.
  \DOIprefix\doi{10.1609/AAAI.V39I22.34564}.
\bibitem[{Xu et~al.(2025)Xu, Li, Zhang, Lin, Zhu, Zheng, Wu, Zhao, Xu, and
  Chen}]{DBLP:conf/aaai/Xu0ZLZ0000C25}
\bibinfo{author}{D.~Xu}, \bibinfo{author}{X.~Li}, \bibinfo{author}{Z.~Zhang},
  \bibinfo{author}{Z.~Lin}, \bibinfo{author}{Z.~Zhu},
  \bibinfo{author}{Z.~Zheng}, \bibinfo{author}{X.~Wu},
  \bibinfo{author}{X.~Zhao}, \bibinfo{author}{T.~Xu},
  \bibinfo{author}{E.~Chen},
\newblock \bibinfo{title}{Harnessing large language models for knowledge graph
  question answering via adaptive multi-aspect retrieval-augmentation},
\newblock in: \bibinfo{editor}{T.~Walsh}, \bibinfo{editor}{J.~Shah},
  \bibinfo{editor}{Z.~Kolter} (Eds.), \bibinfo{booktitle}{AAAI-25, Sponsored by
  the Association for the Advancement of Artificial Intelligence, February 25 -
  March 4, 2025, Philadelphia, PA, {USA}}, \bibinfo{publisher}{{AAAI} Press},
  \bibinfo{year}{2025}, pp. \bibinfo{pages}{25570--25578}. \URLprefix
  \url{https://doi.org/10.1609/aaai.v39i24.34747}.
  \DOIprefix\doi{10.1609/AAAI.V39I24.34747}.
\bibitem[{Luo et~al.(2024)Luo, E, Tang, Peng, Guo, Zhang, Ma, Dong, Song, Lin,
  Zhu, and Luu}]{DBLP:conf/acl/LuoETPG0MDSLZL24}
\bibinfo{author}{H.~Luo}, \bibinfo{author}{H.~E}, \bibinfo{author}{Z.~Tang},
  \bibinfo{author}{S.~Peng}, \bibinfo{author}{Y.~Guo},
  \bibinfo{author}{W.~Zhang}, \bibinfo{author}{C.~Ma},
  \bibinfo{author}{G.~Dong}, \bibinfo{author}{M.~Song},
  \bibinfo{author}{W.~Lin}, \bibinfo{author}{Y.~Zhu}, \bibinfo{author}{A.~T.
  Luu},
\newblock \bibinfo{title}{Chatkbqa: {A} generate-then-retrieve framework for
  knowledge base question answering with fine-tuned large language models},
\newblock in: \bibinfo{editor}{L.~Ku}, \bibinfo{editor}{A.~Martins},
  \bibinfo{editor}{V.~Srikumar} (Eds.), \bibinfo{booktitle}{Findings of the
  Association for Computational Linguistics, {ACL} 2024, Bangkok, Thailand and
  virtual meeting, August 11-16, 2024}, \bibinfo{publisher}{Association for
  Computational Linguistics}, \bibinfo{year}{2024}, pp.
  \bibinfo{pages}{2039--2056}. \URLprefix
  \url{https://doi.org/10.18653/v1/2024.findings-acl.122}.
  \DOIprefix\doi{10.18653/V1/2024.FINDINGS-ACL.122}.
\bibitem[{Li et~al.(2024)Li, Luo, and Lu}]{DBLP:journals/eswa/LiLL24}
\bibinfo{author}{J.~Li}, \bibinfo{author}{X.~Luo}, \bibinfo{author}{G.~Lu},
\newblock \bibinfo{title}{{GS-CBR-KBQA:} graph-structured case-based reasoning
  for knowledge base question answering},
\newblock \bibinfo{journal}{Expert Syst. Appl.} \bibinfo{volume}{257}
  (\bibinfo{year}{2024}) \bibinfo{pages}{125090}. \URLprefix
  \url{https://doi.org/10.1016/j.eswa.2024.125090}.
  \DOIprefix\doi{10.1016/J.ESWA.2024.125090}.
\bibitem[{Feng and He(2025)}]{DBLP:conf/coling/FengH25}
\bibinfo{author}{T.~Feng}, \bibinfo{author}{L.~He},
\newblock \bibinfo{title}{{RGR-KBQA:} generating logical forms for question
  answering using knowledge-graph-enhanced large language model},
\newblock in: \bibinfo{editor}{O.~Rambow}, \bibinfo{editor}{L.~Wanner},
  \bibinfo{editor}{M.~Apidianaki}, \bibinfo{editor}{H.~Al{-}Khalifa},
  \bibinfo{editor}{B.~D. Eugenio}, \bibinfo{editor}{S.~Schockaert} (Eds.),
  \bibinfo{booktitle}{Proceedings of the 31st International Conference on
  Computational Linguistics, {COLING} 2025, Abu Dhabi, UAE, January 19-24,
  2025}, \bibinfo{publisher}{Association for Computational Linguistics},
  \bibinfo{year}{2025}, pp. \bibinfo{pages}{3057--3070}. \URLprefix
  \url{https://aclanthology.org/2025.coling-main.205/}.
\bibitem[{Dasigi et~al.(2019)Dasigi, Gardner, Murty, Zettlemoyer, and
  Hovy}]{DBLP:conf/naacl/Dasigi0MZH19}
\bibinfo{author}{P.~Dasigi}, \bibinfo{author}{M.~Gardner},
  \bibinfo{author}{S.~Murty}, \bibinfo{author}{L.~Zettlemoyer},
  \bibinfo{author}{E.~H. Hovy},
\newblock \bibinfo{title}{Iterative search for weakly supervised semantic
  parsing},
\newblock in: \bibinfo{editor}{J.~Burstein}, \bibinfo{editor}{C.~Doran},
  \bibinfo{editor}{T.~Solorio} (Eds.), \bibinfo{booktitle}{Proceedings of the
  2019 Conference of the North American Chapter of the Association for
  Computational Linguistics: Human Language Technologies, {NAACL-HLT} 2019,
  Minneapolis, MN, USA, June 2-7, 2019, Volume 1 (Long and Short Papers)},
  \bibinfo{publisher}{Association for Computational Linguistics},
  \bibinfo{year}{2019}, pp. \bibinfo{pages}{2669--2680}. \URLprefix
  \url{https://doi.org/10.18653/v1/n19-1273}.
  \DOIprefix\doi{10.18653/v1/n19-1273}.
\bibitem[{Xu et~al.(2020)Xu, Semnani, Campagna, and
  Lam}]{DBLP:conf/emnlp/XuSCL20}
\bibinfo{author}{S.~Xu}, \bibinfo{author}{S.~J. Semnani},
  \bibinfo{author}{G.~Campagna}, \bibinfo{author}{M.~S. Lam},
\newblock \bibinfo{title}{Autoqa: From databases to {QA} semantic parsers with
  only synthetic training data},
\newblock in: \bibinfo{editor}{B.~Webber}, \bibinfo{editor}{T.~Cohn},
  \bibinfo{editor}{Y.~He}, \bibinfo{editor}{Y.~Liu} (Eds.),
  \bibinfo{booktitle}{Proceedings of the 2020 Conference on Empirical Methods
  in Natural Language Processing, {EMNLP} 2020, Online, November 16-20, 2020},
  \bibinfo{publisher}{Association for Computational Linguistics},
  \bibinfo{year}{2020}, pp. \bibinfo{pages}{422--434}. \URLprefix
  \url{https://doi.org/10.18653/v1/2020.emnlp-main.31}.
  \DOIprefix\doi{10.18653/v1/2020.emnlp-main.31}.
\bibitem[{Yin et~al.(2022)Yin, Wieting, Sil, and
  Neubig}]{DBLP:conf/acl/YinWSN22}
\bibinfo{author}{P.~Yin}, \bibinfo{author}{J.~Wieting},
  \bibinfo{author}{A.~Sil}, \bibinfo{author}{G.~Neubig},
\newblock \bibinfo{title}{On the ingredients of an effective zero-shot semantic
  parser},
\newblock in: \bibinfo{editor}{S.~Muresan}, \bibinfo{editor}{P.~Nakov},
  \bibinfo{editor}{A.~Villavicencio} (Eds.), \bibinfo{booktitle}{Proceedings of
  the 60th Annual Meeting of the Association for Computational Linguistics
  (Volume 1: Long Papers), {ACL} 2022, Dublin, Ireland, May 22-27, 2022},
  \bibinfo{publisher}{Association for Computational Linguistics},
  \bibinfo{year}{2022}, pp. \bibinfo{pages}{1455--1474}. \URLprefix
  \url{https://aclanthology.org/2022.acl-long.103}.
\bibitem[{Wu et~al.(2023)Wu, Xin, Lin, Han, Liu, Chen, Yang, Wan, and
  Sun}]{DBLP:conf/acl/WuXLHLCYW023}
\bibinfo{author}{S.~Wu}, \bibinfo{author}{C.~Xin}, \bibinfo{author}{H.~Lin},
  \bibinfo{author}{X.~Han}, \bibinfo{author}{C.~Liu},
  \bibinfo{author}{J.~Chen}, \bibinfo{author}{F.~Yang},
  \bibinfo{author}{G.~Wan}, \bibinfo{author}{L.~Sun},
\newblock \bibinfo{title}{Ambiguous learning from retrieval: Towards zero-shot
  semantic parsing},
\newblock in: \bibinfo{editor}{A.~Rogers}, \bibinfo{editor}{J.~L.
  Boyd{-}Graber}, \bibinfo{editor}{N.~Okazaki} (Eds.),
  \bibinfo{booktitle}{Proceedings of the 61st Annual Meeting of the Association
  for Computational Linguistics (Volume 1: Long Papers), {ACL} 2023, Toronto,
  Canada, July 9-14, 2023}, \bibinfo{publisher}{Association for Computational
  Linguistics}, \bibinfo{year}{2023}, pp. \bibinfo{pages}{14081--14094}.
  \URLprefix \url{https://doi.org/10.18653/v1/2023.acl-long.787}.
  \DOIprefix\doi{10.18653/V1/2023.ACL-LONG.787}.
\bibitem[{Cao et~al.(2022)Cao, Shi, Pan, Nie, Xiang, Hou, Li, He, and
  Zhang}]{DBLP:conf/acl/CaoSPNX0LHZ22}
\bibinfo{author}{S.~Cao}, \bibinfo{author}{J.~Shi}, \bibinfo{author}{L.~Pan},
  \bibinfo{author}{L.~Nie}, \bibinfo{author}{Y.~Xiang},
  \bibinfo{author}{L.~Hou}, \bibinfo{author}{J.~Li}, \bibinfo{author}{B.~He},
  \bibinfo{author}{H.~Zhang},
\newblock \bibinfo{title}{{KQA} pro: {A} dataset with explicit compositional
  programs for complex question answering over knowledge base},
\newblock in: \bibinfo{editor}{S.~Muresan}, \bibinfo{editor}{P.~Nakov},
  \bibinfo{editor}{A.~Villavicencio} (Eds.), \bibinfo{booktitle}{Proceedings of
  the 60th Annual Meeting of the Association for Computational Linguistics
  (Volume 1: Long Papers), {ACL} 2022, Dublin, Ireland, May 22-27, 2022},
  \bibinfo{publisher}{Association for Computational Linguistics},
  \bibinfo{year}{2022}, pp. \bibinfo{pages}{6101--6119}. \URLprefix
  \url{https://doi.org/10.18653/v1/2022.acl-long.422}.
  \DOIprefix\doi{10.18653/v1/2022.acl-long.422}.
\bibitem[{Wang et~al.(2015)Wang, Berant, and Liang}]{DBLP:conf/acl/WangBL15}
\bibinfo{author}{Y.~Wang}, \bibinfo{author}{J.~Berant},
  \bibinfo{author}{P.~Liang},
\newblock \bibinfo{title}{Building a semantic parser overnight},
\newblock in: \bibinfo{booktitle}{Proceedings of the 53rd Annual Meeting of the
  Association for Computational Linguistics and the 7th International Joint
  Conference on Natural Language Processing of the Asian Federation of Natural
  Language Processing, {ACL} 2015, July 26-31, 2015, Beijing, China, Volume 1:
  Long Papers}, \bibinfo{publisher}{The Association for Computer Linguistics},
  \bibinfo{year}{2015}, pp. \bibinfo{pages}{1332--1342}. \URLprefix
  \url{https://doi.org/10.3115/v1/p15-1129}.
  \DOIprefix\doi{10.3115/v1/p15-1129}.
\bibitem[{Yin and Neubig(2018)}]{DBLP:conf/emnlp/YinN18}
\bibinfo{author}{P.~Yin}, \bibinfo{author}{G.~Neubig},
\newblock \bibinfo{title}{{TRANX:} {A} transition-based neural abstract syntax
  parser for semantic parsing and code generation},
\newblock in: \bibinfo{editor}{E.~Blanco}, \bibinfo{editor}{W.~Lu} (Eds.),
  \bibinfo{booktitle}{Proceedings of the 2018 Conference on Empirical Methods
  in Natural Language Processing, {EMNLP} 2018: System Demonstrations,
  Brussels, Belgium, October 31 - November 4, 2018},
  \bibinfo{publisher}{Association for Computational Linguistics},
  \bibinfo{year}{2018}, pp. \bibinfo{pages}{7--12}. \URLprefix
  \url{https://doi.org/10.18653/v1/d18-2002}.
  \DOIprefix\doi{10.18653/v1/d18-2002}.
\bibitem[{Yin et~al.(2018)Yin, Zhou, He, and
  Neubig}]{DBLP:conf/acl/NeubigZYH18}
\bibinfo{author}{P.~Yin}, \bibinfo{author}{C.~Zhou}, \bibinfo{author}{J.~He},
  \bibinfo{author}{G.~Neubig},
\newblock \bibinfo{title}{Structvae: Tree-structured latent variable models for
  semi-supervised semantic parsing},
\newblock in: \bibinfo{editor}{I.~Gurevych}, \bibinfo{editor}{Y.~Miyao} (Eds.),
  \bibinfo{booktitle}{Proceedings of the 56th Annual Meeting of the Association
  for Computational Linguistics, {ACL} 2018, Melbourne, Australia, July 15-20,
  2018, Volume 1: Long Papers}, \bibinfo{publisher}{Association for
  Computational Linguistics}, \bibinfo{year}{2018}, pp.
  \bibinfo{pages}{754--765}. \URLprefix
  \url{https://aclanthology.org/P18-1070/}.
  \DOIprefix\doi{10.18653/v1/P18-1070}.
\bibitem[{Guo et~al.(2019)Guo, Zhan, Gao, Xiao, Lou, Liu, and
  Zhang}]{DBLP:conf/acl/GuoZGXLLZ19}
\bibinfo{author}{J.~Guo}, \bibinfo{author}{Z.~Zhan}, \bibinfo{author}{Y.~Gao},
  \bibinfo{author}{Y.~Xiao}, \bibinfo{author}{J.~Lou},
  \bibinfo{author}{T.~Liu}, \bibinfo{author}{D.~Zhang},
\newblock \bibinfo{title}{Towards complex text-to-sql in cross-domain database
  with intermediate representation},
\newblock in: \bibinfo{editor}{A.~Korhonen}, \bibinfo{editor}{D.~R. Traum},
  \bibinfo{editor}{L.~M{\`{a}}rquez} (Eds.), \bibinfo{booktitle}{Proceedings of
  the 57th Conference of the Association for Computational Linguistics, {ACL}
  2019, Florence, Italy, July 28- August 2, 2019, Volume 1: Long Papers},
  \bibinfo{publisher}{Association for Computational Linguistics},
  \bibinfo{year}{2019}, pp. \bibinfo{pages}{4524--4535}. \URLprefix
  \url{https://doi.org/10.18653/v1/p19-1444}.
  \DOIprefix\doi{10.18653/v1/p19-1444}.
\bibitem[{Wang et~al.(2020)Wang, Shin, Liu, Polozov, and
  Richardson}]{DBLP:conf/acl/WangSLPR20}
\bibinfo{author}{B.~Wang}, \bibinfo{author}{R.~Shin}, \bibinfo{author}{X.~Liu},
  \bibinfo{author}{O.~Polozov}, \bibinfo{author}{M.~Richardson},
\newblock \bibinfo{title}{{RAT-SQL:} relation-aware schema encoding and linking
  for text-to-sql parsers},
\newblock in: \bibinfo{editor}{D.~Jurafsky}, \bibinfo{editor}{J.~Chai},
  \bibinfo{editor}{N.~Schluter}, \bibinfo{editor}{J.~R. Tetreault} (Eds.),
  \bibinfo{booktitle}{Proceedings of the 58th Annual Meeting of the Association
  for Computational Linguistics, {ACL} 2020, Online, July 5-10, 2020},
  \bibinfo{publisher}{Association for Computational Linguistics},
  \bibinfo{year}{2020}, pp. \bibinfo{pages}{7567--7578}. \URLprefix
  \url{https://doi.org/10.18653/v1/2020.acl-main.677}.
  \DOIprefix\doi{10.18653/v1/2020.acl-main.677}.
\bibitem[{Gupta et~al.(2021)Gupta, Singh, and
  Gardner}]{DBLP:conf/acl/Gupta0020}
\bibinfo{author}{N.~Gupta}, \bibinfo{author}{S.~Singh},
  \bibinfo{author}{M.~Gardner},
\newblock \bibinfo{title}{Enforcing consistency in weakly supervised semantic
  parsing},
\newblock in: \bibinfo{editor}{C.~Zong}, \bibinfo{editor}{F.~Xia},
  \bibinfo{editor}{W.~Li}, \bibinfo{editor}{R.~Navigli} (Eds.),
  \bibinfo{booktitle}{Proceedings of the 59th Annual Meeting of the Association
  for Computational Linguistics and the 11th International Joint Conference on
  Natural Language Processing, {ACL/IJCNLP} 2021, (Volume 2: Short Papers),
  Virtual Event, August 1-6, 2021}, \bibinfo{publisher}{Association for
  Computational Linguistics}, \bibinfo{year}{2021}, pp.
  \bibinfo{pages}{168--174}. \URLprefix
  \url{https://doi.org/10.18653/v1/2021.acl-short.22}.
  \DOIprefix\doi{10.18653/v1/2021.acl-short.22}.
\bibitem[{Cao et~al.(2021)Cao, Chen, Chen, Zhao, Zhu, and
  Yu}]{DBLP:conf/acl/CaoC0ZZ020}
\bibinfo{author}{R.~Cao}, \bibinfo{author}{L.~Chen}, \bibinfo{author}{Z.~Chen},
  \bibinfo{author}{Y.~Zhao}, \bibinfo{author}{S.~Zhu}, \bibinfo{author}{K.~Yu},
\newblock \bibinfo{title}{{LGESQL:} line graph enhanced text-to-sql model with
  mixed local and non-local relations},
\newblock in: \bibinfo{editor}{C.~Zong}, \bibinfo{editor}{F.~Xia},
  \bibinfo{editor}{W.~Li}, \bibinfo{editor}{R.~Navigli} (Eds.),
  \bibinfo{booktitle}{Proceedings of the 59th Annual Meeting of the Association
  for Computational Linguistics and the 11th International Joint Conference on
  Natural Language Processing, {ACL/IJCNLP} 2021, (Volume 1: Long Papers),
  Virtual Event, August 1-6, 2021}, \bibinfo{publisher}{Association for
  Computational Linguistics}, \bibinfo{year}{2021}, pp.
  \bibinfo{pages}{2541--2555}. \URLprefix
  \url{https://doi.org/10.18653/v1/2021.acl-long.198}.
  \DOIprefix\doi{10.18653/v1/2021.acl-long.198}.
\bibitem[{Chen et~al.(2021)Chen, Liu, Yu, Lin, Lou, and
  Jiang}]{DBLP:conf/acl/ChenLYLLJ21}
\bibinfo{author}{S.~Chen}, \bibinfo{author}{Q.~Liu}, \bibinfo{author}{Z.~Yu},
  \bibinfo{author}{C.~Lin}, \bibinfo{author}{J.~Lou},
  \bibinfo{author}{F.~Jiang},
\newblock \bibinfo{title}{Retrack: {A} flexible and efficient framework for
  knowledge base question answering},
\newblock in: \bibinfo{editor}{H.~Ji}, \bibinfo{editor}{J.~C. Park},
  \bibinfo{editor}{R.~Xia} (Eds.), \bibinfo{booktitle}{Proceedings of the Joint
  Conference of the 59th Annual Meeting of the Association for Computational
  Linguistics and the 11th International Joint Conference on Natural Language
  Processing, {ACL} 2021 - System Demonstrations, Online, August 1-6, 2021},
  \bibinfo{publisher}{Association for Computational Linguistics},
  \bibinfo{year}{2021}, pp. \bibinfo{pages}{325--336}. \URLprefix
  \url{https://doi.org/10.18653/v1/2021.acl-demo.39}.
  \DOIprefix\doi{10.18653/V1/2021.ACL-DEMO.39}.
\bibitem[{Hochreiter and Schmidhuber(1997)}]{DBLP:journals/neco/HochreiterS97}
\bibinfo{author}{S.~Hochreiter}, \bibinfo{author}{J.~Schmidhuber},
\newblock \bibinfo{title}{Long short-term memory},
\newblock \bibinfo{journal}{Neural Comput.} \bibinfo{volume}{9}
  (\bibinfo{year}{1997}) \bibinfo{pages}{1735--1780}. \URLprefix
  \url{https://doi.org/10.1162/neco.1997.9.8.1735}.
  \DOIprefix\doi{10.1162/neco.1997.9.8.1735}.
\bibitem[{Dou et~al.(2023)Dou, Gao, Pan, Wang, Che, Lou, and
  Zhan}]{DBLP:journals/mlc/DouGPWCLZ23}
\bibinfo{author}{L.~Dou}, \bibinfo{author}{Y.~Gao}, \bibinfo{author}{M.~Pan},
  \bibinfo{author}{D.~Wang}, \bibinfo{author}{W.~Che},
  \bibinfo{author}{J.~Lou}, \bibinfo{author}{D.~Zhan},
\newblock \bibinfo{title}{Unisar: a unified structure-aware autoregressive
  language model for text-to-sql semantic parsing},
\newblock \bibinfo{journal}{Int. J. Mach. Learn. Cybern.} \bibinfo{volume}{14}
  (\bibinfo{year}{2023}) \bibinfo{pages}{4361--4376}. \URLprefix
  \url{https://doi.org/10.1007/s13042-023-01898-3}.
  \DOIprefix\doi{10.1007/S13042-023-01898-3}.
\bibitem[{Knuth(1965)}]{DBLP:journals/iandc/Knuth65}
\bibinfo{author}{D.~E. Knuth},
\newblock \bibinfo{title}{On the translation of languages from left to right},
\newblock \bibinfo{journal}{Inf. Control.} \bibinfo{volume}{8}
  (\bibinfo{year}{1965}) \bibinfo{pages}{607--639}. \URLprefix
  \url{https://doi.org/10.1016/S0019-9958(65)90426-2}.
  \DOIprefix\doi{10.1016/S0019-9958(65)90426-2}.
\bibitem[{Earley(1970)}]{DBLP:journals/cacm/Earley70}
\bibinfo{author}{J.~Earley},
\newblock \bibinfo{title}{An efficient context-free parsing algorithm},
\newblock \bibinfo{journal}{Commun. {ACM}} \bibinfo{volume}{13}
  (\bibinfo{year}{1970}) \bibinfo{pages}{94--102}. \URLprefix
  \url{https://doi.org/10.1145/362007.362035}.
  \DOIprefix\doi{10.1145/362007.362035}.
\bibitem[{O'Sullivan(2019)}]{attoparsec}
\bibinfo{author}{B.~O'Sullivan}, \bibinfo{title}{{attoparsec}: Fast combinator
  parsing for bytestrings and text},
  \bibinfo{howpublished}{\url{https://hackage.haskell.org/package/attoparsec}},
  \bibinfo{year}{2019}.
\bibitem[{Parr and Fisher(2011)}]{DBLP:conf/pldi/ParrF11}
\bibinfo{author}{T.~Parr}, \bibinfo{author}{K.~Fisher},
\newblock \bibinfo{title}{Ll(*): the foundation of the {ANTLR} parser
  generator},
\newblock in: \bibinfo{editor}{M.~W. Hall}, \bibinfo{editor}{D.~A. Padua}
  (Eds.), \bibinfo{booktitle}{Proceedings of the 32nd {ACM} {SIGPLAN}
  Conference on Programming Language Design and Implementation, {PLDI} 2011,
  San Jose, CA, USA, June 4-8, 2011}, \bibinfo{publisher}{{ACM}},
  \bibinfo{year}{2011}, pp. \bibinfo{pages}{425--436}. \URLprefix
  \url{https://doi.org/10.1145/1993498.1993548}.
  \DOIprefix\doi{10.1145/1993498.1993548}.
\bibitem[{Rubin and Berant(2021)}]{DBLP:conf/naacl/RubinB21}
\bibinfo{author}{O.~Rubin}, \bibinfo{author}{J.~Berant},
\newblock \bibinfo{title}{Smbop: Semi-autoregressive bottom-up semantic
  parsing},
\newblock in: \bibinfo{editor}{K.~Toutanova}, \bibinfo{editor}{A.~Rumshisky},
  \bibinfo{editor}{L.~Zettlemoyer}, \bibinfo{editor}{D.~Hakkani{-}T{\"{u}}r},
  \bibinfo{editor}{I.~Beltagy}, \bibinfo{editor}{S.~Bethard},
  \bibinfo{editor}{R.~Cotterell}, \bibinfo{editor}{T.~Chakraborty},
  \bibinfo{editor}{Y.~Zhou} (Eds.), \bibinfo{booktitle}{Proceedings of the 2021
  Conference of the North American Chapter of the Association for Computational
  Linguistics: Human Language Technologies, {NAACL-HLT} 2021, Online, June
  6-11, 2021}, \bibinfo{publisher}{Association for Computational Linguistics},
  \bibinfo{year}{2021}, pp. \bibinfo{pages}{311--324}. \URLprefix
  \url{https://doi.org/10.18653/v1/2021.naacl-main.29}.
  \DOIprefix\doi{10.18653/v1/2021.naacl-main.29}.
\bibitem[{Codd(1970)}]{codd1970relational}
\bibinfo{author}{E.~F. Codd},
\newblock \bibinfo{title}{A relational model of data for large shared data
  banks},
\newblock \bibinfo{journal}{Communications of the ACM} \bibinfo{volume}{13}
  (\bibinfo{year}{1970}) \bibinfo{pages}{377--387}.
\bibitem[{Liang et~al.(2018)Liang, Norouzi, Berant, Le, and
  Lao}]{DBLP:conf/nips/LiangNBLL18}
\bibinfo{author}{C.~Liang}, \bibinfo{author}{M.~Norouzi},
  \bibinfo{author}{J.~Berant}, \bibinfo{author}{Q.~V. Le},
  \bibinfo{author}{N.~Lao},
\newblock \bibinfo{title}{Memory augmented policy optimization for program
  synthesis and semantic parsing},
\newblock in: \bibinfo{editor}{S.~Bengio}, \bibinfo{editor}{H.~M. Wallach},
  \bibinfo{editor}{H.~Larochelle}, \bibinfo{editor}{K.~Grauman},
  \bibinfo{editor}{N.~Cesa{-}Bianchi}, \bibinfo{editor}{R.~Garnett} (Eds.),
  \bibinfo{booktitle}{Advances in Neural Information Processing Systems 31:
  Annual Conference on Neural Information Processing Systems 2018, NeurIPS
  2018, December 3-8, 2018, Montr{\'{e}}al, Canada}, \bibinfo{year}{2018}, pp.
  \bibinfo{pages}{10015--10027}. \URLprefix
  \url{https://proceedings.neurips.cc/paper/2018/hash/f4e369c0a468d3aeeda0593ba90b5e55-Abstract.html}.
\bibitem[{Yin et~al.(2020)Yin, Neubig, Yih, and
  Riedel}]{DBLP:conf/acl/YinNYR20}
\bibinfo{author}{P.~Yin}, \bibinfo{author}{G.~Neubig},
  \bibinfo{author}{W.~Yih}, \bibinfo{author}{S.~Riedel},
\newblock \bibinfo{title}{Tabert: Pretraining for joint understanding of
  textual and tabular data},
\newblock in: \bibinfo{editor}{D.~Jurafsky}, \bibinfo{editor}{J.~Chai},
  \bibinfo{editor}{N.~Schluter}, \bibinfo{editor}{J.~R. Tetreault} (Eds.),
  \bibinfo{booktitle}{Proceedings of the 58th Annual Meeting of the Association
  for Computational Linguistics, {ACL} 2020, Online, July 5-10, 2020},
  \bibinfo{publisher}{Association for Computational Linguistics},
  \bibinfo{year}{2020}, pp. \bibinfo{pages}{8413--8426}. \URLprefix
  \url{https://doi.org/10.18653/v1/2020.acl-main.745}.
  \DOIprefix\doi{10.18653/v1/2020.acl-main.745}.
\bibitem[{Gu and Su(2022)}]{DBLP:conf/coling/Gu022}
\bibinfo{author}{Y.~Gu}, \bibinfo{author}{Y.~Su},
\newblock \bibinfo{title}{Arcaneqa: Dynamic program induction and
  contextualized encoding for knowledge base question answering},
\newblock in: \bibinfo{editor}{N.~Calzolari}, \bibinfo{editor}{C.~Huang},
  \bibinfo{editor}{H.~Kim}, \bibinfo{editor}{J.~Pustejovsky},
  \bibinfo{editor}{L.~Wanner}, \bibinfo{editor}{K.~Choi},
  \bibinfo{editor}{P.~Ryu}, \bibinfo{editor}{H.~Chen},
  \bibinfo{editor}{L.~Donatelli}, \bibinfo{editor}{H.~Ji},
  \bibinfo{editor}{S.~Kurohashi}, \bibinfo{editor}{P.~Paggio},
  \bibinfo{editor}{N.~Xue}, \bibinfo{editor}{S.~Kim},
  \bibinfo{editor}{Y.~Hahm}, \bibinfo{editor}{Z.~He}, \bibinfo{editor}{T.~K.
  Lee}, \bibinfo{editor}{E.~Santus}, \bibinfo{editor}{F.~Bond},
  \bibinfo{editor}{S.~Na} (Eds.), \bibinfo{booktitle}{Proceedings of the 29th
  International Conference on Computational Linguistics, {COLING} 2022,
  Gyeongju, Republic of Korea, October 12-17, 2022},
  \bibinfo{publisher}{International Committee on Computational Linguistics},
  \bibinfo{year}{2022}, pp. \bibinfo{pages}{1718--1731}. \URLprefix
  \url{https://aclanthology.org/2022.coling-1.148}.
\bibitem[{Gu et~al.(2023)Gu, Deng, and Su}]{DBLP:conf/acl/Gu0023}
\bibinfo{author}{Y.~Gu}, \bibinfo{author}{X.~Deng}, \bibinfo{author}{Y.~Su},
\newblock \bibinfo{title}{Don't generate, discriminate: {A} proposal for
  grounding language models to real-world environments},
\newblock in: \bibinfo{editor}{A.~Rogers}, \bibinfo{editor}{J.~L.
  Boyd{-}Graber}, \bibinfo{editor}{N.~Okazaki} (Eds.),
  \bibinfo{booktitle}{Proceedings of the 61st Annual Meeting of the Association
  for Computational Linguistics (Volume 1: Long Papers), {ACL} 2023, Toronto,
  Canada, July 9-14, 2023}, \bibinfo{publisher}{Association for Computational
  Linguistics}, \bibinfo{year}{2023}, pp. \bibinfo{pages}{4928--4949}.
  \URLprefix \url{https://doi.org/10.18653/v1/2023.acl-long.270}.
  \DOIprefix\doi{10.18653/v1/2023.acl-long.270}.
\bibitem[{Wang et~al.(2018)Wang, Tatwawadi, Brockschmidt, Huang, Mao, Polozov,
  and Singh}]{wang2018robust}
\bibinfo{author}{C.~Wang}, \bibinfo{author}{K.~Tatwawadi},
  \bibinfo{author}{M.~Brockschmidt}, \bibinfo{author}{P.-S. Huang},
  \bibinfo{author}{Y.~Mao}, \bibinfo{author}{O.~Polozov},
  \bibinfo{author}{R.~Singh},
\newblock \bibinfo{title}{Robust text-to-sql generation with execution-guided
  decoding},
\newblock \bibinfo{journal}{arXiv preprint arXiv:1807.03100}
  (\bibinfo{year}{2018}).
\bibitem[{Berant et~al.(2013)Berant, Chou, Frostig, and
  Liang}]{DBLP:conf/emnlp/BerantCFL13}
\bibinfo{author}{J.~Berant}, \bibinfo{author}{A.~Chou},
  \bibinfo{author}{R.~Frostig}, \bibinfo{author}{P.~Liang},
\newblock \bibinfo{title}{Semantic parsing on freebase from question-answer
  pairs},
\newblock in: \bibinfo{booktitle}{Proceedings of the 2013 Conference on
  Empirical Methods in Natural Language Processing, {EMNLP} 2013, 18-21 October
  2013, Grand Hyatt Seattle, Seattle, Washington, USA, {A} meeting of SIGDAT, a
  Special Interest Group of the {ACL}}, \bibinfo{publisher}{{ACL}},
  \bibinfo{year}{2013}, pp. \bibinfo{pages}{1533--1544}. \URLprefix
  \url{https://aclanthology.org/D13-1160/}.
\bibitem[{Berant and Liang(2014)}]{DBLP:conf/acl/BerantL14}
\bibinfo{author}{J.~Berant}, \bibinfo{author}{P.~Liang},
\newblock \bibinfo{title}{Semantic parsing via paraphrasing},
\newblock in: \bibinfo{booktitle}{Proceedings of the 52nd Annual Meeting of the
  Association for Computational Linguistics, {ACL} 2014, June 22-27, 2014,
  Baltimore, MD, USA, Volume 1: Long Papers}, \bibinfo{publisher}{The
  Association for Computer Linguistics}, \bibinfo{year}{2014}, pp.
  \bibinfo{pages}{1415--1425}. \URLprefix
  \url{https://doi.org/10.3115/v1/p14-1133}.
  \DOIprefix\doi{10.3115/v1/p14-1133}.
\bibitem[{Zhang et~al.(2017)Zhang, Pasupat, and
  Liang}]{DBLP:conf/emnlp/ZhangPL17}
\bibinfo{author}{Y.~Zhang}, \bibinfo{author}{P.~Pasupat},
  \bibinfo{author}{P.~Liang},
\newblock \bibinfo{title}{Macro grammars and holistic triggering for efficient
  semantic parsing},
\newblock in: \bibinfo{editor}{M.~Palmer}, \bibinfo{editor}{R.~Hwa},
  \bibinfo{editor}{S.~Riedel} (Eds.), \bibinfo{booktitle}{Proceedings of the
  2017 Conference on Empirical Methods in Natural Language Processing, {EMNLP}
  2017, Copenhagen, Denmark, September 9-11, 2017},
  \bibinfo{publisher}{Association for Computational Linguistics},
  \bibinfo{year}{2017}, pp. \bibinfo{pages}{1214--1223}. \URLprefix
  \url{https://doi.org/10.18653/v1/d17-1125}.
  \DOIprefix\doi{10.18653/v1/d17-1125}.
\bibitem[{Misra et~al.(2018)Misra, Chang, He, and
  Yih}]{DBLP:conf/emnlp/MisraC0Y18}
\bibinfo{author}{D.~Misra}, \bibinfo{author}{M.~Chang},
  \bibinfo{author}{X.~He}, \bibinfo{author}{W.~Yih},
\newblock \bibinfo{title}{Policy shaping and generalized update equations for
  semantic parsing from denotations},
\newblock in: \bibinfo{editor}{E.~Riloff}, \bibinfo{editor}{D.~Chiang},
  \bibinfo{editor}{J.~Hockenmaier}, \bibinfo{editor}{J.~Tsujii} (Eds.),
  \bibinfo{booktitle}{Proceedings of the 2018 Conference on Empirical Methods
  in Natural Language Processing, Brussels, Belgium, October 31 - November 4,
  2018}, \bibinfo{publisher}{Association for Computational Linguistics},
  \bibinfo{year}{2018}, pp. \bibinfo{pages}{2442--2452}. \URLprefix
  \url{https://doi.org/10.18653/v1/d18-1266}.
  \DOIprefix\doi{10.18653/v1/d18-1266}.
\bibitem[{Guo et~al.(2021)Guo, Lou, Liu, and Zhang}]{DBLP:conf/emnlp/GuoLLZ21}
\bibinfo{author}{J.~Guo}, \bibinfo{author}{J.~Lou}, \bibinfo{author}{T.~Liu},
  \bibinfo{author}{D.~Zhang},
\newblock \bibinfo{title}{Weakly supervised semantic parsing by learning from
  mistakes},
\newblock in: \bibinfo{editor}{M.~Moens}, \bibinfo{editor}{X.~Huang},
  \bibinfo{editor}{L.~Specia}, \bibinfo{editor}{S.~W. Yih} (Eds.),
  \bibinfo{booktitle}{Findings of the Association for Computational
  Linguistics: {EMNLP} 2021, Virtual Event / Punta Cana, Dominican Republic,
  16-20 November, 2021}, \bibinfo{publisher}{Association for Computational
  Linguistics}, \bibinfo{year}{2021}, pp. \bibinfo{pages}{2603--2617}.
  \URLprefix \url{https://doi.org/10.18653/v1/2021.findings-emnlp.222}.
  \DOIprefix\doi{10.18653/v1/2021.findings-emnlp.222}.
\bibitem[{Wolfson et~al.(2022)Wolfson, Deutch, and
  Berant}]{DBLP:conf/naacl/WolfsonDB22}
\bibinfo{author}{T.~Wolfson}, \bibinfo{author}{D.~Deutch},
  \bibinfo{author}{J.~Berant},
\newblock \bibinfo{title}{Weakly supervised text-to-sql parsing through
  question decomposition},
\newblock in: \bibinfo{editor}{M.~Carpuat}, \bibinfo{editor}{M.~de~Marneffe},
  \bibinfo{editor}{I.~V.~M. Ru{\'{\i}}z} (Eds.), \bibinfo{booktitle}{Findings
  of the Association for Computational Linguistics: {NAACL} 2022, Seattle, WA,
  United States, July 10-15, 2022}, \bibinfo{publisher}{Association for
  Computational Linguistics}, \bibinfo{year}{2022}, pp.
  \bibinfo{pages}{2528--2542}. \URLprefix
  \url{https://doi.org/10.18653/v1/2022.findings-naacl.193}.
  \DOIprefix\doi{10.18653/V1/2022.FINDINGS-NAACL.193}.
\bibitem[{Auer et~al.(2007)Auer, Bizer, Kobilarov, Lehmann, Cyganiak, and
  Ives}]{DBLP:conf/semweb/AuerBKLCI07}
\bibinfo{author}{S.~Auer}, \bibinfo{author}{C.~Bizer},
  \bibinfo{author}{G.~Kobilarov}, \bibinfo{author}{J.~Lehmann},
  \bibinfo{author}{R.~Cyganiak}, \bibinfo{author}{Z.~G. Ives},
\newblock \bibinfo{title}{Dbpedia: {A} nucleus for a web of open data},
\newblock in: \bibinfo{editor}{K.~Aberer}, \bibinfo{editor}{K.~Choi},
  \bibinfo{editor}{N.~F. Noy}, \bibinfo{editor}{D.~Allemang},
  \bibinfo{editor}{K.~Lee}, \bibinfo{editor}{L.~J.~B. Nixon},
  \bibinfo{editor}{J.~Golbeck}, \bibinfo{editor}{P.~Mika},
  \bibinfo{editor}{D.~Maynard}, \bibinfo{editor}{R.~Mizoguchi},
  \bibinfo{editor}{G.~Schreiber}, \bibinfo{editor}{P.~Cudr{\'{e}}{-}Mauroux}
  (Eds.), \bibinfo{booktitle}{The Semantic Web, 6th International Semantic Web
  Conference, 2nd Asian Semantic Web Conference, {ISWC} 2007 + {ASWC} 2007,
  Busan, Korea, November 11-15, 2007}, volume \bibinfo{volume}{4825} of
  \textit{\bibinfo{series}{Lecture Notes in Computer Science}},
  \bibinfo{publisher}{Springer}, \bibinfo{year}{2007}, pp.
  \bibinfo{pages}{722--735}. \URLprefix
  \url{https://doi.org/10.1007/978-3-540-76298-0\_52}.
  \DOIprefix\doi{10.1007/978-3-540-76298-0\_52}.
\bibitem[{Bollacker et~al.(2008)Bollacker, Evans, Paritosh, Sturge, and
  Taylor}]{DBLP:conf/sigmod/BollackerEPST08}
\bibinfo{author}{K.~D. Bollacker}, \bibinfo{author}{C.~Evans},
  \bibinfo{author}{P.~K. Paritosh}, \bibinfo{author}{T.~Sturge},
  \bibinfo{author}{J.~Taylor},
\newblock \bibinfo{title}{Freebase: a collaboratively created graph database
  for structuring human knowledge},
\newblock in: \bibinfo{editor}{J.~T. Wang} (Ed.),
  \bibinfo{booktitle}{Proceedings of the {ACM} {SIGMOD} International
  Conference on Management of Data, {SIGMOD} 2008, Vancouver, BC, Canada, June
  10-12, 2008}, \bibinfo{publisher}{{ACM}}, \bibinfo{year}{2008}, pp.
  \bibinfo{pages}{1247--1250}. \URLprefix
  \url{https://doi.org/10.1145/1376616.1376746}.
  \DOIprefix\doi{10.1145/1376616.1376746}.
\bibitem[{Vrandecic and Kr{\"{o}}tzsch(2014)}]{DBLP:journals/cacm/VrandecicK14}
\bibinfo{author}{D.~Vrandecic}, \bibinfo{author}{M.~Kr{\"{o}}tzsch},
\newblock \bibinfo{title}{Wikidata: a free collaborative knowledgebase},
\newblock \bibinfo{journal}{Commun. {ACM}} \bibinfo{volume}{57}
  (\bibinfo{year}{2014}) \bibinfo{pages}{78--85}. \URLprefix
  \url{https://doi.org/10.1145/2629489}. \DOIprefix\doi{10.1145/2629489}.
\bibitem[{Yih et~al.(2015)Yih, Chang, He, and Gao}]{DBLP:conf/acl/YihCHG15}
\bibinfo{author}{W.~Yih}, \bibinfo{author}{M.~Chang}, \bibinfo{author}{X.~He},
  \bibinfo{author}{J.~Gao},
\newblock \bibinfo{title}{Semantic parsing via staged query graph generation:
  Question answering with knowledge base},
\newblock in: \bibinfo{booktitle}{Proceedings of the 53rd Annual Meeting of the
  Association for Computational Linguistics and the 7th International Joint
  Conference on Natural Language Processing of the Asian Federation of Natural
  Language Processing, {ACL} 2015, July 26-31, 2015, Beijing, China, Volume 1:
  Long Papers}, \bibinfo{publisher}{The Association for Computer Linguistics},
  \bibinfo{year}{2015}, pp. \bibinfo{pages}{1321--1331}. \URLprefix
  \url{https://doi.org/10.3115/v1/p15-1128}.
  \DOIprefix\doi{10.3115/v1/p15-1128}.
\bibitem[{Bordes et~al.(2014)Bordes, Chopra, and
  Weston}]{DBLP:conf/emnlp/BordesCW14}
\bibinfo{author}{A.~Bordes}, \bibinfo{author}{S.~Chopra},
  \bibinfo{author}{J.~Weston},
\newblock \bibinfo{title}{Question answering with subgraph embeddings},
\newblock in: \bibinfo{editor}{A.~Moschitti}, \bibinfo{editor}{B.~Pang},
  \bibinfo{editor}{W.~Daelemans} (Eds.), \bibinfo{booktitle}{Proceedings of the
  2014 Conference on Empirical Methods in Natural Language Processing, {EMNLP}
  2014, October 25-29, 2014, Doha, Qatar, {A} meeting of SIGDAT, a Special
  Interest Group of the {ACL}}, \bibinfo{publisher}{{ACL}},
  \bibinfo{year}{2014}, pp. \bibinfo{pages}{615--620}. \URLprefix
  \url{https://doi.org/10.3115/v1/d14-1067}.
  \DOIprefix\doi{10.3115/v1/d14-1067}.
\bibitem[{Dong et~al.(2015)Dong, Wei, Zhou, and Xu}]{DBLP:conf/acl/DongWZX15}
\bibinfo{author}{L.~Dong}, \bibinfo{author}{F.~Wei}, \bibinfo{author}{M.~Zhou},
  \bibinfo{author}{K.~Xu},
\newblock \bibinfo{title}{Question answering over freebase with multi-column
  convolutional neural networks},
\newblock in: \bibinfo{booktitle}{Proceedings of the 53rd Annual Meeting of the
  Association for Computational Linguistics and the 7th International Joint
  Conference on Natural Language Processing of the Asian Federation of Natural
  Language Processing, {ACL} 2015, July 26-31, 2015, Beijing, China, Volume 1:
  Long Papers}, \bibinfo{publisher}{The Association for Computer Linguistics},
  \bibinfo{year}{2015}, pp. \bibinfo{pages}{260--269}. \URLprefix
  \url{https://doi.org/10.3115/v1/p15-1026}.
  \DOIprefix\doi{10.3115/v1/p15-1026}.
\bibitem[{Fang et~al.(2024)Fang, Zhu, and Gurevych}]{DBLP:conf/acl/FangZG24}
\bibinfo{author}{H.~Fang}, \bibinfo{author}{X.~Zhu},
  \bibinfo{author}{I.~Gurevych},
\newblock \bibinfo{title}{{DARA:} decomposition-alignment-reasoning autonomous
  language agent for question answering over knowledge graphs},
\newblock in: \bibinfo{editor}{L.~Ku}, \bibinfo{editor}{A.~Martins},
  \bibinfo{editor}{V.~Srikumar} (Eds.), \bibinfo{booktitle}{Findings of the
  Association for Computational Linguistics, {ACL} 2024, Bangkok, Thailand and
  virtual meeting, August 11-16, 2024}, \bibinfo{publisher}{Association for
  Computational Linguistics}, \bibinfo{year}{2024}, pp.
  \bibinfo{pages}{3406--3432}. \URLprefix
  \url{https://doi.org/10.18653/v1/2024.findings-acl.203}.
  \DOIprefix\doi{10.18653/V1/2024.FINDINGS-ACL.203}.
\bibitem[{Zhang et~al.(2022)Zhang, Zhang, Yu, Tang, Tang, Li, and
  Chen}]{DBLP:conf/acl/ZhangZY000C22}
\bibinfo{author}{J.~Zhang}, \bibinfo{author}{X.~Zhang},
  \bibinfo{author}{J.~Yu}, \bibinfo{author}{J.~Tang},
  \bibinfo{author}{J.~Tang}, \bibinfo{author}{C.~Li},
  \bibinfo{author}{H.~Chen},
\newblock \bibinfo{title}{Subgraph retrieval enhanced model for multi-hop
  knowledge base question answering},
\newblock in: \bibinfo{editor}{S.~Muresan}, \bibinfo{editor}{P.~Nakov},
  \bibinfo{editor}{A.~Villavicencio} (Eds.), \bibinfo{booktitle}{Proceedings of
  the 60th Annual Meeting of the Association for Computational Linguistics
  (Volume 1: Long Papers), {ACL} 2022, Dublin, Ireland, May 22-27, 2022},
  \bibinfo{publisher}{Association for Computational Linguistics},
  \bibinfo{year}{2022}, pp. \bibinfo{pages}{5773--5784}. \URLprefix
  \url{https://doi.org/10.18653/v1/2022.acl-long.396}.
  \DOIprefix\doi{10.18653/V1/2022.ACL-LONG.396}.
\bibitem[{Luo et~al.(2024)Luo, Zhao, Gong, Haffari, and
  Pan}]{DBLP:journals/corr/abs-2410-13080}
\bibinfo{author}{L.~Luo}, \bibinfo{author}{Z.~Zhao}, \bibinfo{author}{C.~Gong},
  \bibinfo{author}{G.~Haffari}, \bibinfo{author}{S.~Pan},
\newblock \bibinfo{title}{Graph-constrained reasoning: Faithful reasoning on
  knowledge graphs with large language models},
\newblock \bibinfo{journal}{CoRR} \bibinfo{volume}{abs/2410.13080}
  (\bibinfo{year}{2024}). \URLprefix
  \url{https://doi.org/10.48550/arXiv.2410.13080}.
  \DOIprefix\doi{10.48550/ARXIV.2410.13080}.
  \href{http://arxiv.org/abs/2410.13080}{{\tt arXiv:2410.13080}}.
\bibitem[{Sun et~al.(2019)Sun, Bedrax{-}Weiss, and
  Cohen}]{DBLP:conf/emnlp/SunBC19}
\bibinfo{author}{H.~Sun}, \bibinfo{author}{T.~Bedrax{-}Weiss},
  \bibinfo{author}{W.~W. Cohen},
\newblock \bibinfo{title}{Pullnet: Open domain question answering with
  iterative retrieval on knowledge bases and text},
\newblock in: \bibinfo{editor}{K.~Inui}, \bibinfo{editor}{J.~Jiang},
  \bibinfo{editor}{V.~Ng}, \bibinfo{editor}{X.~Wan} (Eds.),
  \bibinfo{booktitle}{Proceedings of the 2019 Conference on Empirical Methods
  in Natural Language Processing and the 9th International Joint Conference on
  Natural Language Processing, {EMNLP-IJCNLP} 2019, Hong Kong, China, November
  3-7, 2019}, \bibinfo{publisher}{Association for Computational Linguistics},
  \bibinfo{year}{2019}, pp. \bibinfo{pages}{2380--2390}. \URLprefix
  \url{https://doi.org/10.18653/v1/D19-1242}.
  \DOIprefix\doi{10.18653/V1/D19-1242}.
\bibitem[{Choi et~al.(2023)Choi, Lee, Chu, and Kim}]{DBLP:conf/nips/ChoiLCK23}
\bibinfo{author}{H.~K. Choi}, \bibinfo{author}{S.~Lee},
  \bibinfo{author}{J.~Chu}, \bibinfo{author}{H.~J. Kim},
\newblock \bibinfo{title}{Nutrea: Neural tree search for context-guided
  multi-hop {KGQA}},
\newblock in: \bibinfo{editor}{A.~Oh}, \bibinfo{editor}{T.~Naumann},
  \bibinfo{editor}{A.~Globerson}, \bibinfo{editor}{K.~Saenko},
  \bibinfo{editor}{M.~Hardt}, \bibinfo{editor}{S.~Levine} (Eds.),
  \bibinfo{booktitle}{Advances in Neural Information Processing Systems 36:
  Annual Conference on Neural Information Processing Systems 2023, NeurIPS
  2023, New Orleans, LA, USA, December 10 - 16, 2023}, \bibinfo{year}{2023}.
  \URLprefix
  \url{http://papers.nips.cc/paper\_files/paper/2023/hash/707a2d58641b2192203b4bf4c532cfe1-Abstract-Conference.html}.
\bibitem[{Jiang et~al.(2023)Jiang, Zhou, Zhao, and
  Wen}]{DBLP:conf/iclr/JiangZ0W23}
\bibinfo{author}{J.~Jiang}, \bibinfo{author}{K.~Zhou},
  \bibinfo{author}{X.~Zhao}, \bibinfo{author}{J.~Wen},
\newblock \bibinfo{title}{Unikgqa: Unified retrieval and reasoning for solving
  multi-hop question answering over knowledge graph},
\newblock in: \bibinfo{booktitle}{The Eleventh International Conference on
  Learning Representations, {ICLR} 2023, Kigali, Rwanda, May 1-5, 2023},
  \bibinfo{publisher}{OpenReview.net}, \bibinfo{year}{2023}. \URLprefix
  \url{https://openreview.net/forum?id=Z63RvyAZ2Vh}.
\bibitem[{Luo et~al.(2024)Luo, Li, Haffari, and Pan}]{DBLP:conf/iclr/LuoLHP24}
\bibinfo{author}{L.~Luo}, \bibinfo{author}{Y.~Li},
  \bibinfo{author}{G.~Haffari}, \bibinfo{author}{S.~Pan},
\newblock \bibinfo{title}{Reasoning on graphs: Faithful and interpretable large
  language model reasoning},
\newblock in: \bibinfo{booktitle}{The Twelfth International Conference on
  Learning Representations, {ICLR} 2024, Vienna, Austria, May 7-11, 2024},
  \bibinfo{publisher}{OpenReview.net}, \bibinfo{year}{2024}. \URLprefix
  \url{https://openreview.net/forum?id=ZGNWW7xZ6Q}.
\bibitem[{Sun et~al.(2024)Sun, Xu, Tang, Wang, Lin, Gong, Ni, Shum, and
  Guo}]{DBLP:conf/iclr/SunXTW0GNSG24}
\bibinfo{author}{J.~Sun}, \bibinfo{author}{C.~Xu}, \bibinfo{author}{L.~Tang},
  \bibinfo{author}{S.~Wang}, \bibinfo{author}{C.~Lin},
  \bibinfo{author}{Y.~Gong}, \bibinfo{author}{L.~M. Ni},
  \bibinfo{author}{H.~Shum}, \bibinfo{author}{J.~Guo},
\newblock \bibinfo{title}{Think-on-graph: Deep and responsible reasoning of
  large language model on knowledge graph},
\newblock in: \bibinfo{booktitle}{The Twelfth International Conference on
  Learning Representations, {ICLR} 2024, Vienna, Austria, May 7-11, 2024},
  \bibinfo{publisher}{OpenReview.net}, \bibinfo{year}{2024}. \URLprefix
  \url{https://openreview.net/forum?id=nnVO1PvbTv}.
\bibitem[{Sui et~al.(2025)Sui, He, Liu, He, Wang, and
  Hooi}]{DBLP:conf/acl/SuiHLHWH25}
\bibinfo{author}{Y.~Sui}, \bibinfo{author}{Y.~He}, \bibinfo{author}{N.~Liu},
  \bibinfo{author}{X.~He}, \bibinfo{author}{K.~Wang},
  \bibinfo{author}{B.~Hooi},
\newblock \bibinfo{title}{Fidelis: Faithful reasoning in large language models
  for knowledge graph question answering},
\newblock in: \bibinfo{editor}{W.~Che}, \bibinfo{editor}{J.~Nabende},
  \bibinfo{editor}{E.~Shutova}, \bibinfo{editor}{M.~T. Pilehvar} (Eds.),
  \bibinfo{booktitle}{Findings of the Association for Computational
  Linguistics, {ACL} 2025, Vienna, Austria, July 27 - August 1, 2025},
  \bibinfo{publisher}{Association for Computational Linguistics},
  \bibinfo{year}{2025}, pp. \bibinfo{pages}{8315--8330}. \URLprefix
  \url{https://aclanthology.org/2025.findings-acl.436/}.
\bibitem[{Ma et~al.(2025)Ma, Gao, Chai, Sun, Wang, Pei, Tao, Song, Liu, Zhang,
  and Cui}]{DBLP:conf/aaai/Ma0CSWPTSLZC25}
\bibinfo{author}{J.~Ma}, \bibinfo{author}{Z.~Gao}, \bibinfo{author}{Q.~Chai},
  \bibinfo{author}{W.~Sun}, \bibinfo{author}{P.~Wang},
  \bibinfo{author}{H.~Pei}, \bibinfo{author}{J.~Tao},
  \bibinfo{author}{L.~Song}, \bibinfo{author}{J.~Liu},
  \bibinfo{author}{C.~Zhang}, \bibinfo{author}{L.~Cui},
\newblock \bibinfo{title}{Debate on graph: {A} flexible and reliable reasoning
  framework for large language models},
\newblock in: \bibinfo{editor}{T.~Walsh}, \bibinfo{editor}{J.~Shah},
  \bibinfo{editor}{Z.~Kolter} (Eds.), \bibinfo{booktitle}{AAAI-25, Sponsored by
  the Association for the Advancement of Artificial Intelligence, February 25 -
  March 4, 2025, Philadelphia, PA, {USA}}, \bibinfo{publisher}{{AAAI} Press},
  \bibinfo{year}{2025}, pp. \bibinfo{pages}{24768--24776}. \URLprefix
  \url{https://doi.org/10.1609/aaai.v39i23.34658}.
  \DOIprefix\doi{10.1609/AAAI.V39I23.34658}.
\bibitem[{Jiang et~al.(2025)Jiang, Zhou, Zhao, Song, Zhu, Zhu, and
  Wen}]{DBLP:conf/acl/JiangZZS0ZW25}
\bibinfo{author}{J.~Jiang}, \bibinfo{author}{K.~Zhou},
  \bibinfo{author}{X.~Zhao}, \bibinfo{author}{Y.~Song},
  \bibinfo{author}{C.~Zhu}, \bibinfo{author}{H.~Zhu}, \bibinfo{author}{J.~Wen},
\newblock \bibinfo{title}{Kg-agent: An efficient autonomous agent framework for
  complex reasoning over knowledge graph},
\newblock in: \bibinfo{editor}{W.~Che}, \bibinfo{editor}{J.~Nabende},
  \bibinfo{editor}{E.~Shutova}, \bibinfo{editor}{M.~T. Pilehvar} (Eds.),
  \bibinfo{booktitle}{Proceedings of the 63rd Annual Meeting of the Association
  for Computational Linguistics (Volume 1: Long Papers), {ACL} 2025, Vienna,
  Austria, July 27 - August 1, 2025}, \bibinfo{publisher}{Association for
  Computational Linguistics}, \bibinfo{year}{2025}, pp.
  \bibinfo{pages}{9505--9523}. \URLprefix
  \url{https://aclanthology.org/2025.acl-long.468/}.
\bibitem[{Ma et~al.(2025)Ma, Xu, Jiang, Li, Qu, Yang, Mao, and
  Guo}]{DBLP:conf/iclr/MaXJLQYMG25}
\bibinfo{author}{S.~Ma}, \bibinfo{author}{C.~Xu}, \bibinfo{author}{X.~Jiang},
  \bibinfo{author}{M.~Li}, \bibinfo{author}{H.~Qu}, \bibinfo{author}{C.~Yang},
  \bibinfo{author}{J.~Mao}, \bibinfo{author}{J.~Guo},
\newblock \bibinfo{title}{Think-on-graph 2.0: Deep and faithful large language
  model reasoning with knowledge-guided retrieval augmented generation},
\newblock in: \bibinfo{booktitle}{The Thirteenth International Conference on
  Learning Representations, {ICLR} 2025, Singapore, April 24-28, 2025},
  \bibinfo{publisher}{OpenReview.net}, \bibinfo{year}{2025}. \URLprefix
  \url{https://openreview.net/forum?id=oFBu7qaZpS}.
\bibitem[{Liu et~al.(2025)Liu, Zhang, Lin, Yang, Peng, and
  Yin}]{DBLP:conf/www/LiuZLYPY25}
\bibinfo{author}{B.~Liu}, \bibinfo{author}{J.~Zhang}, \bibinfo{author}{F.~Lin},
  \bibinfo{author}{C.~Yang}, \bibinfo{author}{M.~Peng},
  \bibinfo{author}{W.~Yin},
\newblock \bibinfo{title}{Symagent: {A} neural-symbolic self-learning agent
  framework for complex reasoning over knowledge graphs},
\newblock in: \bibinfo{editor}{G.~Long}, \bibinfo{editor}{M.~Blumestein},
  \bibinfo{editor}{Y.~Chang}, \bibinfo{editor}{L.~Lewin{-}Eytan},
  \bibinfo{editor}{Z.~H. Huang}, \bibinfo{editor}{E.~Yom{-}Tov} (Eds.),
  \bibinfo{booktitle}{Proceedings of the {ACM} on Web Conference 2025, {WWW}
  2025, Sydney, NSW, Australia, 28 April 2025- 2 May 2025},
  \bibinfo{publisher}{{ACM}}, \bibinfo{year}{2025}, pp.
  \bibinfo{pages}{98--108}. \URLprefix
  \url{https://doi.org/10.1145/3696410.3714768}.
  \DOIprefix\doi{10.1145/3696410.3714768}.
\bibitem[{McCarthy(1978)}]{DBLP:conf/hopl/McCarthy78}
\bibinfo{author}{J.~McCarthy},
\newblock \bibinfo{title}{History of {LISP}},
\newblock in: \bibinfo{editor}{R.~L. Wexelblat} (Ed.),
  \bibinfo{booktitle}{History of Programming Languages, from the {ACM}
  {SIGPLAN} History of Programming Languages Conference, June 1-3, 1978, Los
  Angeles, California, {USA}}, \bibinfo{publisher}{Academic Press / {ACM}},
  \bibinfo{year}{1978}, pp. \bibinfo{pages}{173--185}. \URLprefix
  \url{https://doi.org/10.1145/800025.1198360}.
  \DOIprefix\doi{10.1145/800025.1198360}.
\bibitem[{Wolf et~al.(2020)Wolf, Debut, Sanh, Chaumond, Delangue, Moi, Cistac,
  Rault, Louf, Funtowicz, Davison, Shleifer, von Platen, Ma, Jernite, Plu, Xu,
  Scao, Gugger, Drame, Lhoest, and Rush}]{DBLP:conf/emnlp/WolfDSCDMCRLFDS20}
\bibinfo{author}{T.~Wolf}, \bibinfo{author}{L.~Debut},
  \bibinfo{author}{V.~Sanh}, \bibinfo{author}{J.~Chaumond},
  \bibinfo{author}{C.~Delangue}, \bibinfo{author}{A.~Moi},
  \bibinfo{author}{P.~Cistac}, \bibinfo{author}{T.~Rault},
  \bibinfo{author}{R.~Louf}, \bibinfo{author}{M.~Funtowicz},
  \bibinfo{author}{J.~Davison}, \bibinfo{author}{S.~Shleifer},
  \bibinfo{author}{P.~von Platen}, \bibinfo{author}{C.~Ma},
  \bibinfo{author}{Y.~Jernite}, \bibinfo{author}{J.~Plu},
  \bibinfo{author}{C.~Xu}, \bibinfo{author}{T.~L. Scao},
  \bibinfo{author}{S.~Gugger}, \bibinfo{author}{M.~Drame},
  \bibinfo{author}{Q.~Lhoest}, \bibinfo{author}{A.~M. Rush},
\newblock \bibinfo{title}{Transformers: State-of-the-art natural language
  processing},
\newblock in: \bibinfo{editor}{Q.~Liu}, \bibinfo{editor}{D.~Schlangen} (Eds.),
  \bibinfo{booktitle}{Proceedings of the 2020 Conference on Empirical Methods
  in Natural Language Processing: System Demonstrations, {EMNLP} 2020 - Demos,
  Online, November 16-20, 2020}, \bibinfo{publisher}{Association for
  Computational Linguistics}, \bibinfo{year}{2020}, pp.
  \bibinfo{pages}{38--45}. \URLprefix
  \url{https://doi.org/10.18653/v1/2020.emnlp-demos.6}.
  \DOIprefix\doi{10.18653/v1/2020.emnlp-demos.6}.
\bibitem[{Talmor and Berant(2018)}]{DBLP:conf/naacl/TalmorB18}
\bibinfo{author}{A.~Talmor}, \bibinfo{author}{J.~Berant},
\newblock \bibinfo{title}{The web as a knowledge-base for answering complex
  questions},
\newblock in: \bibinfo{editor}{M.~A. Walker}, \bibinfo{editor}{H.~Ji},
  \bibinfo{editor}{A.~Stent} (Eds.), \bibinfo{booktitle}{Proceedings of the
  2018 Conference of the North American Chapter of the Association for
  Computational Linguistics: Human Language Technologies, {NAACL-HLT} 2018, New
  Orleans, Louisiana, USA, June 1-6, 2018, Volume 1 (Long Papers)},
  \bibinfo{publisher}{Association for Computational Linguistics},
  \bibinfo{year}{2018}, pp. \bibinfo{pages}{641--651}. \URLprefix
  \url{https://doi.org/10.18653/v1/n18-1059}.
  \DOIprefix\doi{10.18653/v1/n18-1059}.
\bibitem[{Gu et~al.(2021)Gu, Kase, Vanni, Sadler, Liang, Yan, and
  Su}]{DBLP:conf/www/GuKVSLY021}
\bibinfo{author}{Y.~Gu}, \bibinfo{author}{S.~Kase}, \bibinfo{author}{M.~Vanni},
  \bibinfo{author}{B.~M. Sadler}, \bibinfo{author}{P.~Liang},
  \bibinfo{author}{X.~Yan}, \bibinfo{author}{Y.~Su},
\newblock \bibinfo{title}{Beyond {I.I.D.:} three levels of generalization for
  question answering on knowledge bases},
\newblock in: \bibinfo{editor}{J.~Leskovec}, \bibinfo{editor}{M.~Grobelnik},
  \bibinfo{editor}{M.~Najork}, \bibinfo{editor}{J.~Tang},
  \bibinfo{editor}{L.~Zia} (Eds.), \bibinfo{booktitle}{{WWW} '21: The Web
  Conference 2021, Virtual Event / Ljubljana, Slovenia, April 19-23, 2021},
  \bibinfo{publisher}{{ACM} / {IW3C2}}, \bibinfo{year}{2021}, pp.
  \bibinfo{pages}{3477--3488}. \URLprefix
  \url{https://doi.org/10.1145/3442381.3449992}.
  \DOIprefix\doi{10.1145/3442381.3449992}.
\bibitem[{Jia et~al.(2021)Jia, Pramanik, Roy, and
  Weikum}]{DBLP:conf/cikm/JiaPRW21}
\bibinfo{author}{Z.~Jia}, \bibinfo{author}{S.~Pramanik}, \bibinfo{author}{R.~S.
  Roy}, \bibinfo{author}{G.~Weikum},
\newblock \bibinfo{title}{Complex temporal question answering on knowledge
  graphs},
\newblock in: \bibinfo{editor}{G.~Demartini}, \bibinfo{editor}{G.~Zuccon},
  \bibinfo{editor}{J.~S. Culpepper}, \bibinfo{editor}{Z.~Huang},
  \bibinfo{editor}{H.~Tong} (Eds.), \bibinfo{booktitle}{{CIKM} '21: The 30th
  {ACM} International Conference on Information and Knowledge Management,
  Virtual Event, Queensland, Australia, November 1 - 5, 2021},
  \bibinfo{publisher}{{ACM}}, \bibinfo{year}{2021}, pp.
  \bibinfo{pages}{792--802}. \URLprefix
  \url{https://doi.org/10.1145/3459637.3482416}.
  \DOIprefix\doi{10.1145/3459637.3482416}.
\bibitem[{Liang(2013)}]{DBLP:journals/corr/Liang13}
\bibinfo{author}{P.~Liang},
\newblock \bibinfo{title}{Lambda dependency-based compositional semantics},
\newblock \bibinfo{journal}{CoRR} \bibinfo{volume}{abs/1309.4408}
  (\bibinfo{year}{2013}). \URLprefix \url{http://arxiv.org/abs/1309.4408}.
  \href{http://arxiv.org/abs/1309.4408}{{\tt arXiv:1309.4408}}.
\bibitem[{Nie et~al.(2022)Nie, Cao, Shi, Sun, Tian, Hou, Li, and
  Zhai}]{DBLP:conf/emnlp/NieCSST0LZ22}
\bibinfo{author}{L.~Nie}, \bibinfo{author}{S.~Cao}, \bibinfo{author}{J.~Shi},
  \bibinfo{author}{J.~Sun}, \bibinfo{author}{Q.~Tian},
  \bibinfo{author}{L.~Hou}, \bibinfo{author}{J.~Li}, \bibinfo{author}{J.~Zhai},
\newblock \bibinfo{title}{Graphq {IR:} unifying the semantic parsing of graph
  query languages with one intermediate representation},
\newblock in: \bibinfo{editor}{Y.~Goldberg}, \bibinfo{editor}{Z.~Kozareva},
  \bibinfo{editor}{Y.~Zhang} (Eds.), \bibinfo{booktitle}{Proceedings of the
  2022 Conference on Empirical Methods in Natural Language Processing, {EMNLP}
  2022, Abu Dhabi, United Arab Emirates, December 7-11, 2022},
  \bibinfo{publisher}{Association for Computational Linguistics},
  \bibinfo{year}{2022}, pp. \bibinfo{pages}{5848--5865}. \URLprefix
  \url{https://aclanthology.org/2022.emnlp-main.394}.
\bibitem[{Nie et~al.(2023)Nie, Sun, Wang, Du, Han, Zhang, Hou, Li, and
  Zhai}]{DBLP:conf/aaai/NieS0DH00LZ23}
\bibinfo{author}{L.~Nie}, \bibinfo{author}{J.~Sun}, \bibinfo{author}{Y.~Wang},
  \bibinfo{author}{L.~Du}, \bibinfo{author}{S.~Han},
  \bibinfo{author}{D.~Zhang}, \bibinfo{author}{L.~Hou},
  \bibinfo{author}{J.~Li}, \bibinfo{author}{J.~Zhai},
\newblock \bibinfo{title}{Unveiling the black box of plms with semantic
  anchors: Towards interpretable neural semantic parsing},
\newblock in: \bibinfo{editor}{B.~Williams}, \bibinfo{editor}{Y.~Chen},
  \bibinfo{editor}{J.~Neville} (Eds.), \bibinfo{booktitle}{Thirty-Seventh
  {AAAI} Conference on Artificial Intelligence, {AAAI} 2023, Thirty-Fifth
  Conference on Innovative Applications of Artificial Intelligence, {IAAI}
  2023, Thirteenth Symposium on Educational Advances in Artificial
  Intelligence, {EAAI} 2023, Washington, DC, USA, February 7-14, 2023},
  \bibinfo{publisher}{{AAAI} Press}, \bibinfo{year}{2023}, pp.
  \bibinfo{pages}{13400--13408}. \URLprefix
  \url{https://ojs.aaai.org/index.php/AAAI/article/view/26572}.
\bibitem[{Odendahl(2022)}]{Hissp_version:0.3.0}
\bibinfo{author}{M.~E. Odendahl}, \bibinfo{title}{{Hissp}},
  \bibinfo{howpublished}{\url{https://pypi.org/project/hissp/0.3.0/}},
  \bibinfo{year}{2022}.
\bibitem[{Gugger et~al.(2022)Gugger, Debut, Wolf, Schmid, Mueller, Mangrulkar,
  Sun, and Bossan}]{accelerate}
\bibinfo{author}{S.~Gugger}, \bibinfo{author}{L.~Debut},
  \bibinfo{author}{T.~Wolf}, \bibinfo{author}{P.~Schmid},
  \bibinfo{author}{Z.~Mueller}, \bibinfo{author}{S.~Mangrulkar},
  \bibinfo{author}{M.~Sun}, \bibinfo{author}{B.~Bossan},
  \bibinfo{title}{Accelerate: Training and inference at scale made simple,
  efficient and adaptable.},
  \bibinfo{howpublished}{\url{https://github.com/huggingface/accelerate}},
  \bibinfo{year}{2022}.
\bibitem[{Loshchilov and Hutter(2017)}]{DBLP:journals/corr/abs-1711-05101}
\bibinfo{author}{I.~Loshchilov}, \bibinfo{author}{F.~Hutter},
\newblock \bibinfo{title}{Fixing weight decay regularization in adam},
\newblock \bibinfo{journal}{CoRR} \bibinfo{volume}{abs/1711.05101}
  (\bibinfo{year}{2017}). \URLprefix \url{http://arxiv.org/abs/1711.05101}.
  \href{http://arxiv.org/abs/1711.05101}{{\tt arXiv:1711.05101}}.
\bibitem[{Herzig and Berant(2017)}]{DBLP:conf/acl/HerzigB17}
\bibinfo{author}{J.~Herzig}, \bibinfo{author}{J.~Berant},
\newblock \bibinfo{title}{Neural semantic parsing over multiple
  knowledge-bases},
\newblock in: \bibinfo{editor}{R.~Barzilay}, \bibinfo{editor}{M.~Kan} (Eds.),
  \bibinfo{booktitle}{Proceedings of the 55th Annual Meeting of the Association
  for Computational Linguistics, {ACL} 2017, Vancouver, Canada, July 30 -
  August 4, Volume 2: Short Papers}, \bibinfo{publisher}{Association for
  Computational Linguistics}, \bibinfo{year}{2017}, pp.
  \bibinfo{pages}{623--628}. \URLprefix
  \url{https://doi.org/10.18653/v1/P17-2098}.
  \DOIprefix\doi{10.18653/V1/P17-2098}.
\bibitem[{Su and Yan(2017)}]{DBLP:conf/emnlp/SuY17}
\bibinfo{author}{Y.~Su}, \bibinfo{author}{X.~Yan},
\newblock \bibinfo{title}{Cross-domain semantic parsing via paraphrasing},
\newblock in: \bibinfo{editor}{M.~Palmer}, \bibinfo{editor}{R.~Hwa},
  \bibinfo{editor}{S.~Riedel} (Eds.), \bibinfo{booktitle}{Proceedings of the
  2017 Conference on Empirical Methods in Natural Language Processing, {EMNLP}
  2017, Copenhagen, Denmark, September 9-11, 2017},
  \bibinfo{publisher}{Association for Computational Linguistics},
  \bibinfo{year}{2017}, pp. \bibinfo{pages}{1235--1246}. \URLprefix
  \url{https://doi.org/10.18653/v1/d17-1127}.
  \DOIprefix\doi{10.18653/V1/D17-1127}.
\bibitem[{Chen et~al.(2018)Chen, Sun, and Han}]{DBLP:conf/acl/SunHC18}
\bibinfo{author}{B.~Chen}, \bibinfo{author}{L.~Sun}, \bibinfo{author}{X.~Han},
\newblock \bibinfo{title}{Sequence-to-action: End-to-end semantic graph
  generation for semantic parsing},
\newblock in: \bibinfo{editor}{I.~Gurevych}, \bibinfo{editor}{Y.~Miyao} (Eds.),
  \bibinfo{booktitle}{Proceedings of the 56th Annual Meeting of the Association
  for Computational Linguistics, {ACL} 2018, Melbourne, Australia, July 15-20,
  2018, Volume 1: Long Papers}, \bibinfo{publisher}{Association for
  Computational Linguistics}, \bibinfo{year}{2018}, pp.
  \bibinfo{pages}{766--777}. \URLprefix
  \url{https://aclanthology.org/P18-1071/}.
  \DOIprefix\doi{10.18653/v1/P18-1071}.
\bibitem[{Cao et~al.(2019)Cao, Zhu, Liu, Li, and Yu}]{DBLP:conf/acl/CaoZLLY19}
\bibinfo{author}{R.~Cao}, \bibinfo{author}{S.~Zhu}, \bibinfo{author}{C.~Liu},
  \bibinfo{author}{J.~Li}, \bibinfo{author}{K.~Yu},
\newblock \bibinfo{title}{Semantic parsing with dual learning},
\newblock in: \bibinfo{editor}{A.~Korhonen}, \bibinfo{editor}{D.~R. Traum},
  \bibinfo{editor}{L.~M{\`{a}}rquez} (Eds.), \bibinfo{booktitle}{Proceedings of
  the 57th Conference of the Association for Computational Linguistics, {ACL}
  2019, Florence, Italy, July 28- August 2, 2019, Volume 1: Long Papers},
  \bibinfo{publisher}{Association for Computational Linguistics},
  \bibinfo{year}{2019}, pp. \bibinfo{pages}{51--64}. \URLprefix
  \url{https://doi.org/10.18653/v1/p19-1007}.
  \DOIprefix\doi{10.18653/v1/p19-1007}.
\bibitem[{Cao et~al.(2020)Cao, Zhu, Yang, Liu, Ma, Zhao, Chen, and
  Yu}]{DBLP:conf/acl/CaoZYLMZCY20}
\bibinfo{author}{R.~Cao}, \bibinfo{author}{S.~Zhu}, \bibinfo{author}{C.~Yang},
  \bibinfo{author}{C.~Liu}, \bibinfo{author}{R.~Ma}, \bibinfo{author}{Y.~Zhao},
  \bibinfo{author}{L.~Chen}, \bibinfo{author}{K.~Yu},
\newblock \bibinfo{title}{Unsupervised dual paraphrasing for two-stage semantic
  parsing},
\newblock in: \bibinfo{editor}{D.~Jurafsky}, \bibinfo{editor}{J.~Chai},
  \bibinfo{editor}{N.~Schluter}, \bibinfo{editor}{J.~R. Tetreault} (Eds.),
  \bibinfo{booktitle}{Proceedings of the 58th Annual Meeting of the Association
  for Computational Linguistics, {ACL} 2020, Online, July 5-10, 2020},
  \bibinfo{publisher}{Association for Computational Linguistics},
  \bibinfo{year}{2020}, pp. \bibinfo{pages}{6806--6817}. \URLprefix
  \url{https://doi.org/10.18653/v1/2020.acl-main.608}.
  \DOIprefix\doi{10.18653/v1/2020.acl-main.608}.
\bibitem[{Agrawal et~al.(2019)Agrawal, Dalmia, Jain, Bansal, Mittal, and
  Sankaranarayanan}]{DBLP:conf/acl/AgrawalDJBMS19}
\bibinfo{author}{P.~Agrawal}, \bibinfo{author}{A.~Dalmia},
  \bibinfo{author}{P.~Jain}, \bibinfo{author}{A.~Bansal},
  \bibinfo{author}{A.~R. Mittal}, \bibinfo{author}{K.~Sankaranarayanan},
\newblock \bibinfo{title}{Unified semantic parsing with weak supervision},
\newblock in: \bibinfo{editor}{A.~Korhonen}, \bibinfo{editor}{D.~R. Traum},
  \bibinfo{editor}{L.~M{\`{a}}rquez} (Eds.), \bibinfo{booktitle}{Proceedings of
  the 57th Conference of the Association for Computational Linguistics, {ACL}
  2019, Florence, Italy, July 28- August 2, 2019, Volume 1: Long Papers},
  \bibinfo{publisher}{Association for Computational Linguistics},
  \bibinfo{year}{2019}, pp. \bibinfo{pages}{4801--4810}. \URLprefix
  \url{https://doi.org/10.18653/v1/p19-1473}.
  \DOIprefix\doi{10.18653/v1/p19-1473}.
\bibitem[{Pasupat and Liang(2016)}]{DBLP:conf/acl/PasupatL16}
\bibinfo{author}{P.~Pasupat}, \bibinfo{author}{P.~Liang},
\newblock \bibinfo{title}{Inferring logical forms from denotations},
\newblock in: \bibinfo{booktitle}{Proceedings of the 54th Annual Meeting of the
  Association for Computational Linguistics, {ACL} 2016, August 7-12, 2016,
  Berlin, Germany, Volume 1: Long Papers}, \bibinfo{publisher}{The Association
  for Computer Linguistics}, \bibinfo{year}{2016}. \URLprefix
  \url{https://doi.org/10.18653/v1/p16-1003}.
  \DOIprefix\doi{10.18653/v1/p16-1003}.

\end{thebibliography}
  \fi

\newcommand{\includingvitae}{\proofreading}
\if \includingvitae 1
  \clearpage
  \section*{Vitae}
  \input{vitae.txt}
  \clearpage
  \else
  \fi

\appendix

\clearpage


\appendixtitle
\\
\hfill
\\

\begin{revisiontwo}
  We present the details of our grammar in \refappendix{sec:grammar-details},
  an working example in \refappendix{sec:working-example},
  an implementation issue of search algorithm in \refappendix{sec:search-implementation},
  an alternative algorithm that increases the speed of constrained decoding in \refappendix{sec:alternative-decoding-speed-optimization},
  the derivation of \(\nabla_\theta \jmml(\theta \scbar \uttersc, \lfgd)\) in \refappendix{sec:derivation-of-grad},
  and qualitative analysis in \refappendix{sec:qualitative-analysis}.
\end{revisiontwo}

\appendixsection{Grammar Details} \label{sec:grammar-details}

\subsection{Types}

The node classes in our grammar have return types and parameter types as properties, and the types have sub-type relations \Crefp{sec:grammars-with-types} \Crefp{fig:kqapro-type-hierarchy,fig:overnight-type-hierarchy}.

\begingroup
\newcommand{\commoncaption}{
  The types that are underlined are used only to group their sub-types without being used as return types or parameter types of node classes; therefore, the underlined types do not affect the lengths of action sequences when sub-type inference is disabled.
}
\begin{figure}[h]            
%

  \tablefontsizethree
  \begin{center}
    \input{\figurepathkqapro-type-hierarchy.tex\inputfilesuffix}
    \caption{
      Type hierarchies for \kqapro.
      \commoncaption
    }
    \label{fig:kqapro-type-hierarchy}
  \end{center}
\end{figure}

\begin{figure}[h]            

  \tablefontsizethree
  \begin{center}
    \input{\figurepathovernight-type-hierarchy.tex\inputfilesuffix}

    \lffontsizecaption
    \caption{
      Type hierarchies for \overnight.
      A type can have multiple super-types as a union type; therefore, \(\lfexpr{\lftyp{object-list}}\) has \(\lfexpr{\lftyp{object-*}}\) and \(\lfexpr{\lftyp{result}}\) as super-types.
    }
    \label{fig:overnight-type-hierarchy}
  \end{center}
\end{figure}
\endgroup


\subsection{Logical Form Templates} \label{sec:lf-template}

\begin{table}[t]  
  \begin{center}
    \caption{
      Example of converting an intermediate representation to the corresponding logical form for \kqapro.
      For each row, a sub-intermediate representation written in bold with a specific color is converted to a sub-logical underlined with the color in the next row.
      The step \mbox{5 -- 6} makes no difference, as the node class \(\lfexpr{find}\) has the same expression as its intermediate representation and as its logical form.
    }

    \tablefontsizefour
    \lffontsizethree
    \input{\tablepathir-to-lf.tex\inputfilesuffix}

    \label{tab:ir-to-lf}
  \end{center}
\end{table}

\begin{revision}
  The time complexity of converting an intermediate representation \(\lfir{\actseq}\) to a logical form \(\lflf{\actseq}\) is \mbox{\(O(\setsize{\actseq} + \setsize{\lflf{\actseq}})\)} which depends on logical form templates.
  We use three kinds of logical form templates that (1) replace placeholders with sub-logical forms, that (2) concatenate \(\lfexpr{\nlt{ *}}\) nodes, and that (3) involve complex operations such as conditional statements.
  The first kind of template delays the concatenation between strings in the templates and sub-logical forms that replace placeholders until the bottom-up process \Crefp{tab:ir-to-lf} is finished, so the replacement time is \(O(\setsize{\actseq})\) and the concatenation time is \(O(\setsize{\lflf{\actseq}})\).
  The second kind of template can concatenate up to \(\setsize{\actseq}\) \(\lfexpr{\nlt{ *}}\) nodes, so the concatenation time is \(O(\setsize{\lflf{\actseq}})\).
  The third kind of template is designed to finish operations in constant time, so it does not affect the total time complexity.
\end{revision}

In the implementation of our grammar, (1) all actions have logical form templates that are closely related to executable code,
and (2) many actions have logical form templates for simplified expressions.
We use two keywords, \(\lfexpr{default}\) and \(\lfexpr{visual}\), to separate the two groups of logical form templates \Crefp{fig:grammar-code}.
For \kqapro, the \(\lfexpr{default}\) templates make logical forms that can be converted to the Python code for KBQA by using a transpiler \citep{Hissp_version:0.3.0}.
The Python code from a logical form is a function that takes a KB and return a denotation, where the function is a \(\lfexpr{lambda}\) expression and the KB is the argument \(\lfexpr{context}\).
For \overnight, the \(\lfexpr{default}\) templates make logical forms that are \lambdadcs expressions, which are executed by the SEMPRE library \citep{DBLP:conf/emnlp/BerantCFL13}.
In contrast, the \(\lfexpr{visual}\) templates make concise logical forms, which are presented throughout this paper.

\ignore{
  As a logical form is constructed from templates rather than directly from actions, the order of the parameter types of an action is customizable.
  We customize the order of the parameter types of an action to put off constructing a sub-intermediate representation that requires relatively many actions.
  For example, some actions, such as \(\lfexpr{filter-number}\), have \(\lfexpr{obj-entity}\) as a parameter type, and an intermediate representation for \(\lfexpr{obj-entity}\) has relatively many nodes; therefore, the actions put \(\lfexpr{obj-entity}\) as the last parameter type.
  The customized order of parameter types reduces the average difference in time steps between the actions that have parent-child relations; we assume that a seq2seq model benefits from the reduced distances between the actions, which are located in sequences.
}


\begin{figure}[t]  
  \tablefontsizeone
  \begin{center}
    \input{\figurepathgrammar-code.tex\inputfilesuffix}
    \lffontsizecaption
    \caption{
      Part of code for the grammar for \kqapro.
      In the code, an action for a node class is defined by \(\lfexpr{define-action}\).
      There are discrepancies in terminology between the code and our grammar;
      \(\lfexpr{act-type}\) means a return type, \(\lfexpr{param-types}\) means parameter types,
      \(\lfexpr{expr-dict}\) means logical form templates,
      and \(\lfexpr{arg-candidate}\) means a function that uses candidate expressions for the node class.
      In addition, the code defines the meta-action for the meta-node class \(\lfexpr{nl-token}\),
      which corresponds to the \(\lfexpr{\nltname}\) symbol in \(\lfexpr{\nlt{ *}}\) node classes.
      The meta-node class and the tokens from a seq2seq PLM create all actions for \(\lfexpr{\nlt{ *}}\) node classes.
    }
    \label{fig:grammar-code}
  \end{center}
\end{figure}

\subsection{Node Classes} \label{sec:node-classes}


Our grammar defines node classes for actions in \(\ActSetCom\) and \(\ActSetNlt\).
The node classes for actions in \(\ActSetCom\) are manually specified \Crefp{tab:kqapro-grammar-part-1,tab:kqapro-grammar-part-2,tab:overnight-grammar-part-1,tab:overnight-grammar-part-2}.
The node classes for actions in \(\ActSetNlt\) are converted from the natural language tokens of a seq2seq PLM by using a few rules.
Exceptionally, the special node class \(\lfexpr{program}\), which is used only to initialize an intermediate representation, does not yield a corresponding action, and the special rule \(\reduceact\) does not originate from any node class.

\newcommand{\skipthingswhenproofreading}{0}
\newcommand{\manywithoutproofreading}{1}
\if \manywithoutproofreading 1
  \if \proofreading 1
    \renewcommand{\skipthingswhenproofreading}{1}
    \fi
  \fi

\afterpage
{
  \begin{table}[t]  
    \begin{center}
      \caption{
        First part of node classes for actions in \(\ActSetCom\) for \kqapro.
      }

      \tablefontsizetwo
      \lffontsizetwo

      \if \skipthingswhenproofreading 0
        \input{\tablepathkqapro-grammar-part-1.tex\inputfilesuffix}
        \fi
      \label{tab:kqapro-grammar-part-1}
    \end{center}
  \end{table}
}

\solelyinpage{
  \begin{table}[t]  
    \begin{center}
      \caption{
        Second part of node classes for actions in \(\ActSetCom\) for \kqapro.
      }

      \tablefontsizetwo
      \lffontsizetwo

      \if \skipthingswhenproofreading 0
        \input{\tablepathkqapro-grammar-part-2.tex\inputfilesuffix}
        \fi
      \label{tab:kqapro-grammar-part-2}
    \end{center}
  \end{table}
}

\solelyinpage{
  \begin{table}[t]  
    \begin{center}
      \caption{
        First part of node classes for actions in \(\ActSetCom\) for \overnight.
      }

      \tablefontsizetwo
      \lffontsizetwo

      \if \skipthingswhenproofreading 0
        \input{\tablepathovernight-grammar-part-1.tex\inputfilesuffix}
        \fi
      \label{tab:overnight-grammar-part-1}
    \end{center}
  \end{table}
}

\solelyinpage{
  \begin{table}[t]  
    \begin{center}
      \caption{
        Second part of node classes for actions in \(\ActSetCom\) for \overnight.
      }

      \tablefontsizetwo
      \lffontsizetwo

      \if \skipthingswhenproofreading 0
        \input{\tablepathovernight-grammar-part-2.tex\inputfilesuffix}
        \fi
      \label{tab:overnight-grammar-part-2}
    \end{center}
  \end{table}
}

\clearpage
\subsection{Union Type Assignment} \label{sec:union-types}

The key differences between \(\lfexpr{\nlt{ *}}\) node classes are return types that are specified as union types.
To assign a proper return type to an \(\lfexpr{\nlt{ *}}\) node class, we use a few rules.
First, all \(\lfexpr{\nlt{ *}}\) node classes have common types that are included in return types. %
In \kqapro, a return type for an \(\lfexpr{\nlt{ *}}\) node class includes the following types by default:
\(\lfexpr{\lftyp{kp-concept}}\), \(\lfexpr{\lftyp{kp-entity}}\), \(\lfexpr{\lftyp{kp-relation}}\), \(\lfexpr{\lftyp{kp-attr-string}}\), \(\lfexpr{\lftyp{kp-attr-number}}\), \(\lfexpr{\lftyp{kp-attr-time}}\), \(\lfexpr{\lftyp{kp-q-string}}\), \(\lfexpr{\lftyp{kp-q-number}}\), \(\lfexpr{\lftyp{kp-q-time}}\), \(\lfexpr{\lftyp{vp-string}}\) and \(\lfexpr{\lftyp{vp-unit}}\).
Second, the return type of a node class additionally includes a specific type if the natural language token of the node class matches a pattern.
In \kqapro, the return type of a node class additionally includes \(\lfexpr{\lftyp{vp-quantity}}\), \(\lfexpr{\lftyp{vp-date}}\) or \(\lfexpr{\lftyp{vp-year}}\) if the natural language token of the node class is a number or a special character for the types;
for example, the return type of \(\lfexpr{\nltwb{7}}\) includes the three additional types, and the return type of \(\lfexpr{\nltwb{.}}\) includes \(\lfexpr{\lftyp{vp-quantity}}\), as the period character (\(\lfexpr{.}\)) is used for rational numbers (e.g., 3.14).

\begin{revision}
  \appendixsection{Working Example} \label{sec:working-example}

  We show details of constrained decoding with \(\ActHybr\) by building an intermediate representation for \kqapro step by step with \Cref{tab:working-example}.
  For each step, a production rule is applied to the leftmost non-terminal, which is identical to the left-hand side of the rule if neither sub-type inference nor a union type is applied.
  For example, in step 2, the rule \(\lfrule{obj-entity}{(filter-concept \lfnt{kw-concept} \lfnt{obj-entity})}\) is applied, where the rule's left-hand side is identical to the leftmost non-terminal and the rule's right-hand side is identical to the expanded expression in the next step. 
  In the following explanations, we address when our constrained decoding method uses sub-type inference, union types, candidate expressions, and mask caching.

  Sub-type inference is applied to the leftmost non-terminals in steps 0 and 7 by following the type hierarchy \Crefp{fig:kqapro-type-hierarchy}.
  In step 0, the non-terminal \(\lfexpr{\lfnt{result}}\) is expanded by applying
  the production rule \(\lfrule{result-entity-name}{(query-name \lfnt{obj-entity})}\),
  where \(\lfexpr{\lftyp{result-entity-name}}\) is a sub-type of \(\lfexpr{\lftyp{result}}\).
  In step 7, the non-terminal \(\lfexpr{\lfnt{obj-entity}}\) is expanded by applying
  the production rule \(\lfrule{obj-entity-with-attr}{(filter-number \lfnt{kw-attr-number} \lfnt{v-number} \lfnt{op-comparison} \lfnt{obj-entity})}\), 
  where \(\lfexpr{\lftyp{obj-entity-with-attr}}\) is a sub-type of \(\lfexpr{\lftyp{obj-entity}}\).
  Sub-type inference makes a seq2seq model skip predicting the rules that replace non-terminals of super types with non-terminals of sub-types; e.g. \(\lfrule{result}{\lfnt{result-entity-name}}\) and \(\lfrule{obj-entity}{\lfnt{obj-entity-with-attr}}\) are skipped.

  A union type is used for the left-hand side of a rule, which is applied to the leftmost non-terminal \(\lfexpr{\lfexpr{\lfnt{vp-quantity}}}\) in step 15.
  In the rule \(\lfrule{vp-quantity vp-year ...}{\nltwb{630}}\), the left-hand side \(\lfexpr{\lfnt{vp-quantity vp-year ...}}\) has a union-type \(\lfexpr{(\uniontyp \lftyp{vp-quantity} \lftyp{vp-year} ...)}\).
  The union-type enables this rule to be applied to different non-terminals including \(\lfexpr{\lfexpr{\lfnt{vp-quantity}}}\) and \(\lfexpr{\lfexpr{\lfnt{vp-year}}}\).

  Candidate expressions and mask caching are not used together, so candidate expressions are used in steps 4-6 and 9-12, and mask caching is used in the other steps.
  The parent nodes of leftmost non-terminals in the steps 4-6 and 9-12 are \(\lfexpr{keyword-concept}\) and \(\lfexpr{keyword-attribute-number}\) respectively, and the parent nodes have sets of candidate expressions \Crefp{tab:kqapro-candidate-expressions}.
  In contrast, mask caching increases decoding speed in the other steps in which candidate expressions are not used.

  In addition, after the action \(\reduceact\) is taken in step 18, the reduction operation is performed twice, then a nested complete sub-expression is constructed \CrefpContent{sec:implementation}{Intermediate representations}.
  First, the action \(\reduceact\) creates the complete sub-expression \(\lfexpr{(constant-unit \uline{reduce})}\).
  Then, the parent node \(\lfexpr{constant-number}\) has no more non-terminals as child nodes, so the parent node and its child nodes are reduced as a nested complete sub-expression \(\lfexpr{(constant-number (constant-quantity \nltwb{630} reduce) (constant-unit reduce))}\).

  \begin{landscape}
    \begin{table}[t]  
      \begin{center}
        \caption{
          \begin{revision}
            Example of building an intermediate representation by taking actions with \(\ActHybr\) for \kqapro.
            For each step, an action expands the leftmost non-terminal, written in bold, into a logical form expression, underlined in the next step.
            The utterance and the complete intermediate representation are \inCref{tab:kqapro-example-sp-concept}.
          \end{revision}
        }

        \tablefontsizefour
        \lffontsizethree
        \begin{revision}
          \input{\tablepathworking-example.tex\inputfilesuffix}
        \end{revision}

        \label{tab:working-example}
      \end{center}
    \end{table}
  \end{landscape}

\end{revision}

\begin{revision}
  \appendixsection{Search Algorithm Implementation of the Transformers Library} \label{sec:search-implementation}

  The implementation of a search algorithm in the transformers library \citep{DBLP:conf/emnlp/WolfDSCDMCRLFDS20} provides \texttt{LogitsProcessor}, which is an API to modify a logit that corresponds to an action's score \refeqp{eq:score-func-base}.
  Our implementation of constrained decoding uses \texttt{LogitsProcessor} to filter actions.
  However, \texttt{LogitsProcessor} does not have the ability to associate a past output sequence with a custom search state, such as \(\lfir{\actseqtmo}\).
  Therefore, retrieving \(\lfir{\actseqtmo}\) from a past output sequence costs \(O(\tstep)\) time.
  If the implementation of a search algorithm is modified to track past output sequences in linked lists,
  a custom search state can be associated with a linked list node and retrieved later in \(O(1)\) time.
\end{revision}

\appendixsection{Alternative Decoding Speed Optimization} \label{sec:alternative-decoding-speed-optimization}

\begin{table}[t]  
  \begin{center}

    \caption{
      Average time to decode an output sequence with the alternative algorithm that increases the speed of constrained decoding.
      When the algorithm is disabled,
      \Cref{alg:compute-valid-actions} does not consider the size \(\setsize{\ActTypeFn{\lfir{\actseqic}}}\),
      then the cache memory \(\Cache\) is only used to store \(\Tuple{\ActTypeFn{\lfir{\actseqic}}, \True}\),
      and \(\textsc{ComputeActionsValidness}(\actseqic, \Tensor)\) is always \(\Tuple{\ActHybrFn{\lfir{\actseqic}}, \True}\).
    }
    \begin{subtable}[h]{\textwidth}
      \begin{center}
        \caption{
          \kqapro
        }
        \tablefontsizeone
        \input{\tablepathkqapro-decoding-time-alternative-only.tex\inputfilesuffix}
        \label{tab:kqapro-decoding-time-alternative-only}
      \end{center}
    \end{subtable}

    \vspace{\subtablevspacelen}

    \begin{subtable}[h]{\textwidth}
      \begin{center}
        \caption{
          \overnight
        }
        \tablefontsizeone
        \input{\tablepathovernight-decoding-time-alternative-only.tex\inputfilesuffix}
        \label{tab:overnight-decoding-time-alternative-only}
      \end{center}
    \end{subtable}
    \tablefontsizeone
    \label{tab:decoding-time-alternative-only}
  \end{center}
\end{table}

\begin{algorithm}[t]
  \begin{center}
    \input{\algorithmpathcompute-actions-validness.tex\inputfilesuffix}
    \caption{
      Method to compute a set that consists of either all valid actions or all invalid actions.
      The input values are a past action sequence \(\actseqic\) and a cache memory \(\Cache\).
      The return value is either \mbox{\(\Tuple{\ActHybrFn{\lfir{\actseqic}}, \True}\)} or \mbox{\(\Tuple{\ActSet - \ActHybrFn{\lfir{\actseqic}}, \False}\)}.
    }
    \label{alg:compute-valid-actions}
  \end{center}
\end{algorithm}

\begin{algorithm}[t!]
  \begin{center}
    \input{\algorithmpathcompute-mask-tensor-with-validness.tex\inputfilesuffix}
    \caption{
      Method to compute a mask tensor for a batch of action sequences.
      The input values are a batch \(\Batch\) of past action sequences and a cache memory \(\Cache\).
      The return value is a scoring mask \(\Tensor\).
      The method reduces the number of iterations that are needed to compute a mask tensor.
    }
    \label{alg:compute-mask-tensor-with-validness}
  \end{center}
\end{algorithm}

For \(\ActBase \in \SetOfActsWC\), we observed that \(\setsize{\ActBaseFn{\lfir{\actseqic}}}\) is usually very small or large.
For example, in \kqapro, when the leftmost non-terminal \(\lmnt(\lfir{\actseqic})\) has the type \(\lfexpr{\lftyp{op-direction}}\), \(\ActBaseFn{\lfir{\actseqic}}\) includes actions that produce \(\lfexpr{'forward}\) and \(\lfexpr{'backward}\), then \(\setsize{\ActBaseFn{\lfir{\actseqic}}} = 2\).
As another example, in \kqapro, when the leftmost non-terminal \(\lmnt(\lfir{\actseqic})\) has the type \(\lfexpr{\lftyp{vp-string}}\), \(\ActBaseFn{\lfir{\actseqic}} = \ActSetNlt\), then \(\setsize{\ActBaseFn{\lfir{\actseqic}}} = 50,260\).

We devise an algorithm that increases the speed of constrained decoding by depending on the size \(\setsize{\ActBaseFn{\lfir{\actseqic}}}\) \Crefp{alg:compute-valid-actions,alg:compute-mask-tensor-with-validness}.
\Cref{alg:compute-valid-actions} retrieves either (1) a set of valid actions from trie data structures for candidate expressions, or (2) a set of valid actions or a set of invalid actions from a cache memory for types.
\Cref{alg:compute-mask-tensor-with-validness} takes the retrieved set of actions as an input, then computes a mask tensor, where the number of iterations with a CPU is reduced.
The algorithm extends the constrained decoding method of \citet{DBLP:conf/iclr/CaoI0P21} which is implemented as \texttt{PrefixConstrainedLogitsProcessor} in the transformers library \citep{DBLP:conf/emnlp/WolfDSCDMCRLFDS20}.

We measured the average decoding time with the alternative algorithm \Crefp{tab:decoding-time-alternative-only}.
Similarly to mask caching \Crefp{tab:kqapro-decoding-time,tab:overnight-decoding-time}, the algorithm achieved meaningful decrease in decoding time, compared to the results of constrained decoding without any algorithm.
In particular, the effect of the algorithm is noticeable when a beam size or a batch size is large.
However, the algorithm is less effective than mask caching.
\ignore{
  This means that our algorithms greatly reduce the overhead that occurs from our constraints.
  Therefore, search algorithms, such as beam search, which perform concurrent operations, benefit from the reduction in the overhead.
  In addition, \(\ActTypeWU\) and \(\ActType\) greatly benefit from the algorithms, because \(\ActTypeWUFn{\lfir{\actseqic}}\) and \(\ActTypeFn{\lfir{\actseqic}}\) have many actions in \(\ActSetNlt\) as valid actions due to the absence of candidate expressions.
}

\ignore{
  We note that the decoding time is proportional to the length of an output sequence.
  As ours and \bartkopl respectively have 28.8 and 35.1 as the average lengths of output sequences, ours with \(\ActNone\) is slightly faster than \bartkopl, although both do not use any constraint.
  \bartkopl tokenizes symbols, such as function names, then its output sequences include slight more tokens.
  If sub-type inference \Crefp{sec:grammars-with-types} is not applied to ours, we should additionally introduce actions that convert a non-terminal with a super-type to a non-terminal with a sub-type; an example action is \(\lfexpr{\lfnt{result}} \rightarrow \lfexpr{\lfnt{result-rel-q-value}}\).
  The average lengths of output sequences of ours without sub-type inference is 32.8, then decoding slightly gets slower.
}

\appendixsection{Derivation of \(\nabla_\theta \jmml(\theta \scbar \uttersc, \lfgd)\)} \label{sec:derivation-of-grad}

\if \skipthingswhenproofreading 0
\begingroup
\begin{align*}
  %
  %
  \nabla_\theta \jmml(\theta \scbar \uttersc, \lfgd)
    & = \nabla_\theta \log p_\theta(\lfgd \mid \uttersc) \\
    & = \frac{1}{p_\theta(\lfgd \mid \uttersc)} \nabla_\theta p_\theta(\lfgd \mid \uttersc) \\
    & = \frac{1}{p_\theta(\lfgd \mid \uttersc)} \nabla_\theta \sum_{\actseq \in \ActSeqSet} p(\lfgd \mid \actseq, \uttersc) p_\theta(\actseq \mid \uttersc)\\
    & = \frac{1}{p_\theta(\lfgd \mid \uttersc)} \sum_{\actseq \in \ActSeqSet} p(\lfgd \mid \actseq, \uttersc) \nabla_\theta p_\theta(\actseq \mid \uttersc) \\
    & = \frac{1}{p_\theta(\lfgd \mid \uttersc)} \sum_{\actseq \in \ActSeqSet} p(\lfgd \mid \actseq, \uttersc) p_\theta(\actseq \mid \uttersc) \frac{1}{p_\theta(\actseq \mid \uttersc)} \nabla_\theta p_\theta(\actseq \mid \uttersc) \\
    & = \frac{1}{p_\theta(\lfgd \mid \uttersc)} \sum_{\actseq \in \ActSeqSet} p(\lfgd \mid \actseq, \uttersc) p_\theta(\actseq \mid \uttersc) \left( \frac{1}{p_\theta(\actseq \mid \uttersc)} \nabla_\theta p_\theta(\actseq \mid \uttersc) \right) \\
    & = \frac{1}{p_\theta(\lfgd \mid \uttersc)} \sum_{\actseq \in \ActSeqSet} p(\lfgd \mid \actseq, \uttersc) p_\theta(\actseq \mid \uttersc) \nabla_\theta \log p_\theta(\actseq \mid \uttersc) \\
    & = \sum_{\actseq \in \ActSeqSet} \frac{1}{p_\theta(\lfgd \mid \uttersc)} p(\lfgd \mid \actseq, \uttersc) p_\theta(\actseq \mid \uttersc) \nabla_\theta \log p_\theta(\actseq \mid \uttersc) \\
    & = \sum_{\actseq \in \ActSeqSet} \frac{1}{p_\theta(\lfgd \mid \uttersc)} p_\theta(\actseq, \lfgd \mid \uttersc) \nabla_\theta \log p_\theta(\actseq \mid \uttersc) \\
    & = \sum_{\actseq \in \ActSeqSet} \frac{p(\uttersc)}{p_\theta(\uttersc, \lfgd)} \frac{p_\theta(\actseq, \uttersc, \lfgd)}{p(\uttersc)} \nabla_\theta \log p_\theta(\actseq \mid \uttersc) \\
    & = \sum_{\actseq \in \ActSeqSet} \frac{p_\theta(\actseq, \uttersc, \lfgd)}{p_\theta(\uttersc, \lfgd)} \nabla_\theta \log p_\theta(\actseq \mid \uttersc) \\
    & = \sum_{\actseq \in \ActSeqSet} p_\theta(\actseq \mid \uttersc, \lfgd) \nabla_\theta \log p_\theta(\actseq \mid \uttersc) \\
    & = \mathop{\mathbb{E}}_{p_\theta(\actseq \mid \uttersc, \lfgd)} \nabla_\theta \log p_\theta(\actseq \mid \uttersc) \\
  \\
  \text{where\qquad\qquad\quad}& \\
    p_\theta(\actseq \mid \uttersc, \lfgd)
    &= \frac{p(\lfgd \mid \actseq, \uttersc) p_\theta(\actseq \mid \uttersc)}{\sum_{\actseq' \in \ActSeqSet} p(\lfgd \mid \actseq', \uttersc) p_\theta(\actseq' \mid \uttersc)} \\[1ex]
    &= \frac{p(\lfgd \mid \actseq) p_\theta(\actseq \mid \uttersc)}{\sum_{\actseq' \in \ActSeqSet} p(\lfgd \mid \actseq') p_\theta(\actseq' \mid \uttersc)} && \because x \independent \lfgd \mid \actseq \qquad \forall \actseq \in \ActSeqSet\\
  \\
  p(\lfgd \mid \actseq) &= \begin{cases}
    1, &\textrm{if \(\denot = \lfgd\)} \\
    0, &\textrm{otherwise} \end{cases} \label{eq:lf-score} .
\end{align*}
\endgroup
\fi
\clearpage  

\appendixsection{Qualitative Analysis} \label{sec:qualitative-analysis}

We compare the outputs of semantic parsing with \(\ActHybr\) and with \(\ActType\) \Crefp{tab:kqapro-example-sp-concept,tab:kqapro-example-sp-entity,tab:kqapro-example-sp-relation,tab:overnight-example-sp-basketball-relation-entity,tab:overnight-example-sp-calendar-entity}, where models are trained with strong supervision from \(\SplitTrain\) \Crefp{sec:results-on-strongsup}.
Examples for the comparisons are extracted from \(\SplitVal\) for \kqapro and from \(\SplitTest\) for \overnight.
For an utterance \(\uttersc\), an intermediate representation \(\lfir{\actseq}\) and a logical form \(\lflf{\actseq}\), we highlight the parts that correspond to a candidate expression.
The highlighted part in \(\uttersc\) has a different representation from that of the corresponding candidate expression.
With the guidance of \(\ActHybr\), the highlighted parts in \(\lfir{\actseq}\) and \(\lflf{\actseq}\) are valid KB elements.
In contrast, with the guidance of \(\ActType\), the highlighted parts in \(\lfir{\actseq}\) and \(\lflf{\actseq}\) are invalid, as a semantic parser copies the parts from \(\uttersc\), or as a semantic parser generates sub-action sequences in \(\SplitTrain\).
We also shows the effectiveness of candidate expressions for the node class \(\lfexpr{keyword-ent-type}\) in the Restaurants domain of \overnight (\Cref{sec:effect-of-candidate-expressions} and \Cref{tab:overnight-example-sp-no-kw-ent-type-in-restaurants}).

\begingroup
\newcommand{\candexpr}[1]{\(\lfexpr{#1}\)}

\begin{table}[t]  
  \begin{center}
    \lffontsizecaption
    \caption{
      Example of semantic parsing for \kqapro when \(\ActHybr\) is effective for \(\lfexpr{keyword-concept}\).
      The correct candidate expression is \candexpr{game show} rather than \candexpr{game}.
    }

    \tablefontsizefive
    \lffontsizetwo

    \if \skipthingswhenproofreading 0
      \input{\tablepathkqapro-example-sp-concept.tex\inputfilesuffix}
      \fi
    \label{tab:kqapro-example-sp-concept}
  \end{center}
\end{table}

\begin{table}[t]  
  \begin{center}
    \lffontsizecaption
    \caption{
      Example of semantic parsing for \kqapro when \(\ActHybr\) is effective for \(\lfexpr{keyword-entity}\).
      The correct candidate expression is \candexpr{Tilda Swinton} rather than \candexpr{Tilde Swinton}.
    }

    \tablefontsizefive
    \lffontsizetwo

    \if \skipthingswhenproofreading 0
      \input{\tablepathkqapro-example-sp-entity.tex\inputfilesuffix}
      \fi
    \label{tab:kqapro-example-sp-entity}
  \end{center}
\end{table}

\begin{table}[t]  
  \begin{center}
    \lffontsizecaption
    \caption{
      Example of semantic parsing for \kqapro when \(\ActHybr\) is effective for \(\lfexpr{keyword-relation}\).
      The correct candidate expression is \candexpr{ethnic group} rather than \candexpr{ethnic community}.
    }

    \tablefontsizefive
    \lffontsizetwo

    \if \skipthingswhenproofreading 0
      \input{\tablepathkqapro-example-sp-relation.tex\inputfilesuffix}
      \fi
    \label{tab:kqapro-example-sp-relation}
  \end{center}
\end{table}

\begin{table}[t]  
  \begin{center}
    \lffontsizecaption
    \caption{
      Example of semantic parsing for the Basketball domain of \overnight when \(\ActHybr\) is effective for \(\lfexpr{keyword-relation-entity}\).
      The correct candidate expression is \candexpr{team} rather than \candexpr{birthplace}, which is a candidate expression of \(\lfexpr{keyword-relation-entity}\) in the Socialnetwork domain.
    }

    \tablefontsizefive
    \lffontsizetwo

    \if \skipthingswhenproofreading 0
      \input{\tablepathovernight-example-sp-basketball-relation-entity.tex\inputfilesuffix}
      \fi
    \label{tab:overnight-example-sp-basketball-relation-entity}
  \end{center}
\end{table}

\begin{table}[t]  
  \begin{center}
    \lffontsizecaption
    \caption{
      Example of semantic parsing for the Calendar domain of \overnight when \(\ActHybr\) is effective for \(\lfexpr{keyword-entity}\).
      The correct candidate expression is \candexpr{greenberg cafe} rather than \candexpr{greenbug cafe}.
    }

    \tablefontsizefive
    \lffontsizetwo

    \if \skipthingswhenproofreading 0
      \input{\tablepathovernight-example-sp-calendar-entity.tex\inputfilesuffix}
      \fi
    \label{tab:overnight-example-sp-calendar-entity}
  \end{center}
\end{table}

\ignore{

  \begin{table}[t]  
    \begin{center}
      \lffontsizecaption
      \caption{
        Example of semantic parsing for the Housing domain of \overnight when \(\ActHybr\) is effective for \(\lfexpr{keyword-ent-type}\).
        The correct candidate expression is \candexpr{housing unit} rather than \candexpr{housing}.
      }

      \tablefontsizefive
      \lffontsizetwo

      \if \skipthingswhenproofreading 0
        \input{\tablepathovernight-example-sp-housing-ent-type.tex\inputfilesuffix}
        \fi
      \label{tab:overnight-example-sp-housing-ent-type}
    \end{center}
  \end{table}
}
\begin{table}[t]  
  \begin{center}
    \lffontsizecaption
    \caption{
      Example of semantic parsing for the Restaurants domain of \overnight when \(\ActHybr\) is effective for \(\lfexpr{keyword-ent-type}\).
      The correct candidate expression is \candexpr{food} rather than \candexpr{meal}.
    }

    \tablefontsizefive
    \lffontsizetwo

    \if \skipthingswhenproofreading 0
      \input{\tablepathovernight-example-sp-no-kw-ent-type-in-restaurants.tex\inputfilesuffix}
      \fi
    \label{tab:overnight-example-sp-no-kw-ent-type-in-restaurants}
  \end{center}
\end{table}

\endgroup

\end{document}